\documentclass[journal]{IEEEtran}

\usepackage[monochrome]{xcolor}
\usepackage{amsmath,amsfonts}
\usepackage{dsfont}
\usepackage{soul}
\usepackage{algorithm}
\usepackage{algorithmic}
\usepackage{array}
\usepackage{subfig}
\usepackage{textcomp}
\usepackage{stfloats}
\usepackage{url}
\usepackage{verbatim}
\usepackage{graphicx}
\usepackage{bm}
\usepackage[style=ieee,backend=biber]{biblatex}
\usepackage{indentfirst}
\usepackage{lettrine}
\usepackage{afterpage}
\usepackage{tikz}
\usepackage{mathdots}
\usepackage{yhmath}
\usepackage{cancel}
\usepackage{siunitx}
\usepackage{array}
\usepackage{multirow}
\usepackage{amssymb}
\usepackage{gensymb}
\usepackage{tabularx}
\usepackage{extarrows}
\usepackage{booktabs}
\usepackage{placeins}
\usepackage{setspace}
\usetikzlibrary{fadings}
\usetikzlibrary{patterns}
\usetikzlibrary{shadows.blur}
\usetikzlibrary{shapes}
\usepackage{tikzscale}
\usepackage{xcolor}

\newcolumntype{Y}{>{\centering\arraybackslash}X}

\definecolor{dgreen}{rgb}{0.05, 0.5, 0.06}
\title{Multi-Sensor Diffusion-Driven Optical Image Translation for Large-Scale Applications}
\author{
  João Gabriel Vinholi, Marco Chini, \textit{Senior Member, IEEE}, Anis Amziane, Renato Machado, \textit{Senior Member, IEEE}, Danilo Silva, \textit{Member, IEEE}, Patrick Matgen
\thanks{
\textbf{This is the accepted version of the manuscript published in IEEE Journal of Selected Topics in Applied Earth Observations and Remote Sensing (JSTARS). Please access the final version at IEEEXplore (Open Access). DOI 10.1109/JSTARS.2024.3506032.}\\
This technology is protected by a patent filed on 23 december 2023 at Office Luxembourgeois de la propriété intellectuelle (LU505861).\\
This work has been supported by the Luxembourg Directorate of Defence through the Chameleon project.\\
João Gabriel Vinholi, Marco Chini, Anis Amziane, and Patrick Matgen are with the Luxembourg Institute of Science and Technology (LIST), Esch-sur-Alzette, Luxembourg. E-mails: \{joao.vinholi, marco.chini, anis.amziane, patrick.matgen\}@list.lu\\
João Gabriel Vinholi is also with the Department of Telecommunications, Aeronautics Institute of Technology (ITA), São José dos Campos, SP, Brazil.\\
Renato Machado is with the Department of Telecommunications, Aeronautics Institute of Technology (ITA), São José dos Campos, SP, Brazil. E-mail: rmachado@ita.br\\
Danilo Silva is with the Department of Electric and Electronic Engineering, Universidade Federal de Santa Catarina (UFSC), Florianópolis, SC, Brazil. E-mail: danilo.silva@ufsc.br
}
}

\begin{document}

\maketitle

\begin{abstract}
\color{dgreen}
Comparing images captured by disparate sensors is a common challenge in remote sensing. This requires image translation---converting imagery from one sensor domain to another while preserving the original content. Denoising Diffusion Implicit Models (DDIM) are potential state-of-the-art solutions for such domain translation due to their proven superiority in multiple image-to-image translation tasks in computer vision. However, these models struggle with reproducing radiometric features of large-scale multi-patch imagery, resulting in inconsistencies across the full image. This renders downstream tasks like Heterogeneous Change Detection impractical. 
To overcome these limitations, we propose a method that leverages denoising diffusion for effective multi-sensor optical image translation over large areas. Our approach super-resolves large-scale low spatial resolution images into high-resolution equivalents from disparate optical sensors, ensuring uniformity across hundreds of patches. Our contributions lie in new forward and reverse diffusion processes that address the challenges of large-scale image translation. 
Extensive experiments using paired Sentinel-II (10m) and Planet Dove (3m) images demonstrate that our approach provides precise domain adaptation, preserving image content while improving radiometric accuracy and feature representation.  
A thorough image quality assessment and comparisons with the standard DDIM framework and five other leading methods are presented.
We reach a mean Learned Perceptual Image Patch Similarity (mLPIPS) of 0.1884 and a Fréchet Inception Distance (FID) of 45.64, expressively outperforming all compared methods, including DDIM, ShuffleMixer, and SwinIR. 
The usefulness of our approach is further demonstrated in two Heterogeneous Change Detection tasks.
\color{black}
\end{abstract}

\section{Introduction}
\label{sec:intro}

\lettrine{I}{mage-to-image} translation (I2I) involves transforming an image from one data domain to another while preserving its original content. This technique is useful for applications such as super-resolution, inpainting, image decompression, and domain adaptation  \cite{Saharia2021, Dhariwal2021, I2ISeo2023, Park2019, I2IIsmael2023, I2ITasar2020, I2IZhang2023, I2ISokolov2023}.

In remote sensing, I2I is valuable when comparing images from different sensors. For example, a deep learning model can translate a low-resolution domain $A$ optical image to a high-resolution domain $B$ optical image.This creates synthetic high-resolution images from low-resolution ones, which are useful when original high-resolution images are unavailable, costly, or affected by weather conditions. However, I2I in this context is challenging due to differences in spatial and spectral characteristics, noise levels, and geometric distortions specific to each sensor. Therefore, an I2I model must preserve the content and structure of the source image while adapting to the radiometric properties and spatial resolution of the target domain.

\color{dgreen}
Despite advancements in multi-sensor image-to-image (I2I) translation for remote sensing, most literature focuses on generating individual synthetic patches for augmenting training data sets used in downstream tasks \cite{I2ITasar2020, I2IIsmael2023, I2ISokolov2023}, on SAR to Optical translation \cite{CDLi2021, I2IZhang2023, I2ISeo2023}, or in super-resolution without tackling multisensor discrepancies beyond resolution differences \cite{yi_tip2024, Min2024, Haut2018}. While highly valuable, these methods do not address the challenge of translating \mbox{large-scale} images across disparate optical sensors with different spatial resolutions, particularly ensuring feature consistency between input and output images and preserving the radiometric characteristics of each sensor. This gap is significant as it hinders the fidelity of the translated images, where preserving salient features without introducing \mbox{non-existent} elements is crucial. Additionally, these methods are usually tested on benchmark datasets or small datasets, which may not adequately demonstrate their effectiveness in \mbox{real-world} scenarios. 
\color{black}

An appealing choice for addressing these challenges is the use of Denoising Diffusion Models (DDMs) \cite{Ho2020}, which have shown exceptional image generation capabilities in other contexts, surpassing adversarial models without common training issues like mode-collapse \cite{mansourifar_i3eaccess2022}. However, standard diffusion methods seem to struggle with high variability in overall patch colorization across samples with similar features in a vast data set. Also, at times they produce inconsistencies in feature generation across neighboring patches due to their stochastic nature \cite{Ho2020}.

To overcome these limitations, we propose a novel method that leverages Denoising Diffusion Implicit Models (DDIM) for effective optical image translation over large areas. Our approach super-resolves large-scale low spatial resolution images into high-resolution equivalents from disparate optical sensors, ensuring uniformity across hundreds of patches. We address the issues of high variability in patch colorization and inconsistencies in feature generation by designing novel forward and reverse diffusion procedures. These procedures enhance spatial resolution and align radiometric differences between domains, ensuring high-quality, consistent results.

Notably, this work is the first to integrate large-scale optical domain adaptation and super-resolution into a single task using a DDPM-based method. This integration generates large multi-patch images with consistent and color-accurate patches, showing significant visual resolution improvement while preserving the original content (see Section \ref{sec:experiments:imagen}). The robust training dataset and scalability of our approach contribute to its state-of-the-art performance, making it suitable for practical applications in remote sensing.

We apply our DDM-based method to generate synthetic high-resolution images for eight regions of interest: Beirut, Austin, Tlaquepaque, and five other locations. In our experiments, Sentinel-II imagery ($10\text{m}^2/\text{pixel}$) is translated into synthetic versions of Planet Dove imagery ($3\text{m}^2/\text{pixel}$). We perform two heterogeneous change detection (HCD) experiments to test the method in a practical application. HCD compares images from different sensors or modes of acquisition to identify changes in a specific area. Valuable for quick disaster damage assessment, achieving accurate and consistent results is challenging due to differences in image characteristics and quality \cite{Chini2023,Touati2020,Zhang2022,WangCD2022,Lv2022,Wu2022}.

While we present heterogeneous change detection (HCD) experiments to showcase one application of our image-to-image (I2I) translation algorithm, it is important to note that these experiments are not intended to demonstrate peak performance in HCD. The simplistic image processing algorithm used in these experiments does not exploit the resolution gains from our diffusion-based image translation model. Instead, the purpose is to illustrate the potential use of our I2I translation method in HCD. Developing a learning-based HCD method that fully leverages the added resolution is beyond the scope of this article.

Our contributions are summarized as follows:
\begin{enumerate}
    \item We propose original training and testing procedures based on denoising diffusion models that translate optical images from a low spatial resolution domain to a higher one over large areas while maintaining high \mbox{inter-patch} and \mbox{input-output} consistency.
    \item We present image quality comparisons between our method and other \mbox{cutting-edge} solutions, along with ablation experiments over hundreds of square kilometers of imagery, highlighting the contributions of our approach compared to traditional denoising diffusion.
    \item We present heterogeneous change detection tests in Beirut, Lebanon, and Austin, USA, indicating the potential usefulness of our method in HCD tasks.
\end{enumerate}

\section{Related Works}\label{sec:apx:relworks}
\subsubsection{Denoising Diffusion Models}
Ho et al. \cite{Ho2020} proposed the use of a denoising framework for image generation, the so-called Denoising Diffusion Probabilistic Model (DDPM). It showed incredible image generation quality and overcame most state-of-the-art methods available at the time. However, the inference process was cumbersome since hundreds of model evaluations per image were necessary to reach optimal performance. The work \textit{Denoising Diffusion Implicit Models} (DDIM) \cite{Song2020} approached this issue by presenting a new interpretation of the reverse diffusion process. Differently from the work in \cite{Ho2020}, they treated reverse diffusion as a non-Markovian process by skipping multiple reverse diffusion steps. Furthermore, they experimented with eliminating the randomness of intermediate diffusion steps---which came from the addition of a random noise matrix to each reverse diffusion step result---and noticed that this enabled an even greater reduction of model evaluations during inference at the cost of sample variety. In practice, these combined modifications reduced the necessary number of model evaluations during inference by up to two orders of magnitude while maintaining high generation quality. 

As originally defined in \cite{Ho2020}, denoising diffusion models are unconditional, i.e., they are trained to generate images with distributions similar to those it has seen during training without any prior knowledge about how a particular image should resemble or what it should contain. This limited the applicability of such models in multiple condition-guided tasks, such as image inpainting, super-resolution, colorization, and domain adaptation. To make them possible, two works \cite{Nichol2021, Saharia2021} experimented with adding a conditioning matrix as an additional model input. This partially worked since, as observed by Dhariwal et al. \cite{Dhariwal2021}, diffusion models have a tendency to ignore the condition and focus only on the noisy input to be denoised. Furthermore, the work in \cite{Dhariwal2021} proposed a possible strategy to alleviate this issue: the use of a separate classifier through which conditioning information is directly inserted into the reverse diffusion procedure. This proved to be useful, as the images generated with this procedure were strongly linked with the conditions added to the model. Still, the need to have a second classifier model is cumbersome and sometimes impractical. Due to that, Ho et al. \cite{ho2022classifierfree} proposed \textit{Classifier-free Guidance}, which, as the name implies, removes the need for a separate classifier network that guides the model with conditioning information. They accomplish that by making a small modification in the DDPM training and inference procedures that, in summary, treats the denoising model as if it consisted of a combination of an unconditional and a conditional model. Hence, during inference, it became possible to determine how much the conditional and unconditional virtual parts of the model influenced the generation and, therefore, how strongly the reverse diffusion should rely on the conditioning information. 

\subsubsection{Image-to-Image Translation}
In the context of image-to-image translation, Saharia et al. \cite{Saharia2021} investigated the use of DDPMs in different I2I translation tasks: colorization, image restoration, inpainting, and uncropping, obtaining remarkable image generation quality and high sample diversity. Moreover, in the presented super-resolution tests, it beat deep regression-based models by an impressively high margin. The work in \cite{Wang2022} used classifier-free guidance \cite{ho2022classifierfree} in the task of semantic image generation---where the user inputs the information on what each area of the generated image should contain---together with a novel conditional model architecture that includes multi-head attention-based mechanisms. In the considered test sets, they beat all state-of-the-art methods in image quality metrics. \textit{Image Super-Resolution via Iterative Refinement} \cite{Saharia2021SR3} employed DDPM in the task of super-resolution. They trained a conditional model, referred to as SR3, to generate high-resolution samples starting from their low-resolution versions, beating all compared methods in image quality metrics and fool rate tests.

Other works focused on remote sensing-related tasks. Ismael et al. \cite{I2IIsmael2023} proposed a GAN-based I2I translation method with the goal of performing remote sensing semantic segmentation of optical images from unknown domains using a segmentation network trained with optical images from a single domain. The methods in \cite{I2ITasar2020} and \cite{I2ISokolov2023} proposed style-transferring domain adapting frameworks that generate extra training data matching the appearance of a particular same-resolution domain to alleviate train/test data drift. The method in \cite{CDLi2021} uses a cycle-consistent GAN to translate optical images into SAR images to perform change detection. It showed promising change detection results. However, training and testing data are limited to just a few acquisitions.
The method in \cite{I2IZhang2023} is a thermodynamics-inspired translation network that obtained interesting results in SAR-to-Optical translation tasks with its unconventional design. 
The work in \cite{I2ISeo2023} proposes a DDM-based Brownian Bridge method for SAR-to-Optical I2I translation. There, the authors suggest the use of a modified forward diffusion procedure that, instead of resulting in a pure-noise image, the target SAR image is iteratively transformed into the source optical image. They reason that this could help the model more easily understand the relationships across domains.
Xiao et al.~\cite{Xiao2023} propose a diffusion-based super-resolution method that seeks to further explore prior information from the low-resolution input. When tested on a small-scale data set, it performed well.

\color{dgreen}
In the context of remote sensing super-resolution, \cite{yi_tip2024} focuses on improving image super-resolution techniques in large-area earth observation scenes by addressing issues with redundant token representation and single-scale modeling. They introduce the Top-k Token Selective Transformer (TTST), which uses a Residual Token Selective Group (RTSG) and a Multi-scale Feed-forward Layer (MFL) for better feature aggregation and accurate reconstruction. The TTST outperforms existing methods without attempting multi-sensor translation.
Min et al. \cite{Min2024} introduce a domain matching (DM) module that enhances reference-based image super-resolution (RefSR) models by addressing domain gaps between reference and low-resolution images in remote sensing. This DM-based RefSR approach outperformed the compared methods. However, its reliance on reference image features can potentially introduce unwanted artifacts not present in the input image, and it requires co-registered reference images with similar geometry of acquisition, limiting its practical applicability.
Haut et al. \cite{Haut2018} present an unsupervised convolutional generator model for super-resolving low-resolution remote sensing data. Data is symmetrically projected to the target resolution while ensuring reconstruction constraints over the LR input image. This study did not attempt multi-sensor translation, but rather downsampled high-resolution images to create low-resolution counterparts.
\color{black}

\begin{figure*} 
\centering 
\subfloat[Beirut Pre Setninel-II]{
\includegraphics[width=.4\linewidth]{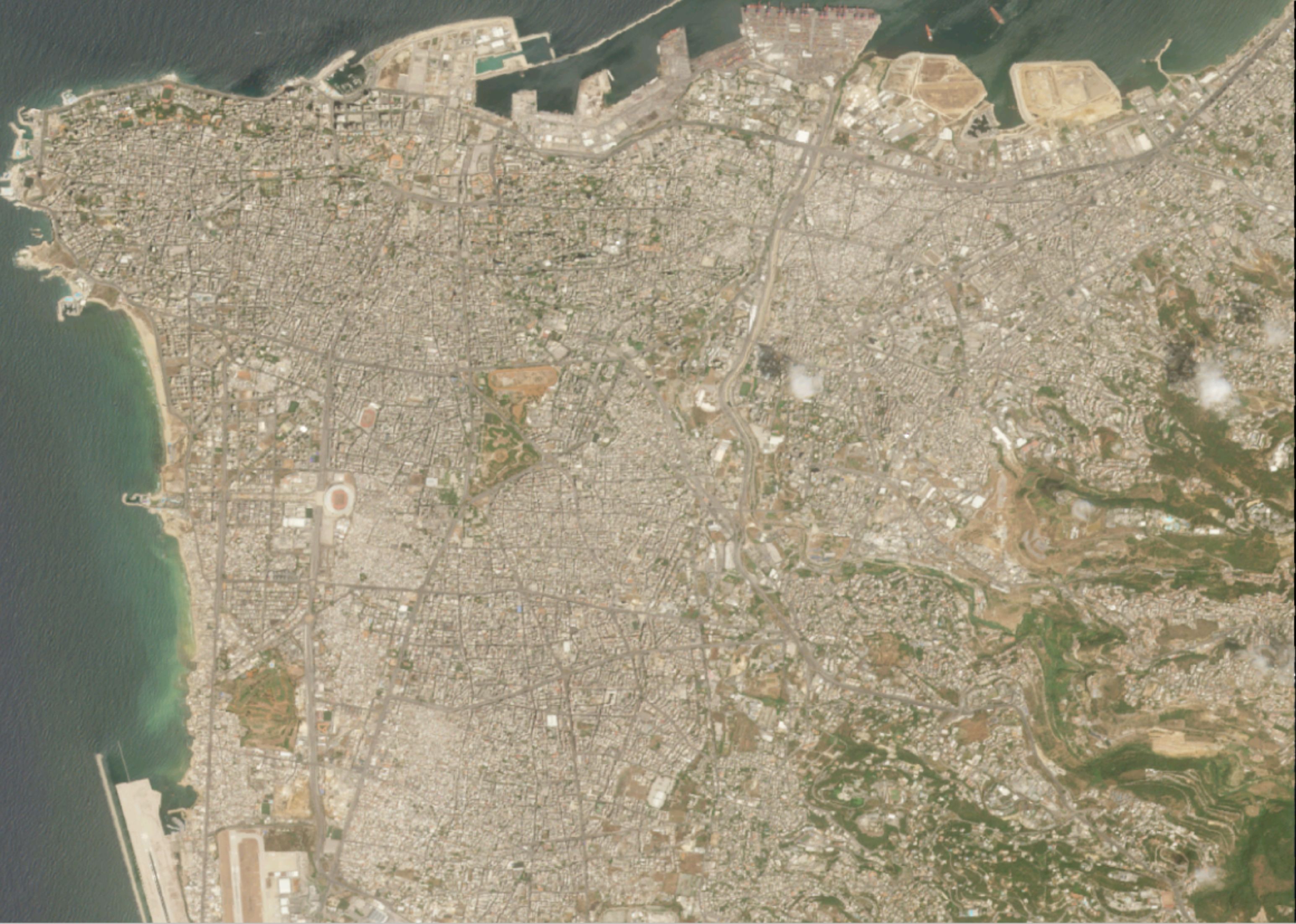} 
\label{fig:pre_s2} 
} 
\hspace{5mm}
\subfloat[Beirut Pre Planet Dove]{
\includegraphics[width=.4\linewidth]{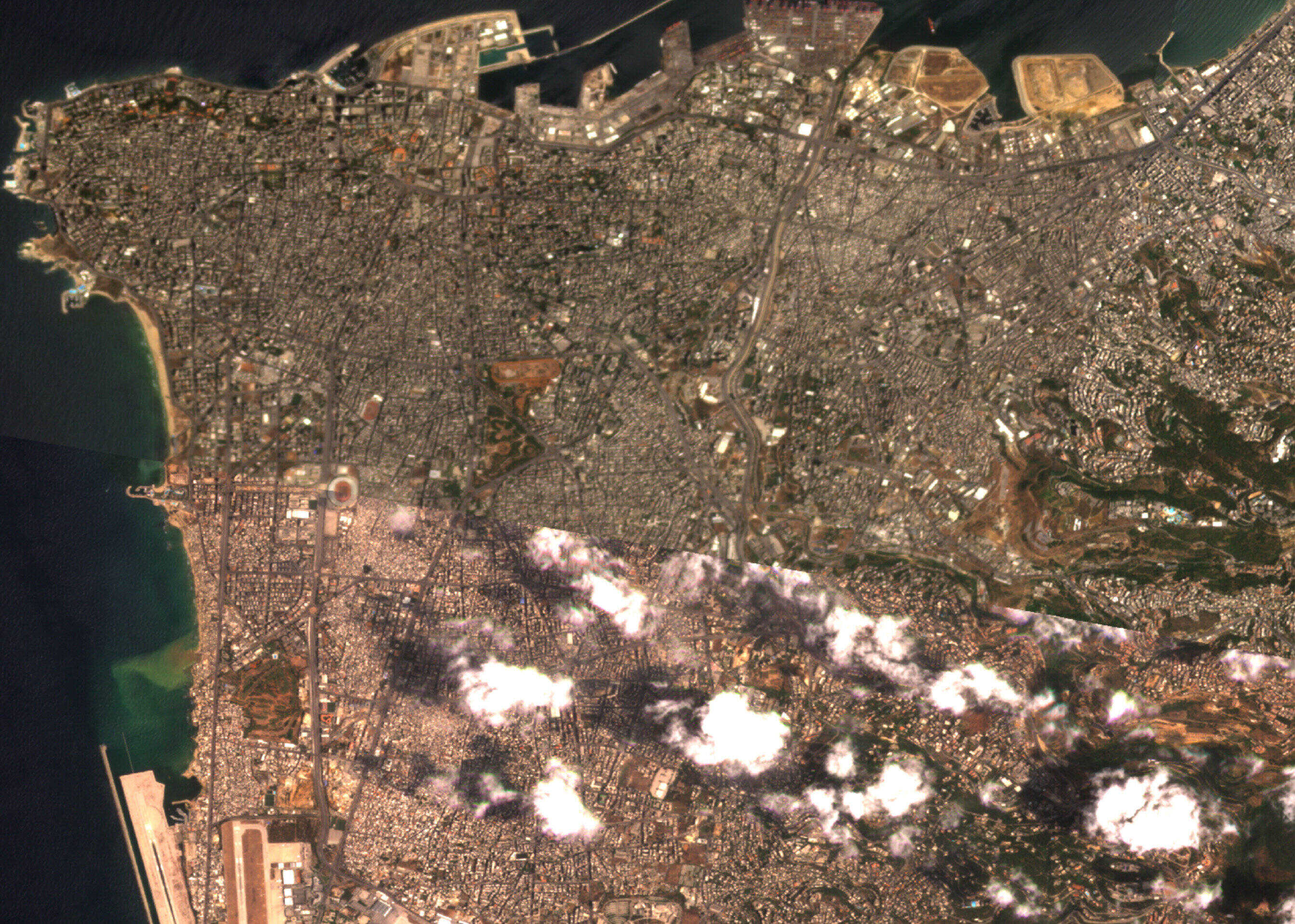} 
\label{fig:pre_planet} 
} 
\\ 
\subfloat[Beirut Post Setninel-II]{
\includegraphics[width=.4\linewidth]{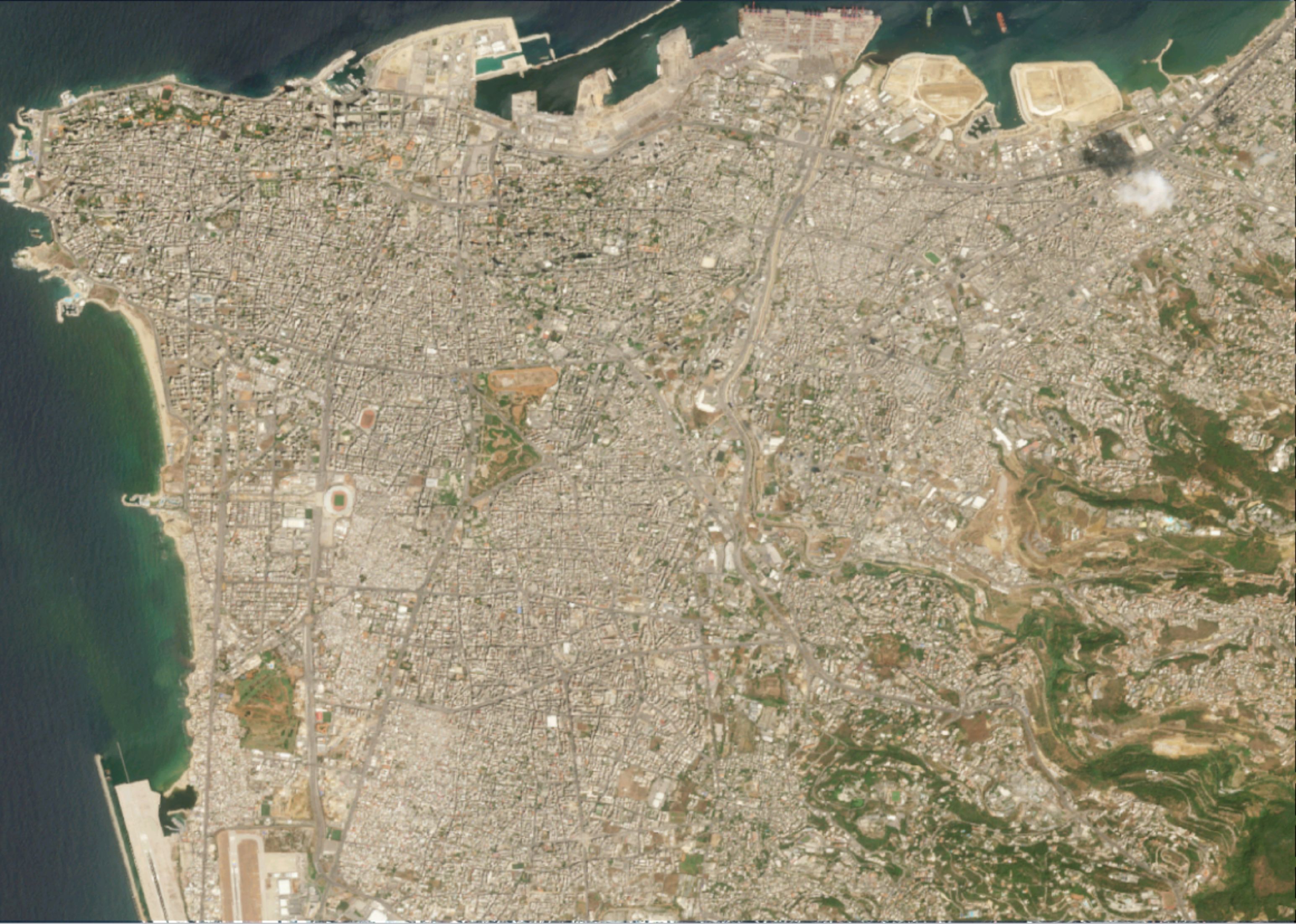} 
\label{fig:post_s2} 
} 
\hspace{5mm}
\subfloat[Beirut Post Planet Dove]{
\includegraphics[width=.4\linewidth]{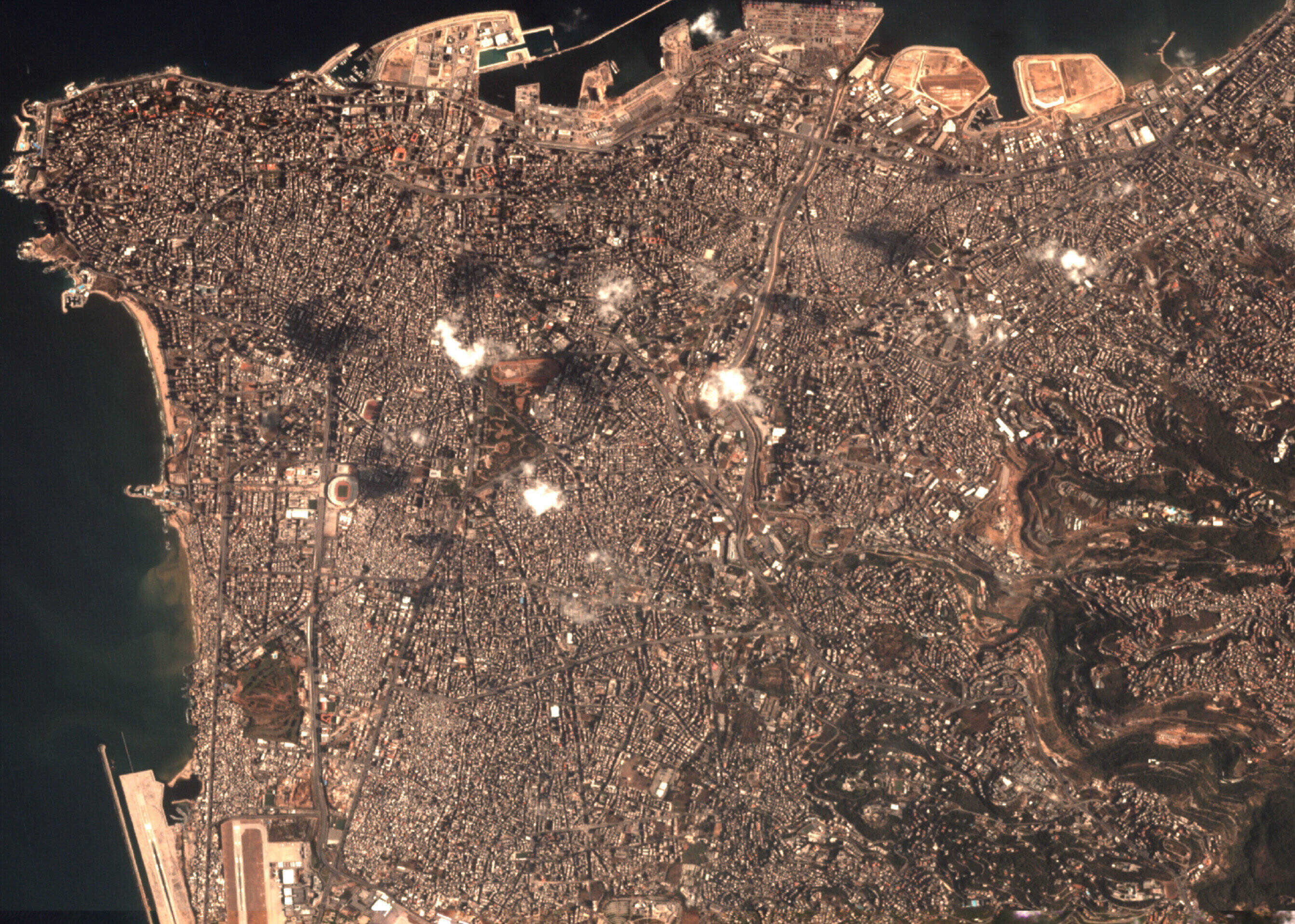} 
\label{fig:post_planet} 
} 
\caption{Surface reflectance imagery from Beirut, Lebanon, of low and high spatial resolutions, captured at similar time intervals, from Sentinel-II (10m) and Planet Dove (3m) sensors. Figures \ref{fig:pre_s2} and \ref{fig:post_s2} are the pre- and post-event images from Sentinel-II, respectively. Figures \ref{fig:pre_planet} and \ref{fig:post_planet} are the pre- and post-event images from Planet Dove, respectively.}
\label{fig:beirut_all} 
\end{figure*}

\section{Proposed Method} \label{sec:pmet}

Our method addresses the challenging task of large-scale, patch-consistent optical-to-optical image translation from a lower to a higher spatial resolution sensor domain. While spatial resolution is the most noticeable difference, other disparities such as radiometric fidelity, illumination conditions, and sensor spectral sensitivity also increase the radiometric discrepancy between the two domains \cite{CuiColor2020}.

To tackle these challenges, we leverage Denoising Probabilistic Diffusion Models (DDPMs) to not only enhance spatial resolution but also align subtle yet significant differences between the two domains, ensuring improved radiometric fidelity and consistency. This requires a training dataset with pairs of co-registered low spatial resolution (LR) domain $A$ and high spatial resolution (HR) domain $B$ images, capturing the same scenes with different sensors for proper domain adaptation.

The method hereby proposed is built upon Denoising Diffusion Implicit Models (DDIM) \cite{Song2020}, which are advantageous due to their reduced number of model forward passes during inference and their deterministic behavior in intermediate steps.
However, our approach differs from the standard classifier-free guided DDIM framework commonly used in traditional computer vision tasks \cite{ho2022classifierfree, Saharia2021, Dhariwal2021, Song2020, Nichol2021, Wang2022}. The standard framework struggles with the high variability in patch coloration within large training sets, leading to inaccurate colorization in generated patches. 
This color shifting behavior in diffusion models has been originally described by Deck and Bischoff \cite{deck2023easing}. They observed that diffusion models can exhibit color shifts that are worsened with larger input patch sizes.

Additionally, due to the stochastic nature of denoising diffusion models, different initial noise matrices can result in varying outputs for the same input \cite{Ho2020}. This randomness can cause inconsistencies in feature generation across neighboring patches, which is problematic for image-to-image translation in remote sensing.

The proposed method addresses these deficiencies by implementing novel forward and reverse diffusion procedures, ensuring high-quality, consistent results.

The model architecture and the associated hyperparameters used for our experiments are discussed in Appendix \ref{sec:apx:model_s}. Details about the denoising diffusion working principles are presented in Appendix \ref{sec:apx:fund}.

\begin{figure}
\centering
\includegraphics[width=\linewidth]{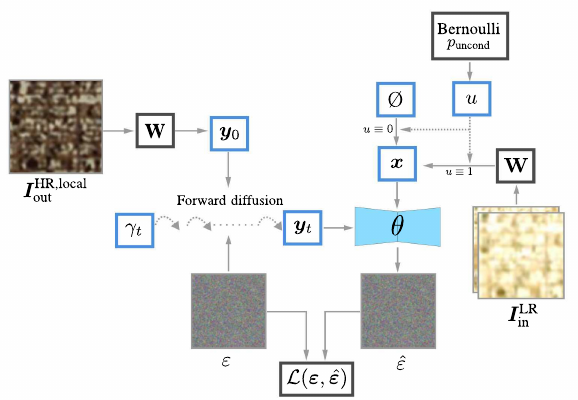}
\caption{Diagram illustrating the training step procedure for the proposed DDM-based image-to-image translation method.}
\label{fig:diag:training}
\end{figure}

\begin{algorithm}
\setstretch{1.25}
\caption{Patchwise Training Step}
\label{alg:training}
\begin{algorithmic}[1]
\REQUIRE  $\bm{I}_{\text{in}}^{\text{LR,local}}$, $\bm{I}_{\text{in}}^{\text{LR,global}}$, $\bm{I}_{\text{out}}^{\text{HR}}$, $\bm{\gamma}$, $p_{\text{uncond}}$, $\bm{\theta}_{\tau}$ \COMMENT{The model at training step $\tau$}
\ENSURE $\bm{\theta}_{\tau+1}$
\STATE $\bm{I}_{\text{in}}^{\text{LR}} = \left[
\bm{I}_{\text{in}}^{\text{LR,local}} \: ;  \bm{I}_{\text{in}}^{\text{LR,global}}
\right] $ \label{alg:training:I_lr_def}
    \STATE $u \thicksim \text{Bernoulli}(p_{\text{uncond}})$ \label{alg:training:u}
    \IF{$u \equiv 1$}
        \STATE $\bm{x} = {\O}_{2n_{\text{ch}} \times h \times w}$ \label{alg:training:uncondx}
    \ELSE
        \STATE $\{\bm{x}, \_, \_, \_\} = \mathbf{W}\left(\bm{I}_{\text{in}}^{\text{LR}}, 2n_{\text{ch}}, h, w\right)$ \label{alg:training:x_def}
    \ENDIF
    \STATE $\{\bm{y}_0, \_, \_, \_\} = \mathbf{W}
     \left(\bm{I}_{\text{out}}^{\text{HR}}, n_{\text{ch}}, h, w\right)$ \label{alg:training:y0_def}
    \STATE $\bm{\varepsilon} = \left(\varepsilon_{i,j,k}\right)_{1 \leq i \leq n_{\text{ch}},\: 1 \leq j \leq h,\: 1 \leq k \leq w}$\\
    $\quad \varepsilon_{i,j,k} \sim \mathcal{N}(0, 1) \: \forall \: i,j,k$ \label{alg:training:epsgen}
    \STATE $\gamma_t \sim \text{Uniform}(\bm{\gamma})$
    \STATE $\bm{y}_t \gets \sqrt{\gamma_t} \bm{y}_0 + \sqrt{1-\gamma_t} \bm{\varepsilon}$ \COMMENT{Forward diffusion}
    \STATE $\hat{\bm{\varepsilon}} \gets \bm{\theta}_{\tau}\left(\bm{y}_t, \bm{x}\right)$
    \STATE $\bm{\theta}_{\tau+1} \gets$ Optimization step on 
            $\nabla_{\bm{\theta}_{\tau}} \: \mathcal{L}({\bm{\varepsilon}}, \hat{\bm{\varepsilon}})$ \label{alg:training:optstep}
\end{algorithmic}
\end{algorithm}

\begin{algorithm}
\setstretch{1.25}
\caption{Image Whitening $\mathbf{W}$}
\label{alg:whitening}
\begin{algorithmic}[1]
\REQUIRE $\bm{I}_{\text{c}}$, $n_{\text{ch}}$, $h$, $w$
\ENSURE $\bm{I}_{\text{w}}$, $\bm{m}_1, m_2, m_3$
\STATE $\bm{I}_{\text{w}} \gets \bm{0}_{n_\text{ch} \times w \times h}$
\STATE $\bm{m}_1 \gets \bm{0}_{n_\text{ch}}$
\STATE $m_2 \gets 0$
\STATE $m_3 \gets 0$
\FOR{$i \gets 0 \text{ to } n_{\text{ch}} - 1$}
    \STATE ${m}_1^i \gets \mu(\bm{I}_{\text{c}}^i)$
    \STATE $\bm{I}_{\text{w}}^i \gets \bm{I}_{\text{c}}^i - {m}_1^i \cdot \mathbf{J}_{h \times w}$ \COMMENT{$\mathbf{J}_{h \times w}$ is the all-ones $h \times w$ matrix}
\ENDFOR
\STATE $m_2 \gets \min \bm{I}_{\text{w}}$
\STATE $\bm{I}_{\text{w}} \gets \bm{I}_{\text{w}} - m_2 \cdot \mathbf{J}_{n_\text{ch} \times h \times w}$
\STATE $m_3 \gets \max \bm{I}_{\text{w}}$
\STATE $\bm{I}_{\text{w}} \gets 2\left(\bm{I}_{\text{w}}/m_3-0.5 \cdot \mathbf{J}_{n_\text{ch} \times h \times w}\right)$
\end{algorithmic}
\end{algorithm}
\subsection{Training}
\label{sec:pmet:train}
Our training procedure diverges from the conventional DDPM by incorporating modifications to the forward diffusion process through Whitening and Coloring techniques. These modifications are critical as they prevent the diffusion network from being confounded by the high variability in patch coloration found within extensive training datasets, particularly when patches have highly similar content. Without these measures, the network fails to accurately predict the channel mean values for the output patches, as shown in Section \ref{sec:experiments:imagen}. Details of the implementation of DDPM and DDIM are presented in Appendices \ref{sec:apx:dif} and \ref{sec:apx:ddim}.
\paragraph*{Algorithm Description}
Algorithm \ref{alg:training} defines the training step for a single pair of low spatial resolution patch from domain $A$, $\bm{I}_{\text{in}}^{\text{LR}} \in \mathds{R}^{2n_{\text{ch}} \times h \times w}$, and high resolution patch, $\bm{I}_{\text{out}}^{\text{HR}} \in \mathds{R}^{n_{\text{ch}} \times h \times w}$, from domain $B$. The goal is to train a model $\bm{\theta}$ that is able to convert $\bm{I}_{\text{in}}^{\text{LR, local}}$ into $\bm{I}_{\text{out}}^{\text{HR}}$. The training step is illustrated by Figure \ref{fig:diag:training}.
The input and output images have the same pixel resolution but differ in spatial resolution, i.e., $\bm{I}_{\text{in}}^{\text{LR, local}}$ is up-scaled with cubic interpolation to have the pixel resolution of $\bm{I}_{\text{out}}^{\text{HR}}$, while covering the exact same area and the similar features. Thus, patches aligned in time are required for training to work.
Line \ref{alg:training:I_lr_def} of the algorithm defines $\bm{I}_{\text{in}}^{\text{LR}}$ as the channel-wise stacking of $\bm{I}_{\text{in}}^{\text{LR,local}}$ and $\bm{I}_{\text{in}}^{\text{LR, global}}$. The matrix $\bm{I}_{\text{in}}^{\text{LR,local}}$ is defined as the region of interest (ROI) of the low-resolution image from domain $A$, which is the patch to be translated to domain $B$ in inference. $\bm{I}_{\text{in}}^{\text{LR, global}}$ represents the surroundings of the region of interest, down-sampled to have the same dimensions of $\bm{I}_{\text{in}}^{\text{LR,local}}$.
$\bm{I}_{\text{in}}^{\text{LR, global}}$ is defined to contain four times the visible area of $\bm{I}_{\text{in}}^{\text{LR,local}}$, i.e., the area that is contained in $\bm{I}_{\text{in}}^{\text{LR,local}}$ plus three additional equally sized regions in its vicinity, thus depicting north, east and southeast patches that are connected to $\bm{I}_{\text{in}}^{\text{LR,local}}$. 
\color{dgreen}
That is, $\bm{I}_{\text{in}}^{\text{LR,local}}$ is the southwestern patch in a 2x2 patch grid. This grid is downsampled to the pixel dimensions of $\bm{I}_{\text{in}}^{\text{LR,local}}$ to form $\bm{I}_{\text{in}}^{\text{LR, global}}$.
\color{black}
This is done to give more context to the model of what exactly lies in the vicinity the input patch.
Line \ref{alg:training:u} samples a random variable $u$ from a Bernoulli distribution of probability $p_{\text{uncond}}$. If it turns out to be $1$, the model behaves as an unconditional diffusion model for the current training step (Line \ref{alg:training:uncondx}), as in \cite{ho2022classifierfree}. Otherwise, it behaves as a conditional diffusion model. ${\O}$ is the null-condition matrix (like in \cite{ho2022classifierfree}). To ensure that ${\O}$ is correctly interpreted by the model, we set it as a matrix whose all elements are equal to $-2$, which is outside the range of values of input pixels $[-1, 1]$. Thus, the model does not interpret the null condition as a regular input image and vice-versa.
Lines \ref{alg:training:x_def} and \ref{alg:training:y0_def} define the conditioning input from domain $A$, $\bm{x} \in \mathds{R}^{2n_{\text{ch}} \times h \times w}$, and the domain $B$ ground truth $\bm{y}_0 \in \mathds{R}^{n_{\text{ch}} \times h \times w}$, respectively. They are obtained by applying the function $\mathbf{W}$, defined in Algorithm \ref{alg:whitening}, to $\bm{I}_{\text{in}}^{\text{LR}}$ and $\bm{I}_{\text{out}}^{\text{HR}}$. This function extracts out the overall color information from an input image, i.e., it zeroes the mean of each channel. A normalization between $[-1, 1]$ follows. $\mathbf{W}$ also returns the previous means, minimum, and maximum values extracted from the input, which can be used for reverting the whitening process during inference. However, during training, these values are not used. Lines \ref{alg:training:epsgen} to \ref{alg:training:optstep} of Algorithm \ref{alg:training} are as in the original conditional DDPM training procedure \cite{Saharia2021}. This process is repeated for all existing training patches until convergence.

\begin{figure}
\centering
\includegraphics[width=\linewidth]{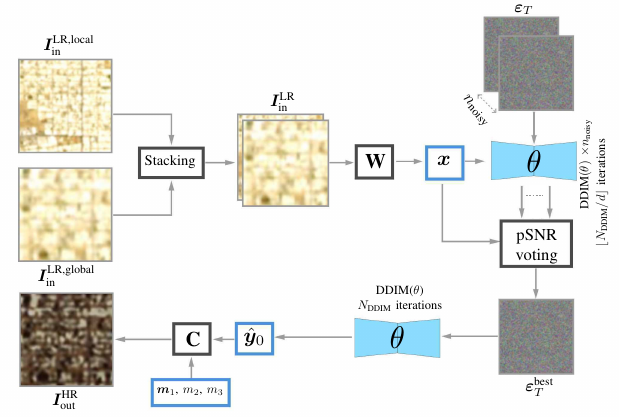}
\caption{Diagram illustrating the inference procedure for the proposed DDM-based image-to-image translation method. Here, the color information variables $\bm{m}_1$, $m_2$, $m_3$ are provided from an external image, e.g., a post-event Planet Dove image.}
\label{fig:diag:inference}
\end{figure}
\begin{algorithm}
\setstretch{1.25}
    \caption{Patchwise Inference}
    \label{alg:inference}
    \begin{algorithmic}[1]
        \REQUIRE $\bm{\theta}, \bm{I}_{\text{in}}^{\text{LR,local}}$, $\bm{I}_{\text{in}}^{\text{LR,global}}$, $n_{\text{ch}}$, $h$, $w$, $\omega_{\text{uncond}}$, $N_{\text{DDIM}}$, $d$, $n_{\text{noisy}}$ \\ \textbf{Optional}: $\bm{m}_1$, $m_2$, $m_3$ \COMMENT{Color information variables.}
        \ENSURE $\bm{I}_{\text{out}}^{\text{HR,local}}$
        \STATE $\bm{I}_{\text{in}}^{\text{LR}} = \left[
        \bm{I}_{\text{in}}^{\text{LR,local}} \:  \bm{I}_{\text{in}}^{\text{LR,global}}
        \right]^T $ \label{alg:inference:I_lr_def}
        \IF{$\bm{m}_1$, $m_2$, $m_3$ are all provided}
            \STATE $\{\bm{x}_{\text{local}}, \_, \_, \_\} = \mathbf{W}\left(\bm{I}_{\text{in}}^{\text{LR,local}}, n_{\text{ch}}, w, h\right)$
            \STATE $\{\bm{x}_{\text{global}}, \_, \_, \_\} = \mathbf{W}\left(\bm{I}_{\text{in}}^{\text{LR,global}}, n_{\text{ch}}, w, h\right)$
        \ELSE
            \STATE $\{\bm{x}_{\text{local}}, \bm{m}_1, m_2, m_3\} = \mathbf{W}\left(\bm{I}_{\text{in}}^{\text{LR,local}}, n_{\text{ch}}, w, h\right)$
            \STATE $\{\bm{x}_{\text{global}}, \_, \_, \_\} = \mathbf{W}\left(\bm{I}_{\text{in}}^{\text{LR,global}}, n_{\text{ch}}, w, h\right)$
        \ENDIF \label{alg:inference:endif}
        \STATE $\bm{x} = \left[ \bm{x}_{\text{local}}; \: \bm{x}_{\text{global}} \right]$
        \STATE $\bm{\mathcal{E}}_T \gets \left(\varepsilon^{i,j,k,l}_T\right)_{1 \leq i \leq n_{\text{noisy}},\: 1 \leq j \leq n_{\text{ch}},\: 1 \leq k \leq h,\: 1 \leq l \leq w}$\\
        $\quad \varepsilon^{i,j,k, l}_T \sim \mathcal{N}(0, 1) \: \forall \: i,j,k,l$ \label{alg:inference:E_T}
        \STATE $\hat{\bm{Y}}_{0,\text{lowqual}} \gets \mathbf{0}_{n_{\text{noisy}} \times n_{\text{ch}} \times h \times w}$
        \FOR{$i \gets 1 \text{ to } n_{\text{noisy}}$} 
            \STATE $\hat{\bm{Y}}_{0,\text{lowqual}}^i \gets \text{DDIM}\left(\bm{\theta}, \bm{\mathcal{E}}_T^i, \omega_{\text{uncond}}, \bm{x}, N_\text{pre} \right)$ \label{alg:inference:Y_0_hat_lq}
        \ENDFOR
        \STATE $\bm{\mathcal{E}}_T^\text{best} = \arg\max_{\bm{\mathcal{E}}_T^j} 
        \text{PSNR}\left(\bm{x}_{\text{local}}, \hat{\bm{Y}}_{0,\text{lowqual}}^i\right)$ \label{alg:inference:eps_t_best} 
        \STATE $\hat{\bm{y}}_{0} \gets \text{DDIM}\left(\bm{\theta}, \bm{\mathcal{E}}_T^{\text{best}}, \omega_{\text{uncond}}, \bm{x}, N_\text{final}\right)$ \label{alg:inference:y_0_hat}
        \STATE $\bm{I}_{\text{out}}^{\text{HR,local}} = \mathbf{C}(\hat{\bm{y}}_{0}, n_{\text{ch}}, h, w, \bm{m}_1, m_2, m_3)$ \label{alg:inference:I_out_hr}
    \end{algorithmic}
\end{algorithm}

\subsection{Inference}
\label{sec:pmet:inf}
The inference procedure illustrated by Figure 3 aims to maximize the image generation quality by selecting the best possible initial noise matrix to be fed to the reverse diffusion process. We propose a novel reverse diffusion procedure by integrating the same Whitening and Coloring techniques used during training. To further enhance the fidelity of the generated patches to their inputs, we refine the selection process for the initial noise condition input for the reverse diffusion process, ensuring optimal consistency and translation quality. Implementation details for the DDIM inference process are presented in Appendix \ref{sec:apx:ddim}.

In our experiments, we observed that, for the same conditioning input $\bm{x}$, the quality of DDIM reverse diffusion outputs generated with different initial noise matrices varied significantly. That is, for a combination of independent and identically distributed random initial noise matrices, fixing $\bm{x}$, the DDIM, in some cases, generates samples entirely consistent with $\bm{x}$, whereas sometimes it generates inconsistent samples with undesired hallucinations. Since the DDIM reverse diffusion is a deterministic process, i.e., given fixed $\bm{\varepsilon}_T$ and $\bm{x}$, its output remains the same, the cause for varying consistency levels lies only in the different sampled initial noise matrices, which are the source of the randomness of the whole inference process. Thus, we propose a procedure that seeks to select an initial noise matrix such that the DDIM output possesses the highest consistency compared to $\bm{x}$. This procedure consists of repeating the DDIM reverse diffusion process with a reduced number of iterations, using as input $n_{\text{noisy}}$ different initial noise matrices. The resulting outputs are tested for consistency by individually calculating the peak signal-to-noise ratio (PSNR) between them and $\bm{x}_{\text{local}}$. The one that was responsible for the maximum PSNR is used as the definitive starting noise matrix for the inference process, which takes place with a higher number of iterations to ensure maximal generation quality.

\paragraph*{Algorithm Description}
Algorithm \ref{alg:inference} defines the inference process for a single low spatial resolution domain $A$ patch $\bm{I}_{\text{in}}^{\text{LR,local}}$, which results in a translated high spatial resolution patch $\bm{I}_{\text{out}}^{\text{HR}}$ resembling those of domain $B$, containing the same elements of $\bm{I}_{\text{in}}^{\text{LR, local}}$. As in Algorithm \ref{alg:training}, the input and output images have the same pixel resolution but differ in spatial resolution, meaning that the lower-resolution input is up-sampled to the output's pixel dimensions.
In the first line of the algorithm, we define the conditioning input $\bm{I}_{\text{in}}^{\text{LR}}$ as the concatenation of the domain $A$ patch to be translated $\bm{I}_{\text{in}}^{\text{LR,local}}$ and its down-sampled surroundings $\bm{I}_{\text{in}}^{\text{LR,global}}$, as done in training. 
Lines \ref{alg:inference:I_lr_def} to \ref{alg:inference:endif} prepare the low spatial resolution domain $A$ input to be in the format required by the trained neural network $\bm{\theta}$. The color information-related variables $\bm{m}_1$, $m_2$, $m_3$ are assigned, which are extracted during the whitening process of the local domain $A$ patch if the colors from the input patch are desired to be reassigned to the output image from domain $B$. Otherwise, color information from an external image, e.g., a post-event domain $B$ image, can be used.
Following, line \ref{alg:inference:E_T} defines $\bm{\mathcal{E}}_T$ as a Gaussian noise matrix that can be interpreted as $n_{\text{noisy}}$ individual matrices stacked together in a new dimension. Each of these is used as initial noise matrices for the DDIM reverse diffusion process \cite{Song2020} described in line \ref{alg:inference:Y_0_hat_lq}. Thus, DDIM inference is executed $n_\text{noisy}$ times using the trained model $\bm{\theta}$, with different initial noise matrices $\bm{\mathcal{E}}_T^i$, but with a small number of iterations $N_{\text{pre}}$. It performs classifier-free guidance inference using $\omega_{\text{uncond}}$ as the weighting parameter, which controls how much the unconditional score estimate is weighted relative to the conditional score estimate. 
Each of the $n_\text{noisy}$ inferences generates a prediction $\hat{\bm{Y}}_{0,\text{lowqual}}^i$, that, due to the lower number of DDIM iterations, are coarse predictions, whose image quality is nevertheless good enough for the following step: they are used to evaluate which of the $n_\text{noisy}$ noise matrices contained in $\bm{\mathcal{E}}_T$ created the prediction of highest quality with respect to the peak signal-to-noise-ratio (PSNR) metric, comparing it to the first $n_{\text{ch}}$ channels of the conditioning input $\bm{x}$, which refer to $\bm{I}_{\text{in}}^{\text{LR,local}}$. The chosen noise matrix is denoted as $\bm{\mathcal{E}}_T^\text{best}$, and is used to generate the final predicted output of the DDIM reverse diffusion $\hat{\bm{y}}_0$, now using $N_{\text{final}} > N_{\text{pre}}$ iterations. Finally, the high spatial resolution output patch $\bm{I}_{\text{out}}^{\text{HR,local}}$ is obtained by applying Algorithm \ref{alg:coloring} to $\hat{\bm{y}}_0$, which gives the overall color information to the DDIM's prediction.
\begin{algorithm}
\setstretch{1.25}

\begin{algorithmic}[1]
\REQUIRE $\bm{I}_{\text{w}}$, $n_{\text{ch}}$, $h$, $w$, $\bm{m}_1$, $m_2$, $m_3$
\ENSURE $\bm{I}_{\text{c}}$
\STATE $\bm{I}_{\text{c}} \gets \bm{0}_{n_\text{ch} \times h \times w}$
\STATE $\bm{I}_{\text{w}} \gets \frac{\bm{I}_{\text{w}} - \min \bm{I}_{\text{w}}\mathbf{J}_{n_\text{ch} \times h \times w}}{\max \bm{I}_{\text{w}} - \min \bm{I}_{\text{w}}}$
\STATE $\bm{I}_{\text{w}} \gets m_3 \cdot \bm{I}_{\text{w}}$
\STATE $\bm{I}_{\text{w}} \gets \bm{I}_{\text{w}} + m_2 \cdot \mathbf{J}_{n_\text{ch} \times h \times w}$
\FOR{$i \gets 0 \text{ to } n_{\text{ch}} - 1$}
    \STATE $\bm{I}_{\text{c}}^i \gets \bm{I}_{\text{w}}^i + m_1^i \cdot \mathbf{J}_{h \times w} $
\ENDFOR
\end{algorithmic}
\caption{Image Colorization $\mathbf{C}$}
\label{alg:coloring}
\end{algorithm}


\section{Dataset}
\label{sec:dataset}
For the proposed experiments, paired images from satellites Sentinel-II and Planet Dove were gathered. The former is a widely known \mbox{open-access} optical satellite with a spatial resolution of $10\text{m}^2/\text{pixel}$, whereas the latter is a class of commercial satellites aimed at producing high-quality images at a spatial resolution of $3\text{m}^2/\text{pixel}$. 
A dataset containing 77 pairs of surface reflectance 
rasters has been assembled for training. These rasters are parts of or a combination of acquisitions that were captured at similar time periods. They are properly geolocated, georeferenced, and checked for image quality issues. In total, a vast area of $7894 \text{ km}^2$ is covered by the training set, which is comparable to the area of Puerto Rico ($9104 \text{ km}^2$).
As the test set for the image-to-image translation algorithm, 8 pairs of images were selected, which, in total, cover an area of $707 \text{ km}^2$. With these, image quality metrics are extracted for the proposed method and compared to other possible approaches for the I2I translation task.

For evaluation of the performance of our method in a change detection task, we use pre and post-images from the port region in Beirut, Lebanon. On August 4th, 2020, an explosion dramatically affected the port and surrounding areas. As they have recovered over time, many changes took place, e.g., new and repaired constructions and structures. Thus, for the change detection test, we use as pre-event images a pair of images acquired hours after the explosion, between 04 and 20 of August 2020, when the effects of the event are still evident. As post images, acquisitions from July 2023 of the exact same region are used. Figure \ref{fig:beirut_all} shows the Beirut pre and post-event images from Sentinel-II and Planet Dove. 
We also perform change detection tests in a region in Austin, USA, where many urban changes can be observed between the selected pre- and post-event images. They were captured in July 2021 and 2023, respectively.

\begin{figure*}
    \centering 
    \subfloat[Sentinel-II]{
    \includegraphics[width=.33\linewidth]{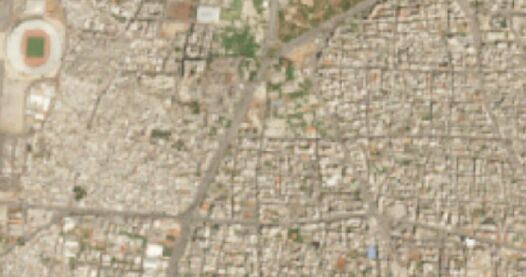}
    \label{fig:gencomp2:sub1}
    }
    \subfloat[Planet Dove (Post)]{
    \includegraphics[width=.33\linewidth]{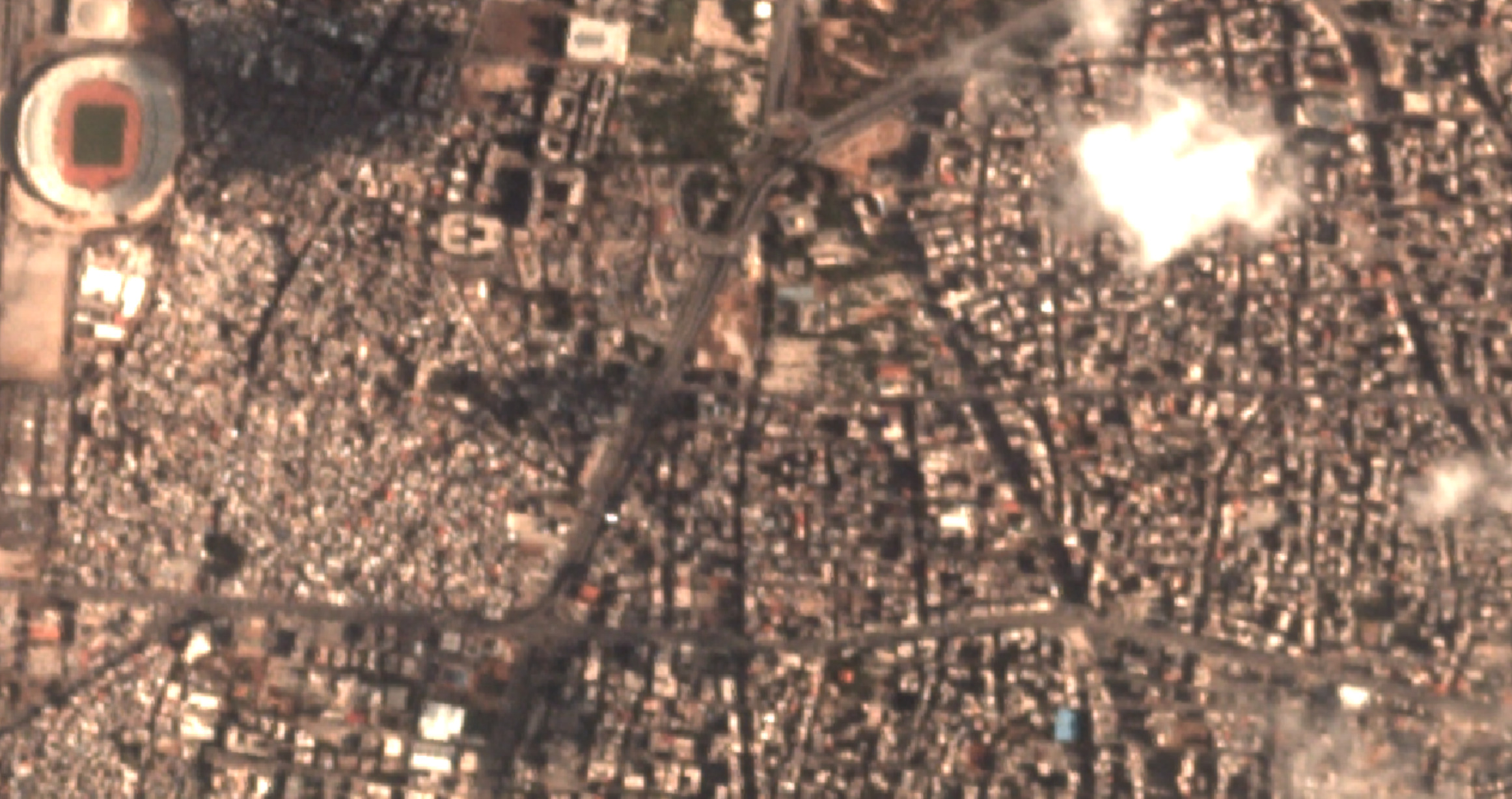}
    \label{fig:gencomp2:sub2} 
    }
    \subfloat[Proposed: DDIM + $\mathbf{W}, \mathbf{C}$ + PSNR Voting]{
    \includegraphics[width=.33\linewidth]{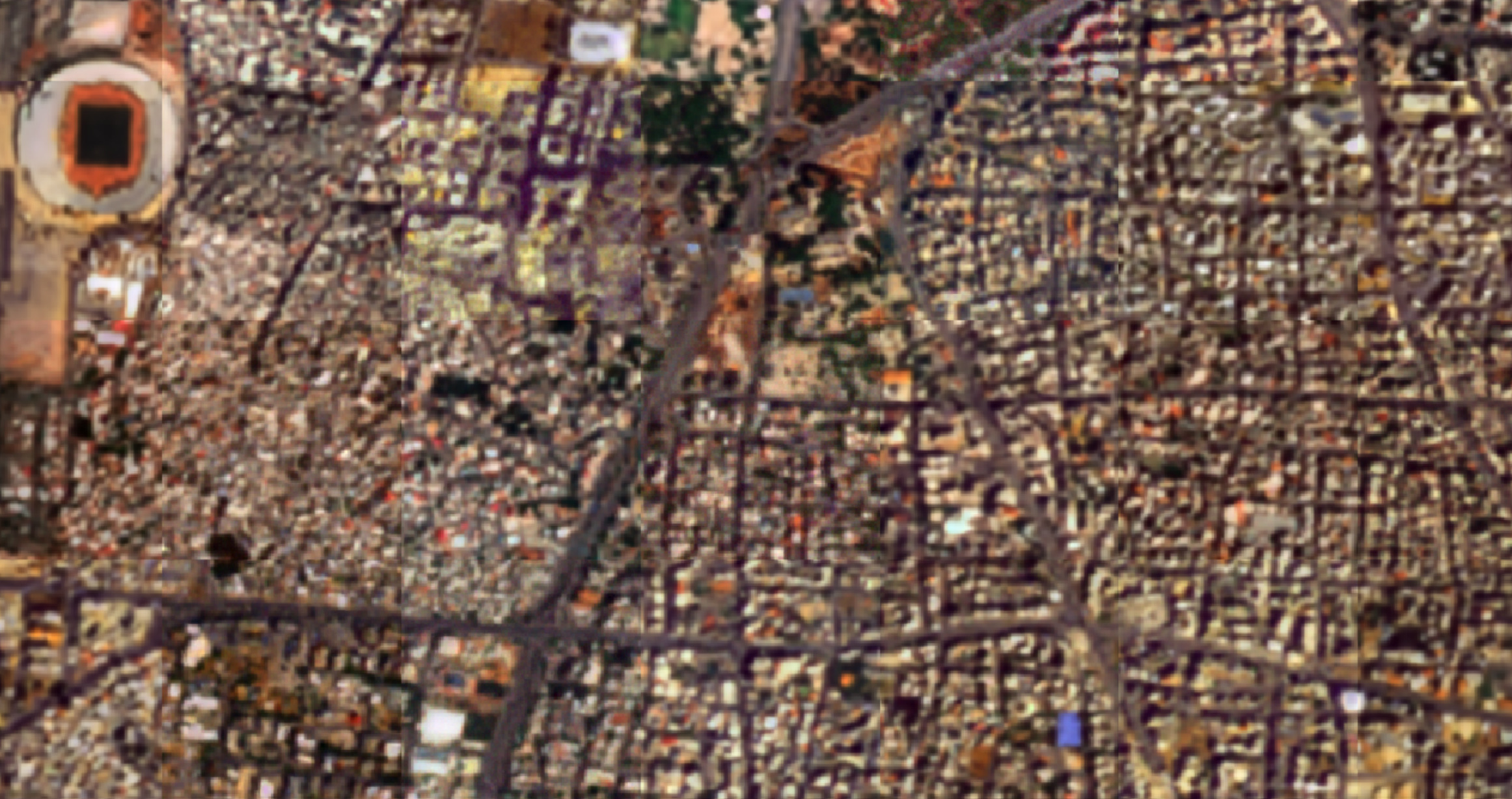}
    \label{fig:gencomp2:sub3} 
    } 
    \hfill

    \subfloat[DDIM + $\mathbf{W}, \mathbf{C}$]{
    \includegraphics[width=.33\linewidth]{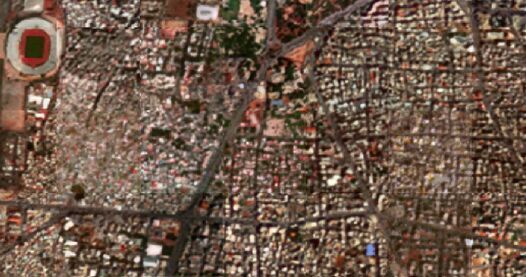} 
    \label{fig:gencomp2:sub4} 
    } 
    \subfloat[DDIM]{
    \includegraphics[width=.33\linewidth]{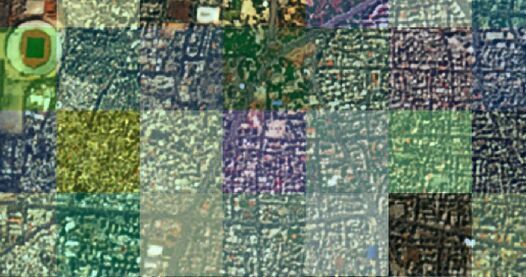}
    \label{fig:gencomp2:sub5}
    }
    \subfloat[Regression]{
    \includegraphics[width=.33\linewidth]{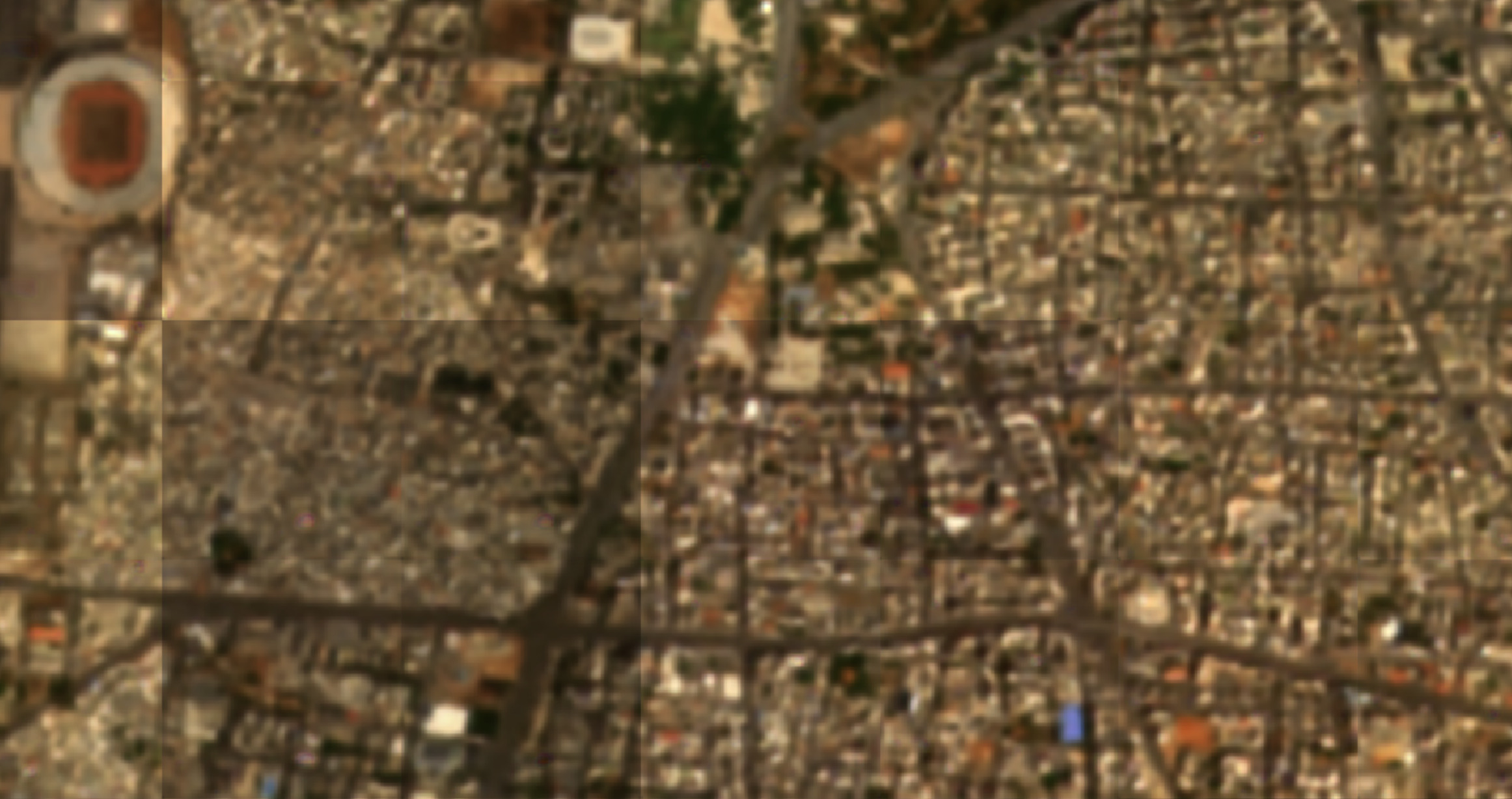} 
    \label{fig:gencomp2:sub6} 
    } 
    \hfill
    
    \subfloat[Regression + $\mathbf{W}, \mathbf{C}$]{
    \includegraphics[width=.33\linewidth]{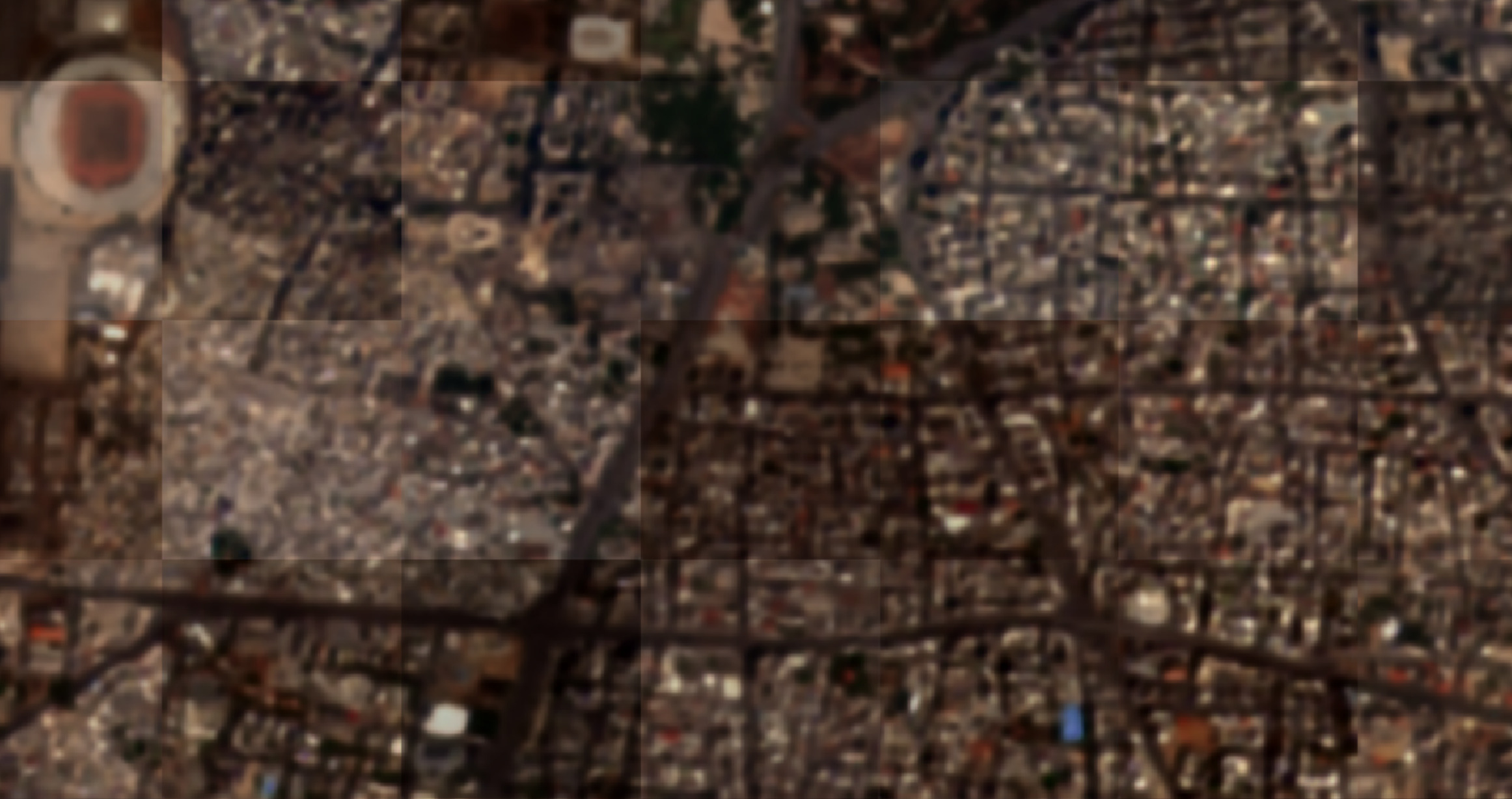}
    \label{fig:gencomp2:sub7} 
    } 
    \subfloat[Pix2Pix \cite{Pix2Pix2017}]{
    \includegraphics[width=.33\linewidth]{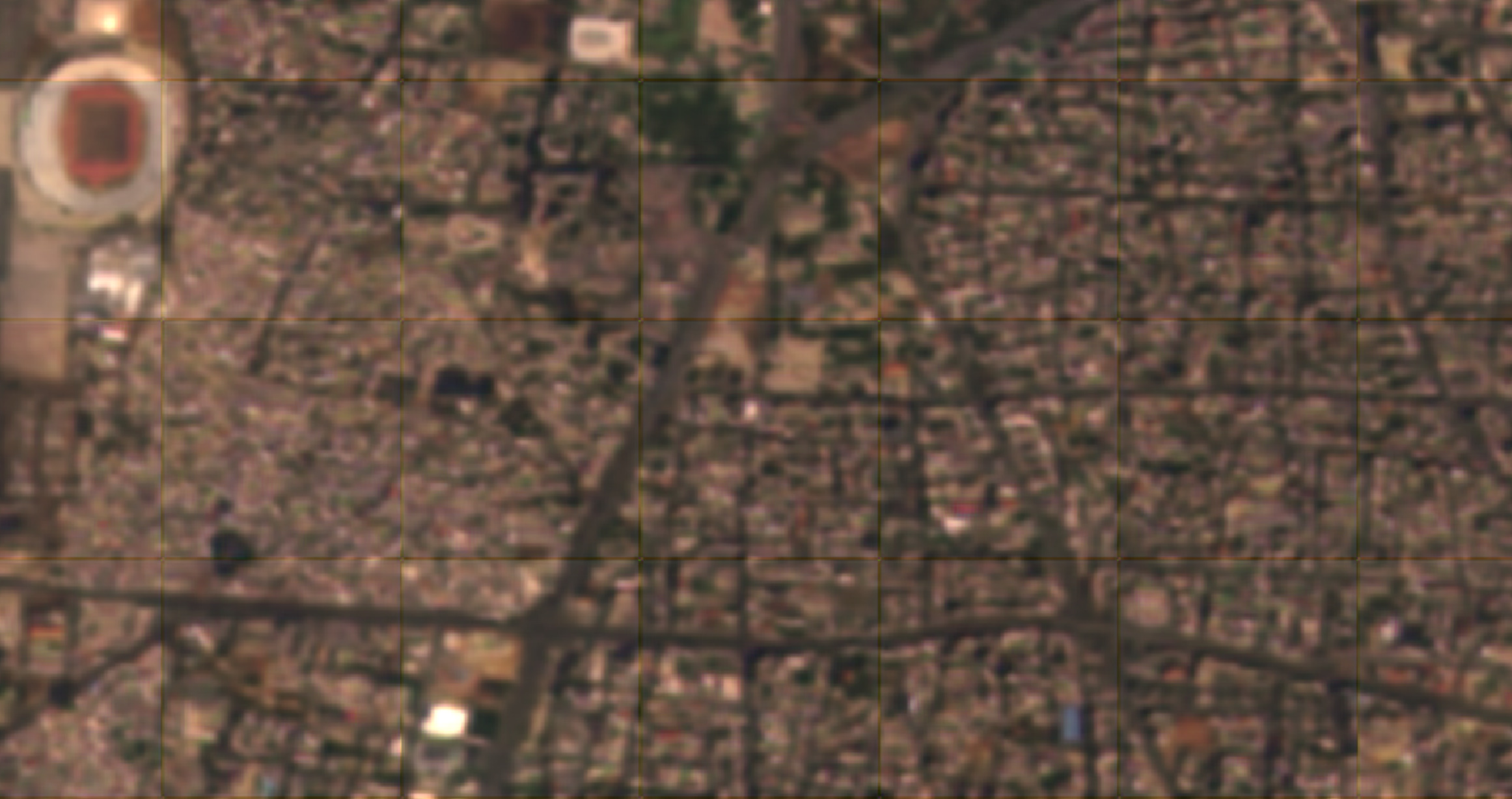}
    \label{fig:gencomp2:sub8} 
    }
    \subfloat[ShuffleMixer \cite{sun2022shufflemixer}]{
    \includegraphics[width=.33\linewidth]{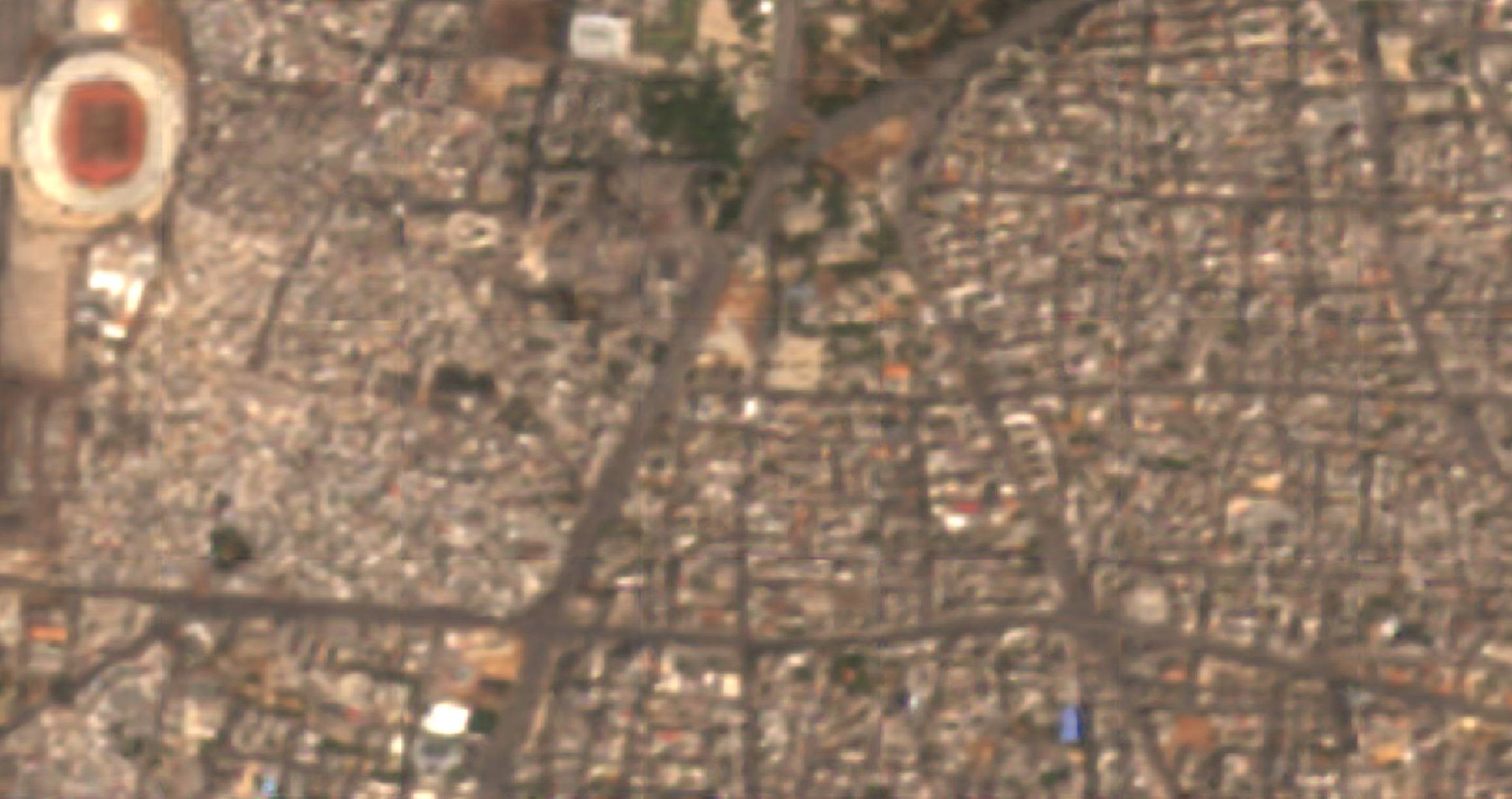}
    \label{fig:gencomp2:sub9} 
    }
    \hfill

    \subfloat[SRDenseNet \cite{zhang2018residual}]{
    \includegraphics[width=.33\linewidth]{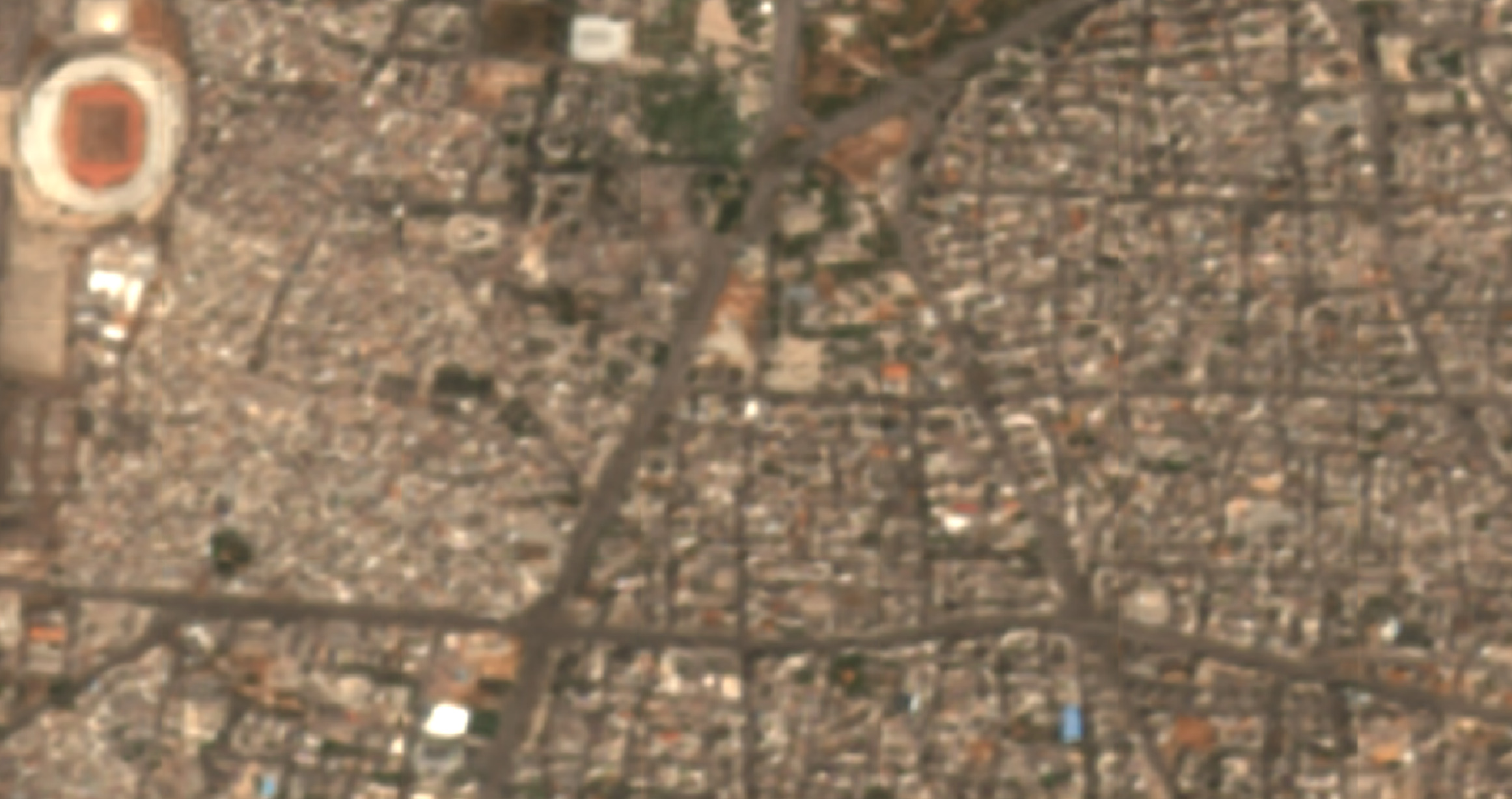}
    \label{fig:gencomp2:sub10} 
    } 
    \subfloat[SwinIR \cite{liang2021swinir}]{
    \includegraphics[width=.33\linewidth]{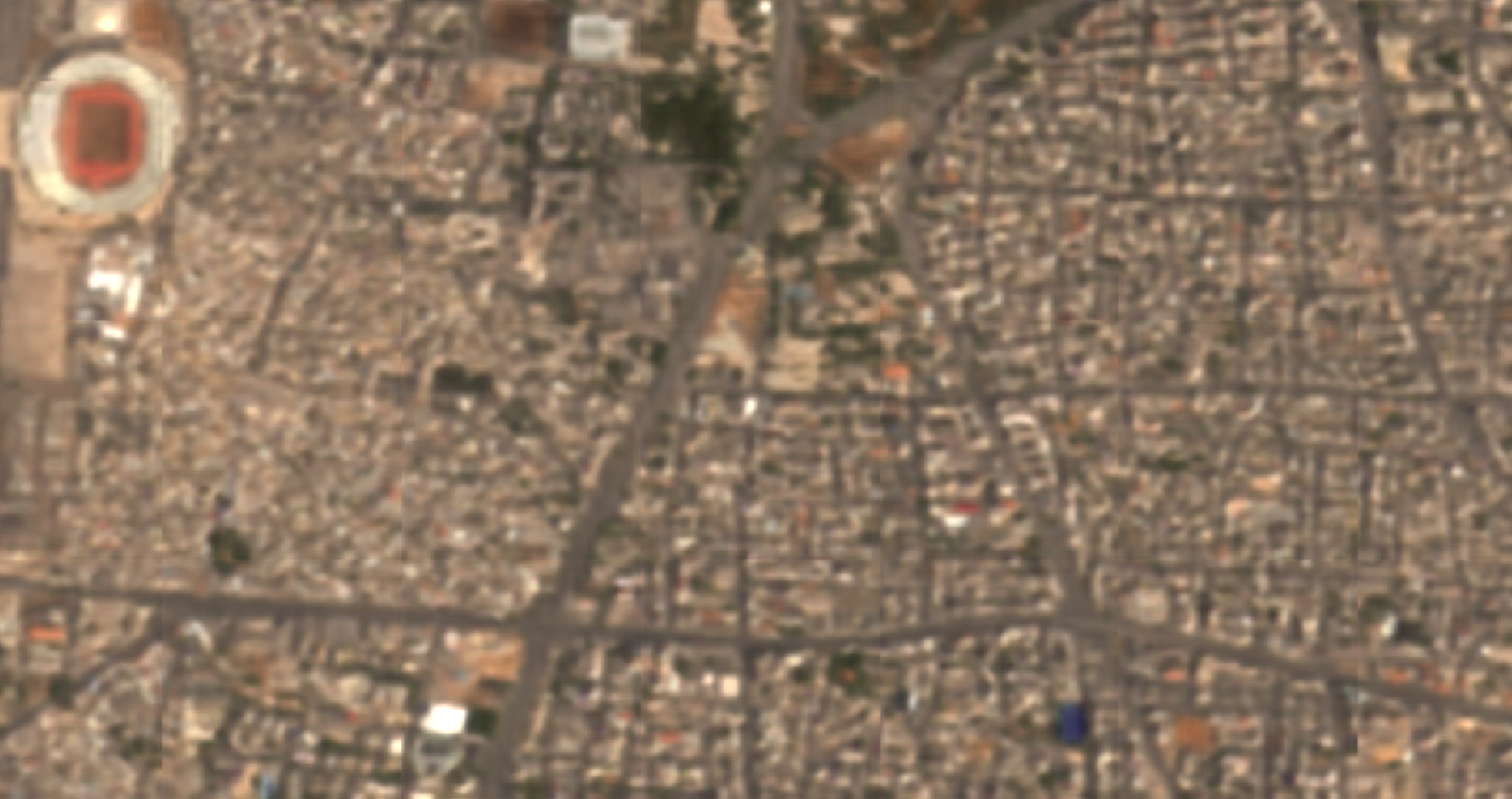}
    \label{fig:gencomp2:sub11} 
    }
    \subfloat[DDIM (SR3 Backbone) \cite{Saharia2021SR3}]{
    \includegraphics[width=.33\linewidth]{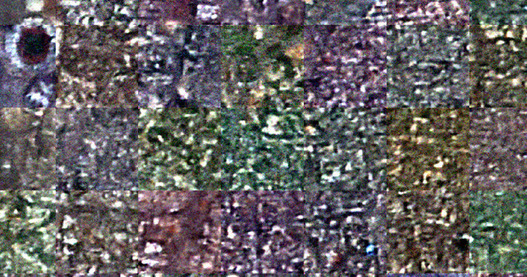}
    \label{fig:gencomp2:sub12} 
    }
    \hfill

    \caption{Comparison among images generated by the tested I2I methods (c)-(l) and the original Sentinel-II (a) and Planet Dove (b) images. The image generated with the proposed model (c) displays higher resolution features not observable among the others, while maintaining patch-wise feature consistency and avoiding blurriness.} 
    \label{fig:gencomp2}
\end{figure*}

\section{Experiments} 
\label{sec:experiments:imagen}
\subsection{Image Quality Metrics and Ablation Experiments}
\begin{table}[t]
\color{dgreen}
\begingroup
\setstretch{1.25}
\centering
\caption{Image Quality Metrics Comparison with the State-of-the-Art Using $N_{\text{final}}=64$, $N_{\text{pre}}=8$, $n_{\text{noisy}}=8$}
\label{tab:imagequal}
\begin{tabularx}{\linewidth}{Xccc}
\hline
\hline
\textbf{Method}  & \textbf{mLPIPS}$\downarrow$ & \textbf{FID}$\downarrow$ & \textbf{mPSNR}$\uparrow$  \\
\hline
Sentinel-II (Input Images) & 0.2977 & 98.08 & 12.76 \\

Regression $^{1}$ & 0.2797 & 64.12 & 16.72 \\
Regression + $\mathbf{W}, \mathbf{C}$ $^{1}$ & 0.2491 & 67.04 & \textbf{17.48}\\
Conditional DDIM $^{1}$ & 0.4769 & 87.93 & 9.919 \\
Conditional DDIM + $\mathbf{W}, \mathbf{C}$ $^{1}$& 0.1993 & 46.89 & 15.39 \\
Conditional DDIM + $\mathbf{W}, \mathbf{C}$ + \mbox{PSNR Voting} (Proposed) $^{1}$ & \textbf{0.1884} & \textbf{45.64} & 15.72 \\
\hline
\multicolumn{4}{c}{\textbf{Other State of the Art Models}} \\
\hline
Pix2Pix \cite{Pix2Pix2017} & 0.3211 & 105.0 & 14.87 \\
ShuffleMixer \cite{sun2022shufflemixer} & 0.2249 & 73.82 & 16.06 \\
SRDenseNet \cite{zhang2018residual} & 0.2296 & 57.70 & 17.12 \\
SwinIR \cite{liang2021swinir} & 0.2325 & 56.36 & 16.68 \\
\hline
\multicolumn{4}{c}{\textbf{Diffusion Using the SR3 Backbone \cite{Saharia2021SR3}}} \\
\hline
Conditional DDIM & 0.6235 & 152.6 & 9.906 \\
Conditional DDIM + $\mathbf{W}, \mathbf{C}$ + \mbox{PSNR Voting} & 0.3730 & 82.15 & 15.87\\
\hline
\hline
\end{tabularx}\\
\endgroup
\vspace{0.1cm}
\footnotesize{$^{1}$Using the SDM \cite{Wang2022} backbone.}
\end{table}

\begin{table}[t]
\begingroup
\setstretch{1.25}
\centering
\caption{Comparison Metrics for Different Inference Hyperparameters}
\label{tab:comp_nit}
\begin{tabularx}{\linewidth}{YYYYccc}
\hline
\hline
$N_{\text{final}}$ & $N_{\text{pre}}$ & $n_{\text{noisy}}$ & $N_{\text{total}}$ & \textbf{mLPIPS}$\downarrow$ & \textbf{FID}$\downarrow$ & \textbf{mPSNR}$\uparrow$  \\
\hline
16 & 2 & 2 & 20 & 0.2229 & 52.46 & 14.60  \\
32 & 4 & 4 & 48 & 0.1967 & \textbf{45.17} & 15.29\\
48 & 6 & 6 & 84 & 0.2013 & 46.49 & 15.15\\
64 & 8 & 8 & 128 & \textbf{0.1884} & 45.64 & \textbf{15.72} \\
\hline
\hline
\end{tabularx}\\
\endgroup

\end{table}

\begin{figure*}[t]
    \centering 

    \subfloat[]{
    \includegraphics[width=.29\linewidth]{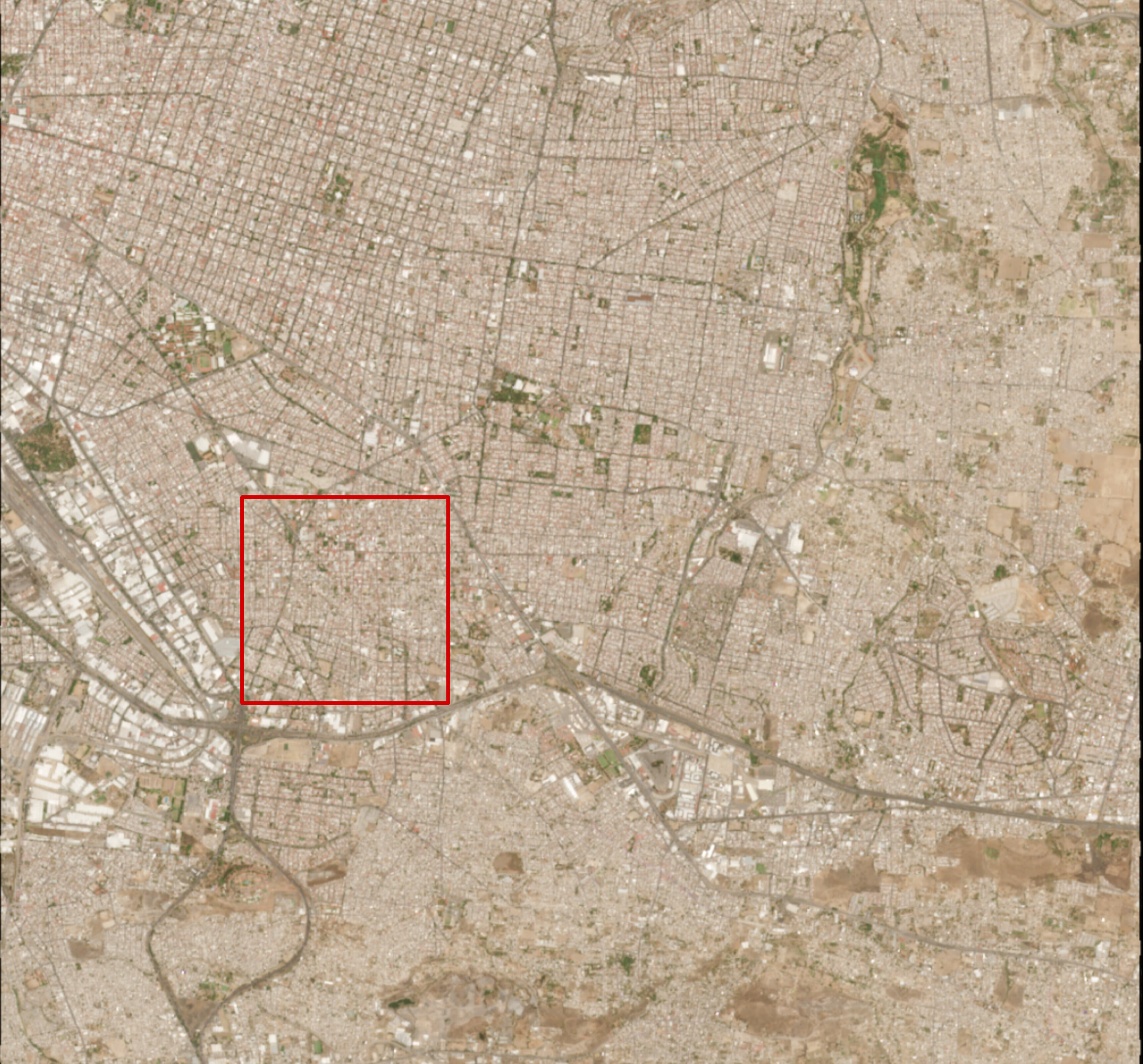} 
    \label{fig:gencomp:sub2} 
    } 
    \subfloat[]{
    \includegraphics[width=.29\linewidth]{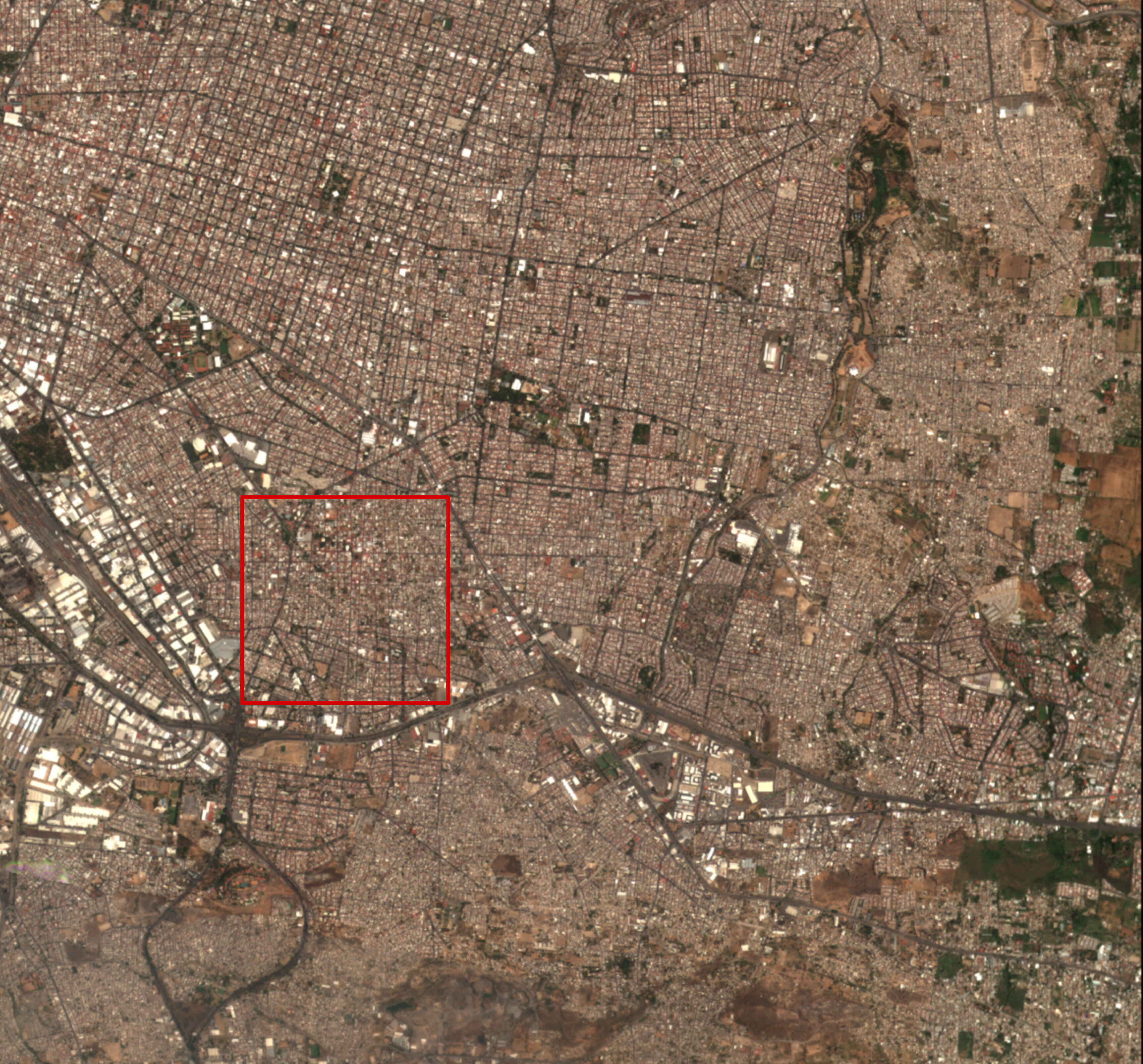}
    \label{fig:gencomp:sub1}
    }
    \subfloat[]{
    \includegraphics[width=.29\linewidth]{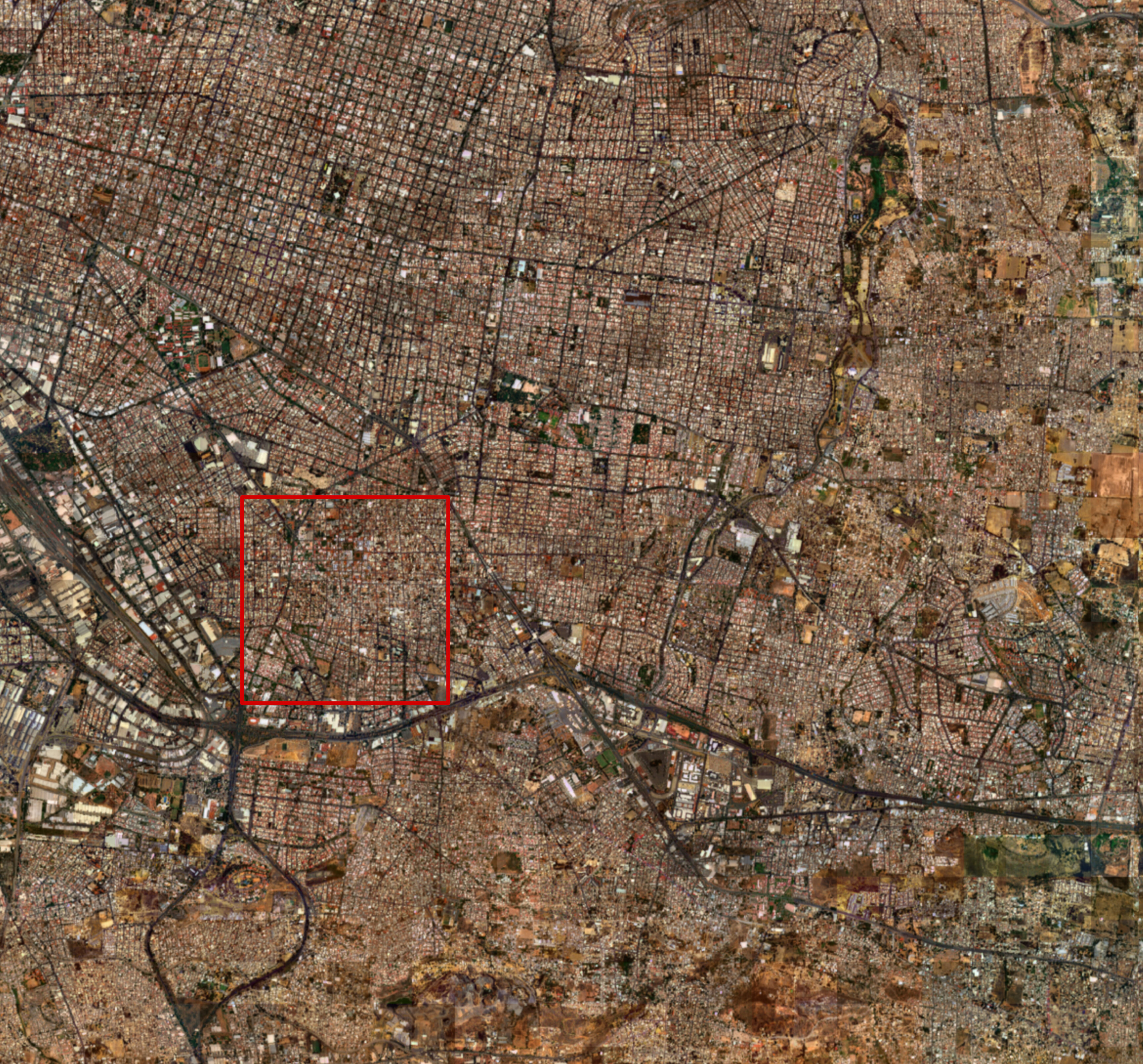}
    \label{fig:gencomp:sub3} 
    } 
    \hfill

    \subfloat[]{
    \includegraphics[width=.29\linewidth]{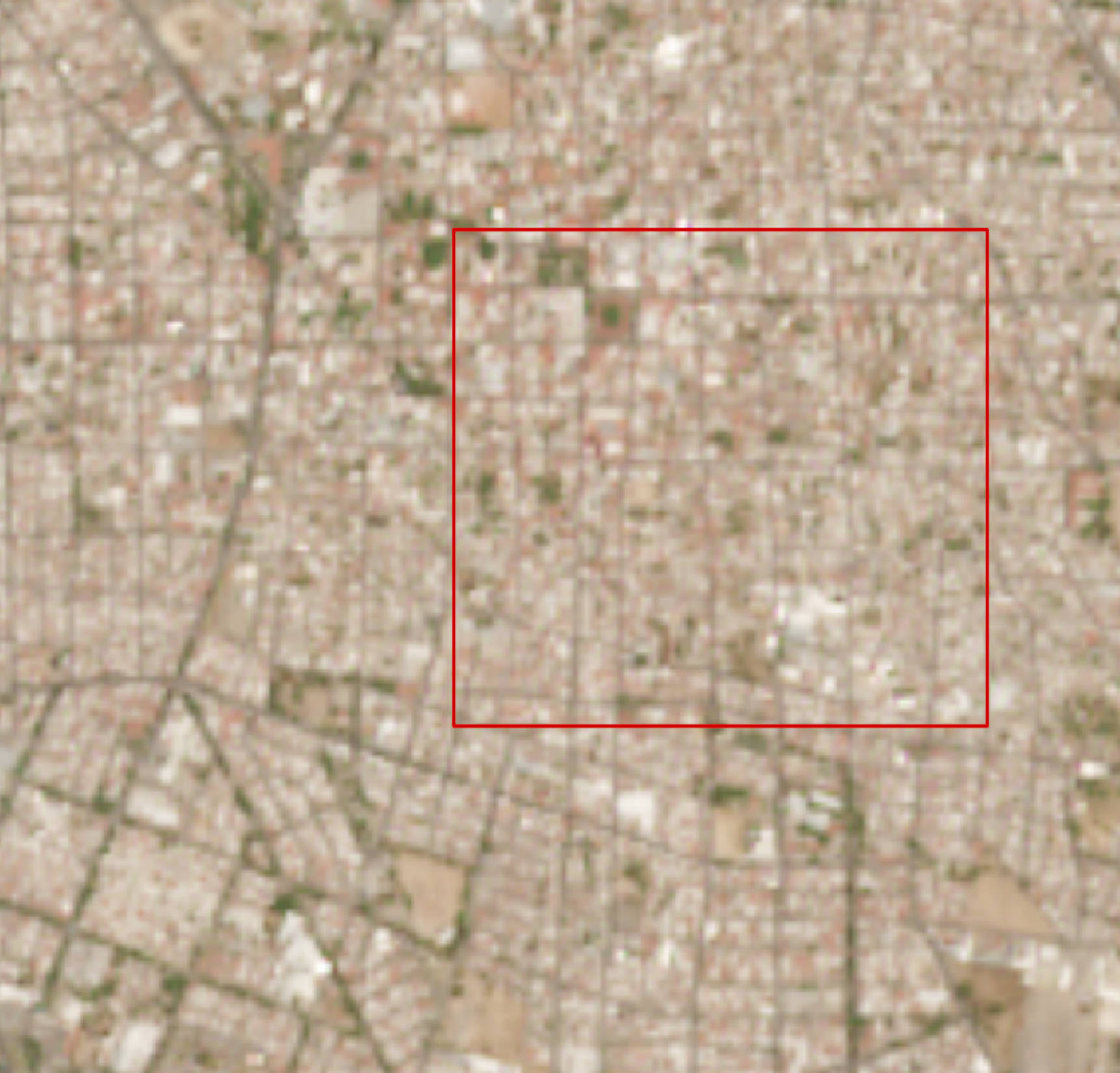} 
    \label{fig:gencomp:sub5} 
    } 
    \subfloat[]{
    \includegraphics[width=.29\linewidth]{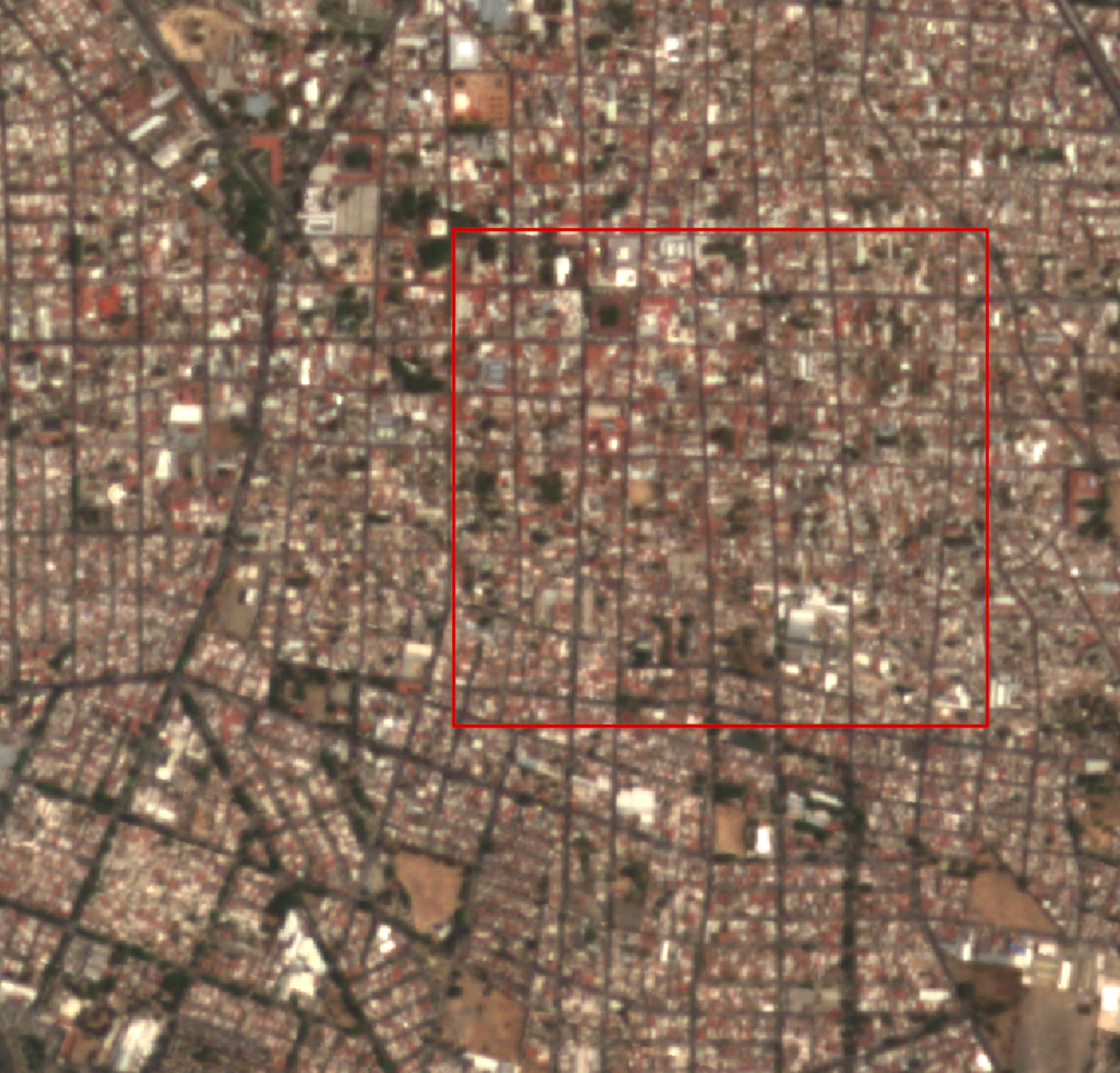}
    \label{fig:gencomp:sub4}
    }
    \subfloat[]{
    \includegraphics[width=.29\linewidth]{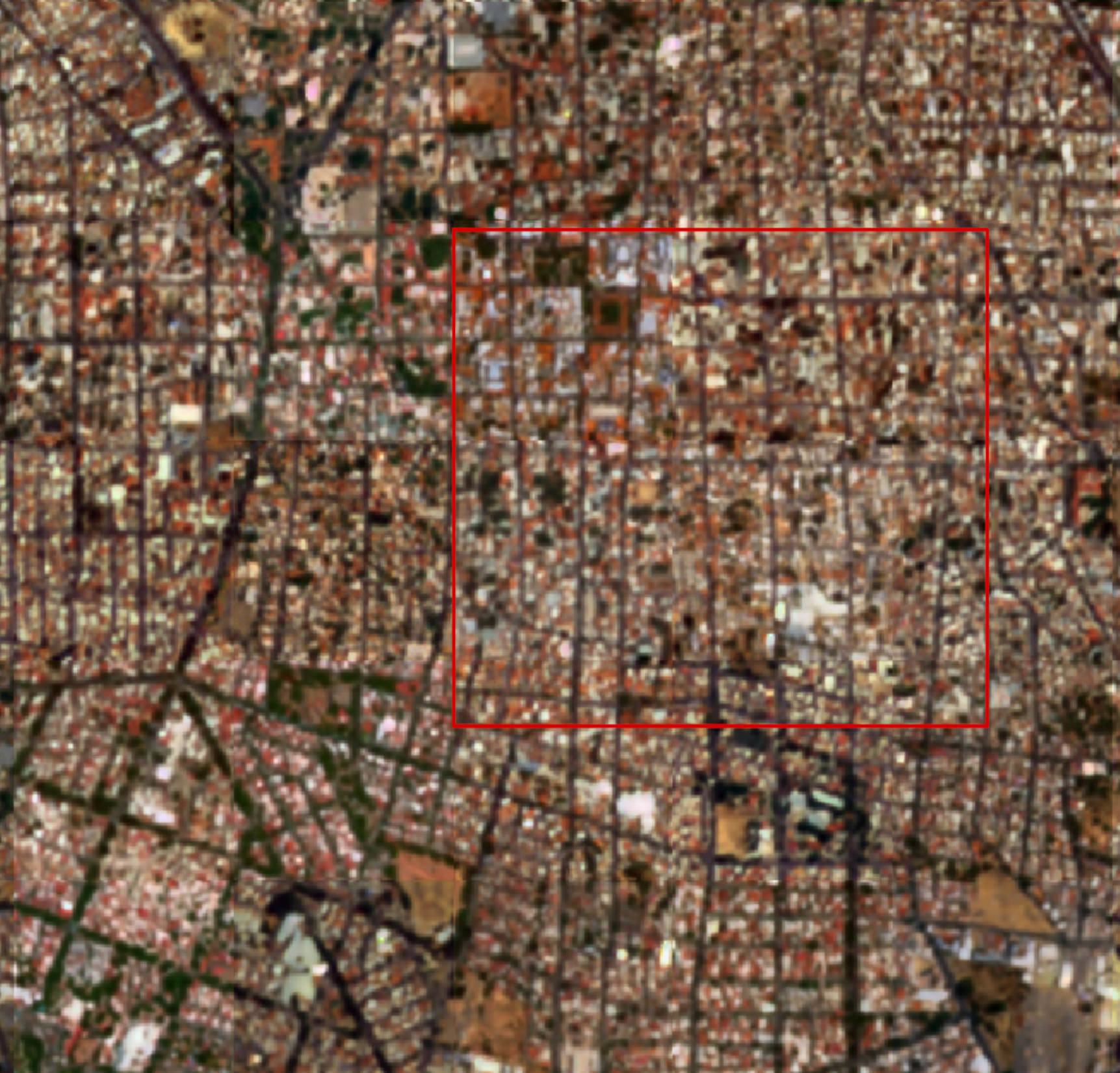}
    \label{fig:gencomp:sub6} 
    } 
    \hfill

    \subfloat[]{
    \includegraphics[width=.29\linewidth]{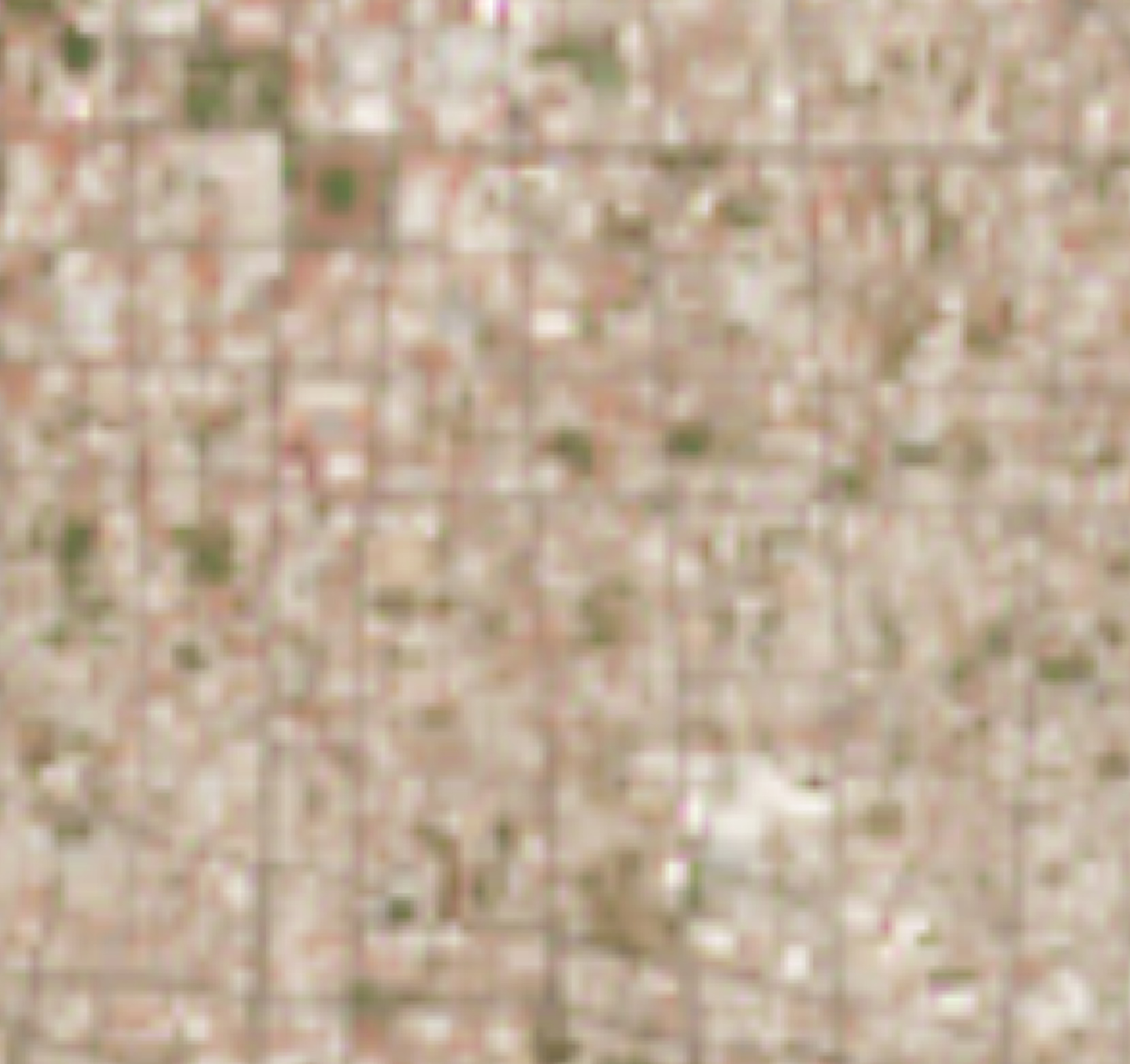} 
    \label{fig:gencomp:sub8} 
    } 
    \subfloat[]{
    \includegraphics[width=.29\linewidth]{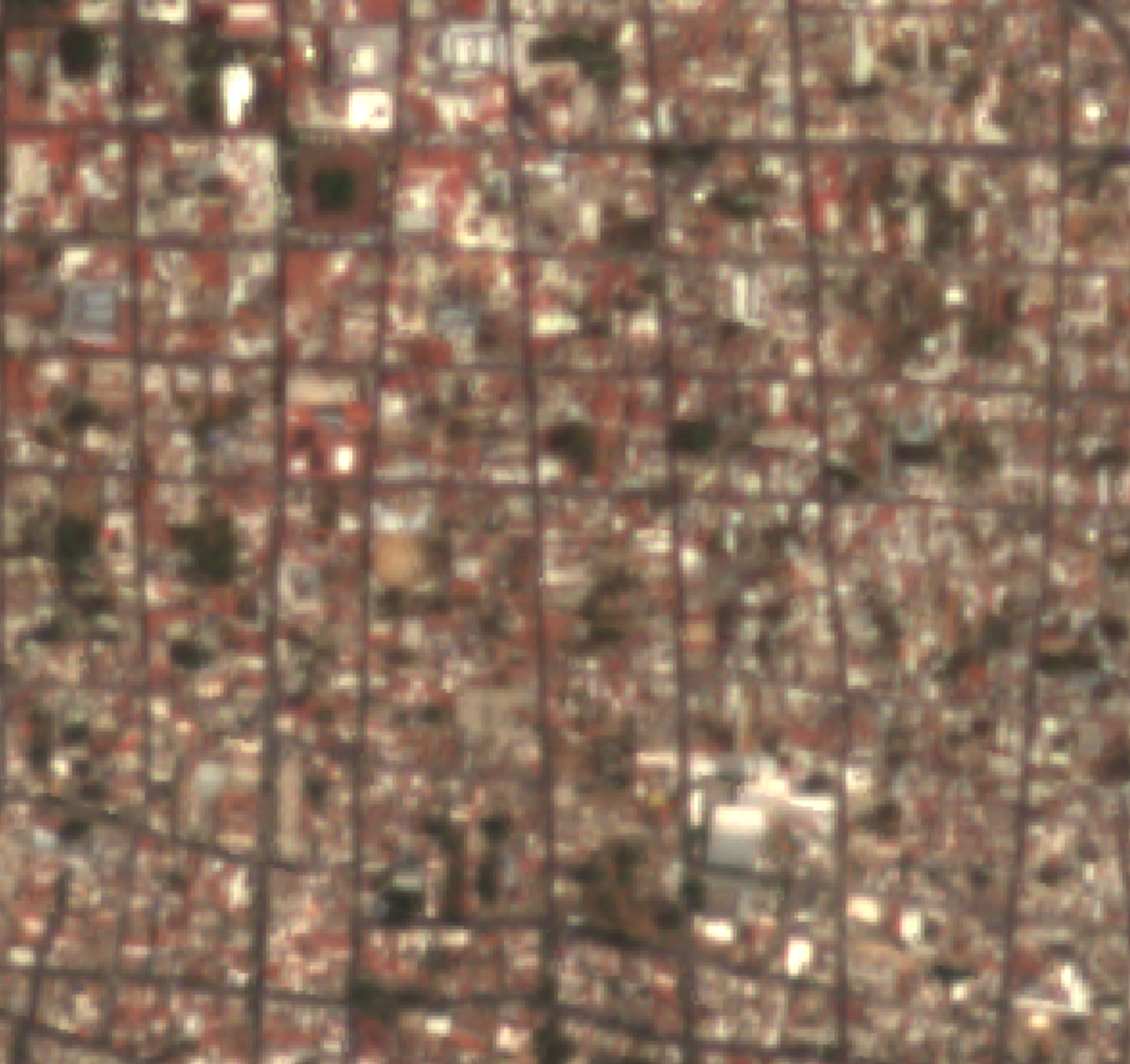}
    \label{fig:gencomp:sub7}
    }
    \subfloat[]{
    \includegraphics[width=.29\linewidth]{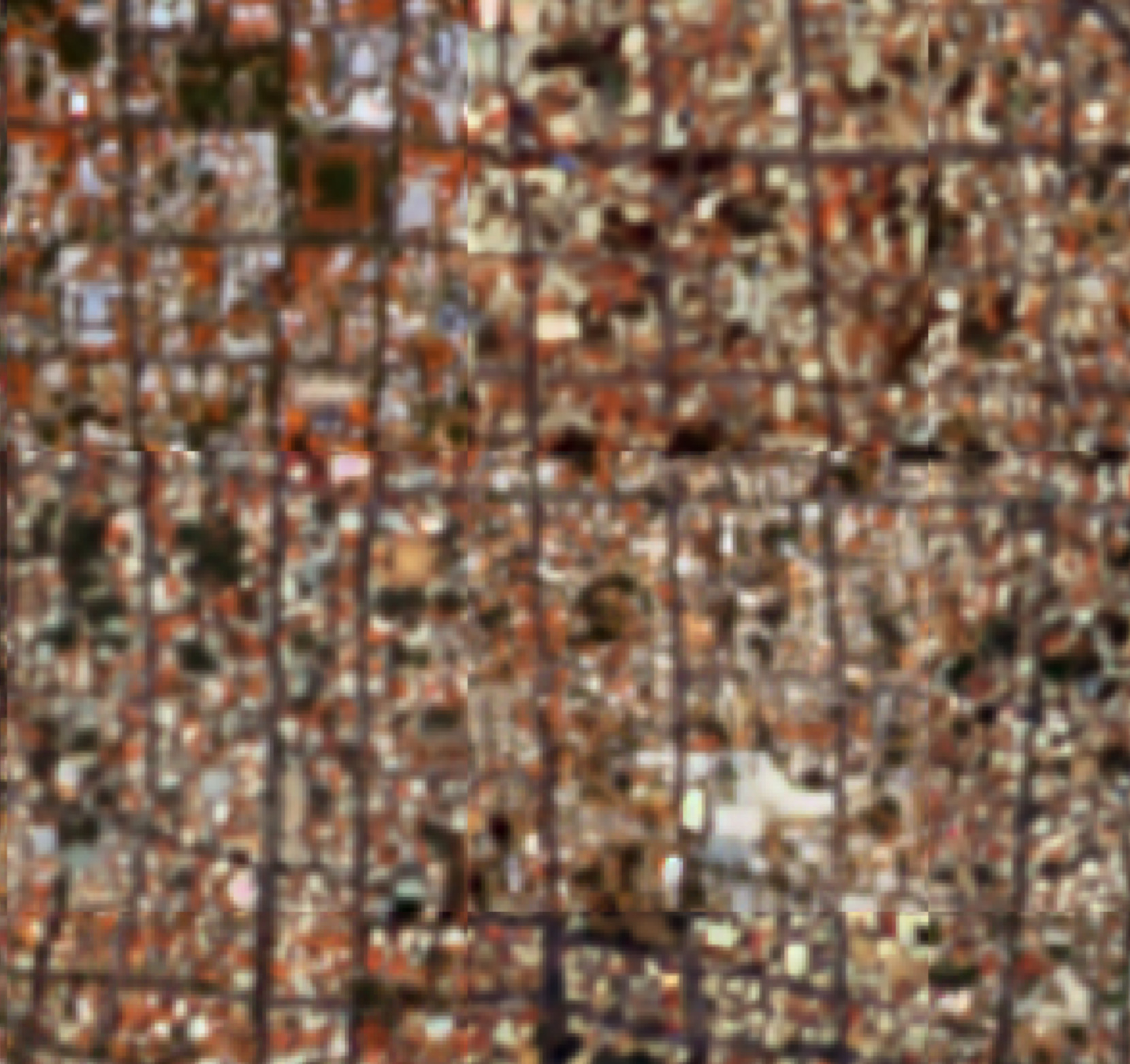}
    \label{fig:gencomp:sub9} 
  } 
    \caption{Figures of the region of interest in Tlaquepaque, Mexico. In the first row, images of a far view of the city are displayed. In the second row, we can observe a cropped part of the images in the first row for closer inspection. At the third row, a cropped part of the images from the second row is displayed, for the inspection of meter-scale features. The first, second, and third columns contain images from Planet Dove, Sentinel-II, and Synthetic Planet domains, respectively.}
    \label{fig:gencomp}
\end{figure*}

To effectively demonstrate the performance and advantages of the proposed I2I algorithm, a comparison in terms of image quality metrics is proposed. Table \ref{tab:imagequal} displays the values of three image quality/comparison metrics: Frechét Inception Distance (FID), mean Peak Signal-to-Noise Ratio (PSNR), and mean Learned Perceptual Image Patch Similarity (LPIPS). To obtain these values, we use the test set described in Section \ref{sec:dataset}.
The FID \cite{Heusel2017} measures the similarity between two data sets of images---usually a data set of generated images and one of reference images---by computing the distance between their feature representations extracted by a pre-trained Inception network. A lower FID indicates better feature alignment of the generated images with the reference data set.
The PSNR \cite{Hore2010} measures the reconstruction error between two images in terms of the ratio of the maximum possible pixel value and the mean squared error. A higher PSNR indicates a lower \mbox{pixel-wise} reconstruction error. It is important to note that PSNR is not a reliable metric for perceptual quality, as it does not account for the human visual system and the semantic content of the images \cite{Saharia2021}. Nonetheless, in the presented results, we compute the mean PSNR (mPSNR) of all Planet and Synthetic Planet corresponding patches, to further enrich the discussion.
The LPIPS \cite{Zhang2018LPIPS} measures the perceptual similarity between a pair of images by computing the distance between their feature representations extracted by a pre-trained deep neural network. A lower LPIPS indicates higher perceptual similarity and better visual quality. LPIPS is aligned with human perception, as it captures the semantic and structural differences between images \cite{Zhang2018LPIPS}. As with the PSNR, we compute the mean LPIPS (mLPIPS) of all corresponding Planet and Synthetic Planet patches. In a patch-wise comparison, the mLPIPS metric is better suited as a comparison metric than FID since its calculation is done on a patch-by-patch basis instead of focusing on the whole test set at once. Thus, since our tests involve direct comparison between corresponding images, mLPIPS is also employed.

We compare our proposed I2I algorithm, which has been trained using the SDM \cite{Wang2022} neural network as backbone, with other possible I2I solutions. First, the Sentinel-II test set is directly compared to Planet Dove images without any I2I translation method. The obtained metrics are shown in the first row of Table \ref{tab:imagequal}.
Second, we train two deep regression models \cite{Lathuiliere2020} whose loss objective is to directly match the pixel values of Planet patches when Sentinel-II corresponding patches are input. They share the same neural network backbone as the proposed DDM-based method. These models are identical to the one used in our experiments, except that the output linear layer directly predicts the pixel values of the corresponding Planet patch for a given input. 
The first deep regression model, whose metrics are displayed in the second row of Table \ref{tab:imagequal}, is a conventional regression model without any pre- or post-processing apart from simple $[-1, 1]$ normalization. The second regression model, shown in the third row, is identical but includes the whitening and coloring algorithms defined in Algorithms \ref{alg:whitening} and \ref{alg:coloring}. In the table, we refer to this as Regression + $\mathbf{W}, \mathbf{C}$. Testing this configuration assesses the influence of these processing methods in a conventional regression model and their impact on the metrics.
Next, we test the standard classifier-free guided DDIM \cite{ho2022classifierfree}, i.e., without the proposed whitening, coloring, and PSNR voting processes, using the exact same model architecture as the proposed method, to perform an ablation of these techniques. It is referred to in Table \ref{tab:imagequal} as Conditional DDIM. We then train an identical model, now adding the whitening and coloring procedures. To measure the impact of the PSNR voting in the inference results, the image quality metrics for this model are separated into two rows, the first being without PSNR voting---denoted as Conditional DDIM + $\mathbf{W}, \mathbf{C}$---and the second with it---Conditional DDIM + $\mathbf{W}, \mathbf{C}$ + PSNR Voting. All DDIM results were obtained using $N_{\text{final}}=64$. The proposed method, which includes PSNR voting, used $N_{\text{pre}}=8$ and $n_\text{noisy}=8$.

We also compare the performance of our method with five other \mbox{state-of-the-art} approaches for super resolution and \mbox{image-to-image} translation tasks. The first method is the \mbox{well-established} Pix2Pix \cite{Pix2Pix2017}, which has been proposed for multiple generative tasks, including \mbox{image-to-image} translation. It is an adversarial framework that uses both regression and discriminative losses to guide training of a generator \mbox{UNet-like} network. Following, we compare our method against the SwinIR \cite{liang2021swinir} super resolution model, which leverages the Swin Transformer architecture for enhanced image processing capabilities. The ShuffleMixer \cite{sun2022shufflemixer} architecture, known for its efficient mixing of features through channel shuffling and spatial mixing, is another model we compare against. The SRDenseNet \cite{zhang2018residual} is also included in our comparison---a model that utilizes dense connections to improve the flow of information and gradients throughout the network, which is particularly beneficial for the task of super resolution. Finally, we present results for the SR3 super resolution diffusion-based framework \cite{Saharia2021SR3}. The UNet structure of SR3 is analogous to the one used in the article that introduced the DDPM procedure \cite{Ho2020}. Thus, comparing it to our method provides an insight on the advantages of the neural network structure chosen for our experiments. 

\subsubsection{Comparison Metrics}
Looking at the numbers of Table \ref{tab:imagequal}, we observe that the full proposed technique, Conditional DDIM + $\mathbf{W}, \mathbf{C}$ + PSNR Voting, reaches, by a good margin, better results in mLPIPS and FID metrics when compared to all other methods and to the Sentinel-II baseline. FID dropped by 52.44 points when compared to Sentinel-II images---an expressive improvement of $53.46\%$. When compared to Regression and Regression + $\mathbf{W}$, the FID has been respectively reduced by 18.48 and by 21.8 points---an improvement of $28.82\%$ and $31.92\%$. Among the state-of-the-art methods, only SwinIR and SRDenseNet reached comparable mLPIPS and FID values to the ones from our approach. 

Interestingly, the vanilla conditional DDIM-based tested models---using SDM and SR3 backbones---—perform worse than all other methods in all metrics. They even underperform compared to the original images in both mLPIPS and mPSNR. The empirical reason is that the vanilla diffusion training seems confused by the multiple different overall tonalities of patches containing similar features, impacting the model's convergence ability. Deck and Bischoff \cite{deck2023easing} noticed that diffusion models tend to present such color-shifting problems, which worsen for larger images.

This issue is mitigated when whitening and coloring procedures are used, as evidenced by the performance superiority of both Conditional DDIM + $\mathbf{W}, \mathbf{C}$ and Conditional DDIM + $\mathbf{W}, \mathbf{C}$ + PSNR Voting (Proposed). Looking at mLPIPS results, we observe that the regression-based models slightly outperform the original Sentinel-II images, whereas Conditional DDIM + $\mathbf{W}, \mathbf{C}$ and Conditional DDIM + $\mathbf{W}, \mathbf{C}$ + PSNR Voting show a much more significant improvement. This indicates that, in a patch-by-patch basis, only DDIM + $\mathbf{W}, \mathbf{C}$ and DDIM + $\mathbf{W}, \mathbf{C}$ + PSNR Voting models were able to heavily reduce the perceptual distance of patches when compared to their Planet Dove versions. The proposed method with PSNR Voting has reduced mLPIPS by 0.1093 points, meaning that the produced patches are, on average, $36.71\%$ more perceptually similar to their corresponding Planet Dove patches than their Sentinel-II versions. Comparing the last two rows of the table, PSNR voting brings a significant improvement in all metrics: $-5.469\%$ for mLPIPS, $-2.665\%$ for FID, and $+2.144\%$ for mPSNR. Moreover, we observed that PSNR voting is specially useful to discard rare but existent undesired model hallucinations and poor translations, which would negatively impact, for example, change detection results. These improvements, due to the increased number of model evaluations, come at the cost of higher numerical complexity.

While regression models achieve higher mPSNR values, PSNR's preference for blurry regression-produced outputs does not align well with human perception. The same applies to the SwinIR, ShuffleMixer, and SRDenseNet methods, as they were also trained using regression objectives. Consequently, learning-based metrics like FID and LPIPS have become the standard in image generation comparisons, rendering the PSNR performance superiority of regression models less relevant.

\color{dgreen}
As discussed, the very poor performance shown by the conditional DDIM using the SR3 backbone is in part due to what plagued the performance of the classically trained DDIM model using the SDM backbone. However, even compared to it, the SR3 model presented far worse numbers. This fact shows how the model architecture chosen for our method has been able to better extract the complex features of the conditioning input and generate better translations.
To further explore the influence of the backbone architecture, we also tested the proposed method using SR3 as the backbone, applying our forward and reverse diffusion methods. This dramatically improved the performance compared to its vanilla DDIM training, indicating that our approach enhances the capabilities across different network architectures. However, when compared to using the SDM backbone, the proposed method with the SR3 backbone performed much worse in both mLPIPS and FID, and could not effectively super-resolve the images. This suggests that while our proposed method improves performance for different backbones, the choice of architecture remains critical for achieving optimal results. The SDM backbone has been better able to extract complex features from the conditioning input and generate superior translations, highlighting the importance of selecting an appropriate architecture in conjunction with our proposed methods. Further details about the SDM backbone and underlying hyperparameters are discussed in Appendix \ref{sec:apx:model_s}.
\color{black}

\subsubsection{Inference Hyperparameters}
To verify the impact of the choice of inference hyperparameters linked to the proposed method, we present a performance comparison for different values of $N_\text{final}$, $N_\text{pre}$ and $n_\text{noisy}$ in Table \ref{tab:comp_nit}. In terms of mLPIPS, $N_\text{final}=64$, $N_\text{pre}=8$ and $n_\text{noisy}=8$ reached the best result of $0.1884$. We use this configuration in our change detection experiments, as a lower perceptual distance between identical patches is crucial for a high change detection performance. Interestingly, however, $N_\text{final}=32$, $N_\text{pre}=4$ and $n_\text{noisy}=4$ reached a lower FID, which suggests that further increasing the number of iterations does not bring any value with regards to image quality. Moreover, unintuitively, $N_\text{final}=48$, $N_\text{pre}=6$ and $n_\text{noisy}=6$ showed worse performance metrics compared to the former configuration. This may be due to the fact that the number of training diffusion steps was set to 1024, and 48 is not a divisor of such number. This implicates that the particular timestep values input to the model during the DDIM sampling were not seen during training, leading to such worse performance values. Finally, for $N_\text{final}=16$, $N_\text{pre}=2$ and $n_\text{noisy}=2$, the performance metrics were clearly degraded, which suggests that $N_\text{final}=32$, $N_\text{pre}=4$ and $n_\text{noisy}=4$ would be a good balance between numerical complexity and performance. In addition, the usefulness of PSNR voting is confirmed when this configuration is compared to the results of the model using 64 DDIM iterations without PSNR voting, displayed in Table \ref{tab:imagequal}. That is, even with a lower total number of iterations (48 vs 64), the configuration $N_\text{final}=32$, $N_\text{pre}=4$ and $n_\text{noisy}=4$ reached better mLPIPS and FID values compared to the DDIM without PSNR voting.

\begin{figure*}
    \centering
    \subfloat[Pre-Event Sentinel-II]{
    \includegraphics[width=0.6\linewidth]{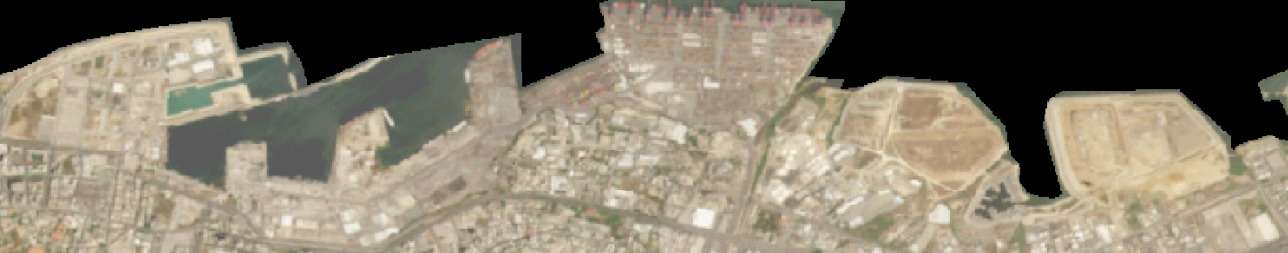}
    \label{fig:genres:sub1} 
    }
    \hfill
    \subfloat[Pre-Event Synthetic Planet Dove (Regression + $\mathbf{W}, \mathbf{C}$)]{
    \includegraphics[width=0.6\linewidth]{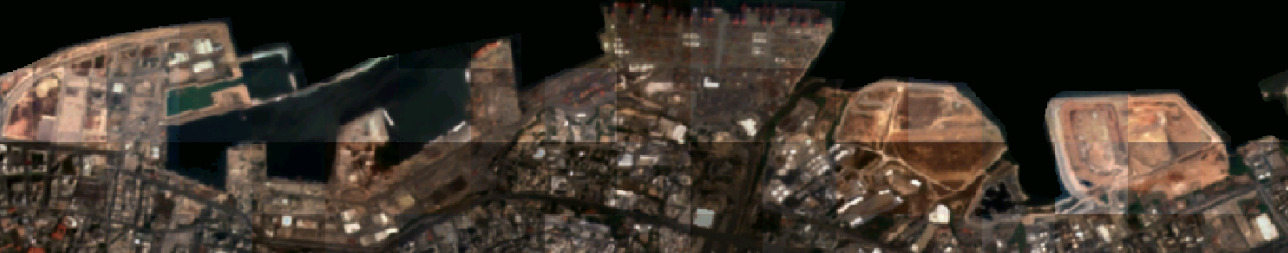} 
    \label{fig:genres:sub2}
    } 
    \hfill
    \subfloat[Pre-Event Synthetic Planet Dove (Proposed)]{
    \includegraphics[width=0.6\linewidth]{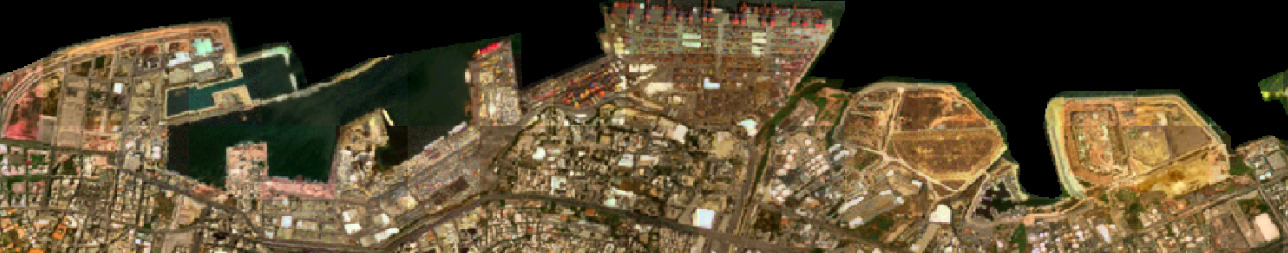} 
    \label{fig:genres:sub3}
    } 
    \hfill 
    \subfloat[Pre-Event Planet Dove]{
    \includegraphics[width=0.6\linewidth]{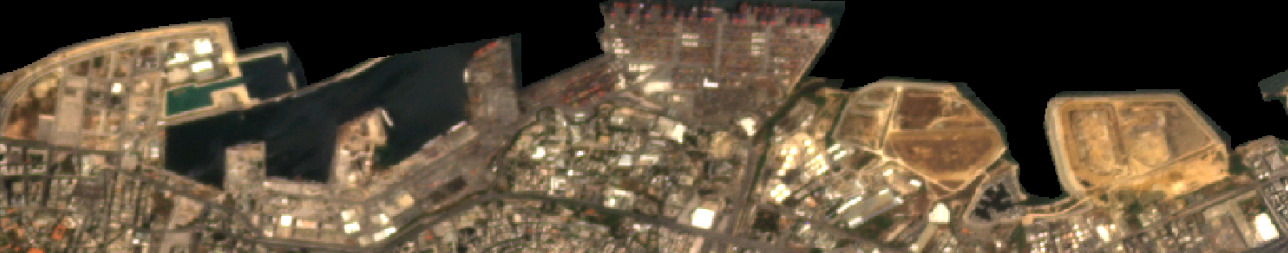}
    \label{fig:genres:sub4}
    }
    \hfill
    \subfloat[Post-Event Planet Dove]{
    \includegraphics[width=0.6\linewidth]{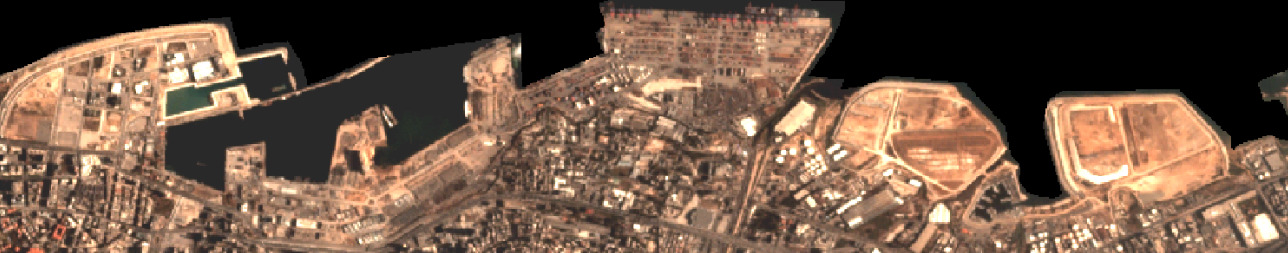}
    \label{fig:genres:sub5} 
    }
    \hfill
    \subfloat[Change Detection Ground Truth]{
    \includegraphics[width=0.6\linewidth]{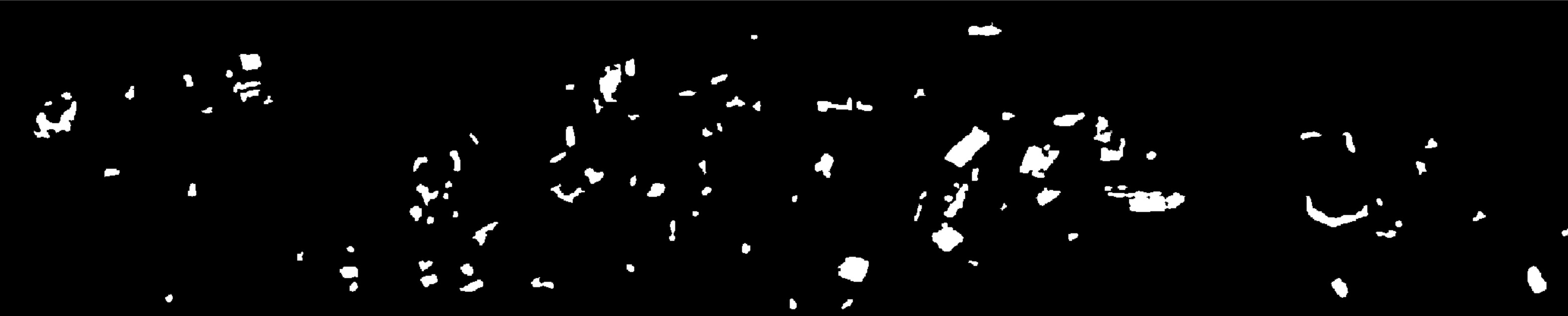}
    \label{fig:genres:cd} 
    } 
    \caption{Figures of the region of interest in the port of Beirut.} 
    \label{fig:genres}
\end{figure*} 

\subsection{Visual Examples}

In our evaluation of image generation methods, as detailed in Table \ref{tab:imagequal}, Figure \ref{fig:gencomp2} showcases the visual differences using examples from a region in Beirut. The sequence begins with our proposed method (\ref{fig:gencomp2:sub3}), which utilizes the DDIM approach enhanced by whitening and coloring processes (\(\mathbf{W}, \mathbf{C}\)) and PSNR voting, and ends with the SR3 diffusion-based method (\ref{fig:gencomp2:sub12}). The Sentinel-II (\ref{fig:gencomp2:sub1}) and Planet Dove (\ref{fig:gencomp2:sub2}) images serve as input and target references, respectively.

Our method (\ref{fig:gencomp2:sub3}) demonstrates superior patch consistency and detail resolution, outperforming the other techniques. The addition of \(\mathbf{W}, \mathbf{C}\) significantly enhances contrast and clarity, as evidenced by the comparison with the vanilla DDIM result (\ref{fig:gencomp2:sub5}), which suffers from tonal inconsistencies and a mosaic-like appearance. These issues are notably absent in Figure \ref{fig:gencomp2:sub4}, where \(\mathbf{W}, \mathbf{C}\) have been applied.

Despite these improvements, some regions in Figure \ref{fig:gencomp2:sub4} exhibit deficiencies. The center-west patch is notably blurry and washed-out, suggesting a failure in the generation process. The south-west and south-east patches display a darker tone compared to adjacent areas, with the south-east patch also presenting blurry, low-resolution features not observed in the corresponding region of Figure \ref{fig:gencomp2:sub3}. This demonstrates how the PSNR voting mechanism improves overall generation quality. 

The regression-based methods (\ref{fig:gencomp2:sub6} and \ref{fig:gencomp2:sub7}), which share the same neural network backbone as the proposed method, yield images with a blurry appearance, which does not significantly enhance the apparent resolution when compared to the Sentinel-II image. 

In a similar venue, the algorithms Pix2Pix (\ref{fig:gencomp2:sub8}), ShuffleMixer (\ref{fig:gencomp2:sub9}), SRDenseNet (\ref{fig:gencomp2:sub10}), and SwinIR (\ref{fig:gencomp2:sub11}) maintain good inter-patch consistency but fall short in capturing the high-resolution details that our method achieves. The SR3 method (\ref{fig:gencomp2:sub12}) performs poorly, with color inconsistencies and feature representation problems that are more severe than those observed when our method is applied without \(\mathbf{W}, \mathbf{C}\).

\begin{figure*}
    \centering 
    \subfloat[Pre-Event Sentinel-II]{
    \includegraphics[width=0.6\linewidth]{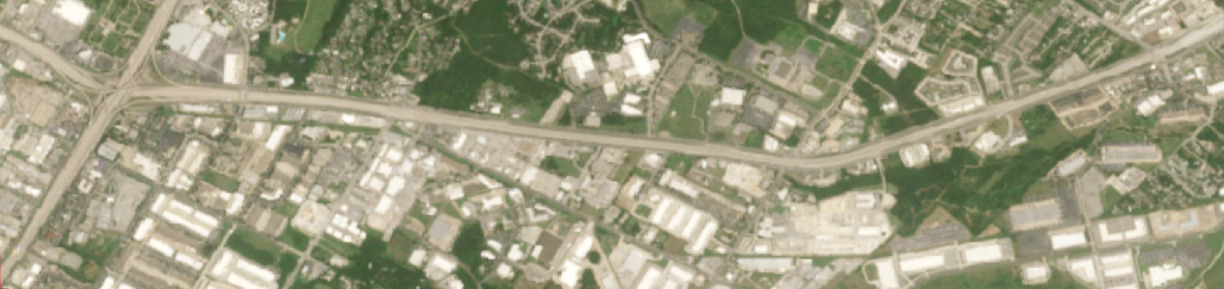}
    \label{fig:genres_austin:sub1} 
    } 
    \hfill 
    \subfloat[Pre-Event Synthetic Planet Dove (Regression + $\mathbf{W,C}$)]{
    \includegraphics[width=0.6\linewidth]{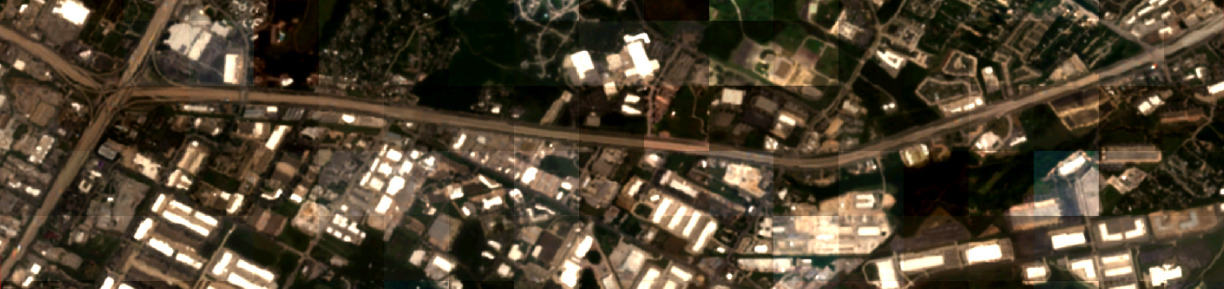} 
    \label{fig:genres_austin:sub2}
    }
    \hfill
    \subfloat[Pre-Event Synthetic Planet Dove (Proposed)]{
    \includegraphics[width=0.6\linewidth]{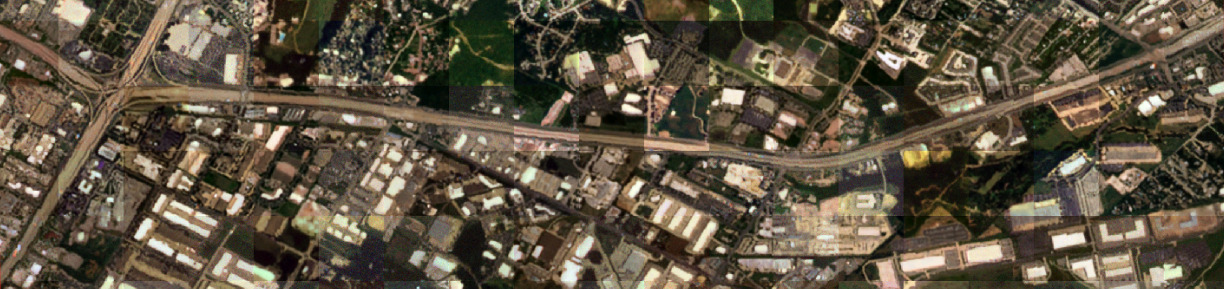} 
    \label{fig:genres_austin:sub3}
    }
    \hfill
    \subfloat[Pre-Event Planet Dove]{
    \includegraphics[width=0.6\linewidth]{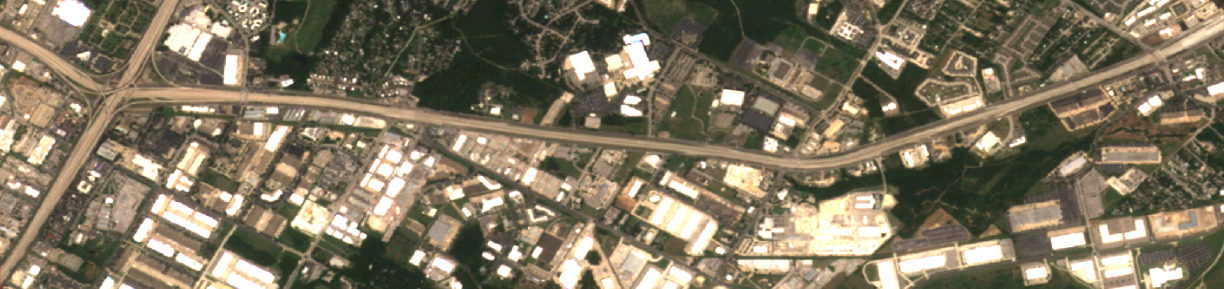}
    \label{fig:genres_austin:sub4}
    }
    \hfill
    \subfloat[Post-Event Planet Dove]{
    \includegraphics[width=0.6\linewidth]{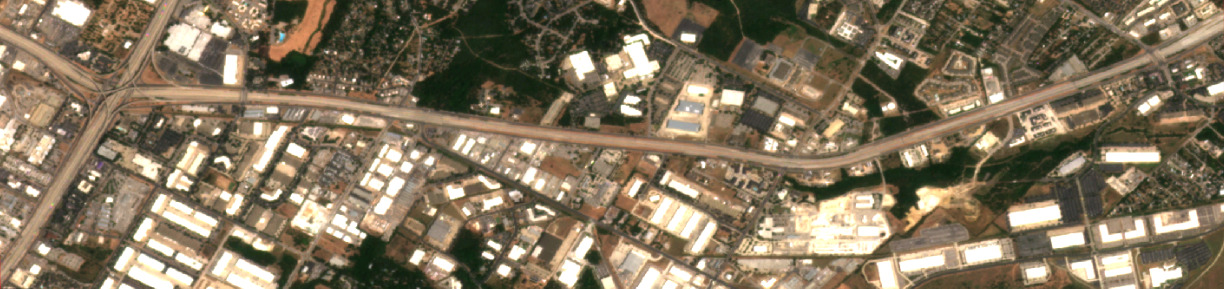} 
    \label{fig:genres_austin:sub5} 
    }
    \hfill
    \subfloat[Change Detection Ground Truth]{
    \includegraphics[width=0.6\linewidth]{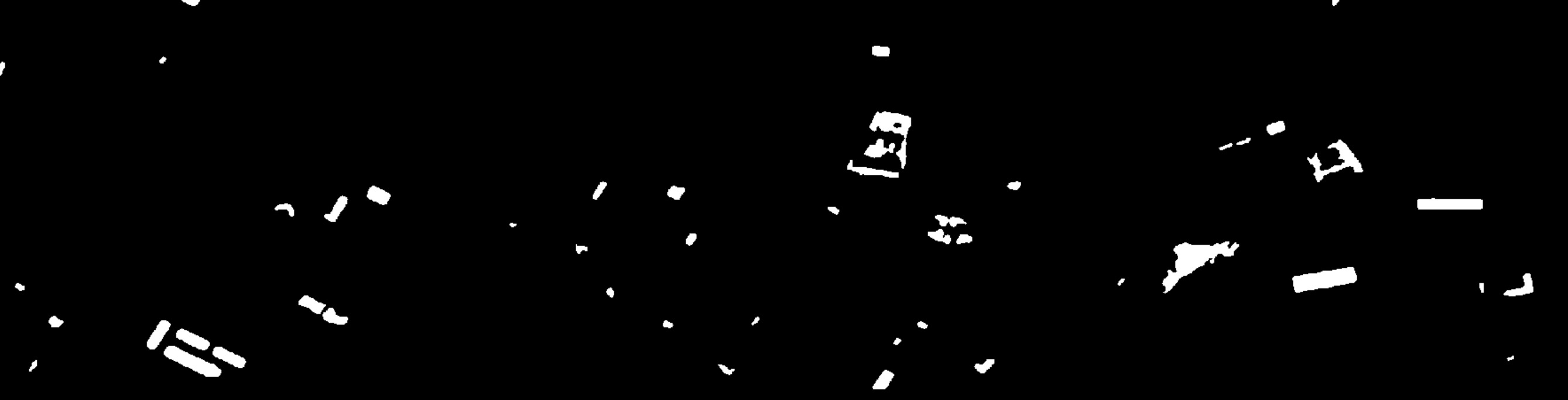}
    \label{fig:genres_austin:gt}
    }
    \caption{Figures of the region of interest in Austin.} 
    \label{fig:genres_austin} 
\end{figure*} 

In a direct comparison between Sentinel-II, Planet Dove, and the synthetic Planet Dove image generated by the proposed method, Figure \ref{fig:gencomp} presents a region from the city of Tlaquepaque, Mexico. Each of the three rows of figures shows a different view of the city, with different scales, to better compare different aspects of the images. The three columns show images from Planet Dove, Sentinel-II, and images generated by the proposed method, respectively. The first row shows a far view of the city that depicts multiple complex urban features: houses, buildings, storage facilities, factories, roads, and scattered vegetation. When looking at the quality of the figures, Figure \ref{fig:gencomp:sub1} clearly shows great contrast between features due to it coming from a higher spatial resolution satellite when compared to Figure \ref{fig:gencomp:sub2}. Figure \ref{fig:gencomp:sub3} is the synthetic high-resolution image generated with the proposed model using Figure \ref{fig:gencomp:sub2} as input. It exhibits a higher contrast between features, compared to the Sentinel-II version, while maintaining feature consistency between patches to the point that it becomes almost impossible to distinguish individual patches in the full images. The same can be said about Figure \ref{fig:gencomp:sub6}, which shows a close-up view of a region located around the center of Figure \ref{fig:gencomp:sub3}, where homogeneity is still observed, and the super-resolution aspect of the method starts to be noticeable. That is, high-frequency information not clearly observable in Figure \ref{fig:gencomp:sub5} gets enhanced by the algorithm, resulting in Figure \ref{fig:gencomp:sub6}. Finally, the last row depicts a highly zoomed-up area of the city, where individual residential buildings can be visualized. It is clearly noticeable in Figure \ref{fig:gencomp:sub7} how the higher spatial resolution information provided by the Planet Dove sensor enables the visualization of details otherwise imperceptible in the Sentinel-II image (Figure \ref{fig:gencomp:sub8}), as well as properly separates features coming from different elements in the image, e.g., roads are easily distinguishable from buildings, whereas in \ref{fig:gencomp:sub8} the pixels from narrow roads are interleaved with pixels from buildings. This drawback of Sentinel-II images is alleviated by the proposed method, as can be observed in Figure \ref{fig:gencomp:sub9}. There, pixel-level details are almost as clear and as distinguishable as in Figure \ref{fig:gencomp:sub7}. In addition to this, the synthetic image does not show visible undesirable hallucinations, which enables change detection tasks.

Figures \ref{fig:genres} and \ref{fig:genres_austin} respectively show images from the selected regions of Beirut and Austin. Their first, second, third, and fourth rows individually show pre-event Sentinel-II, pre-event Synthetic Planet Dove (Regression + $\mathbf{W}$), pre-event Synthetic Planet Dove (Proposed), pre-event Planet Dove, post-event Planet Dove, and the change detection ground truth obtained by applying Algorithm \ref{alg:cd}, which is discussed in Section \ref{sec:experiments:change_detection}, using Planet Dove pre and post-event images. Instead of manual annotations, we use this ground truth since our I2I translation method aims to closely match low-resolution input to the style and resolution of high-resolution images, specifically Planet Dove in our experiments.

In both regions, the tonality of the synthetic images generated using the proposed method (Figs. \ref{fig:genres:sub3} and \ref{fig:genres_austin:sub3}) closely resembles those of the Planet pre-event images (Figs. \ref{fig:genres:sub4} and \ref{fig:genres_austin:sub4}). Also, when compared with the pre-event Sentinel-II images used as inputs to the DDM-based method (Figs. \ref{fig:genres:sub1} and \ref{fig:genres_austin:sub1}), it is easily observable how they have an improved apparent resolution and contrast. It is also noticeable how no perceptible undesired hallucinated features are present, favoring their use for change detection. Compared to the Regression + $\mathbf{W}$ synthetic images (Figs. \ref{fig:genres:sub2} and \ref{fig:genres_austin:sub2}), our method delivered images with higher contrast and visual resolution, which translated into images that better match the characteristics of Planet Dove imagery.

The few patching brightness mismatches can be easily alleviated by performing inference with overlapping patches, as done in \cite{wan_psc_2024}. However, this would greatly increase the numerical complexity of the inference, e.g. by a factor of 4 if the overlap is 1/2 of the size of the patch, or by a factor of 16 if the overlap is 3/4 of the size of the patch. The reverse diffusion process is costly because it needs numerous model forward passes to infer a single patch. Therefore, we do not explore this possibility in our paper.

\begin{figure*}
    \centering
    \includegraphics[width=0.8\linewidth]{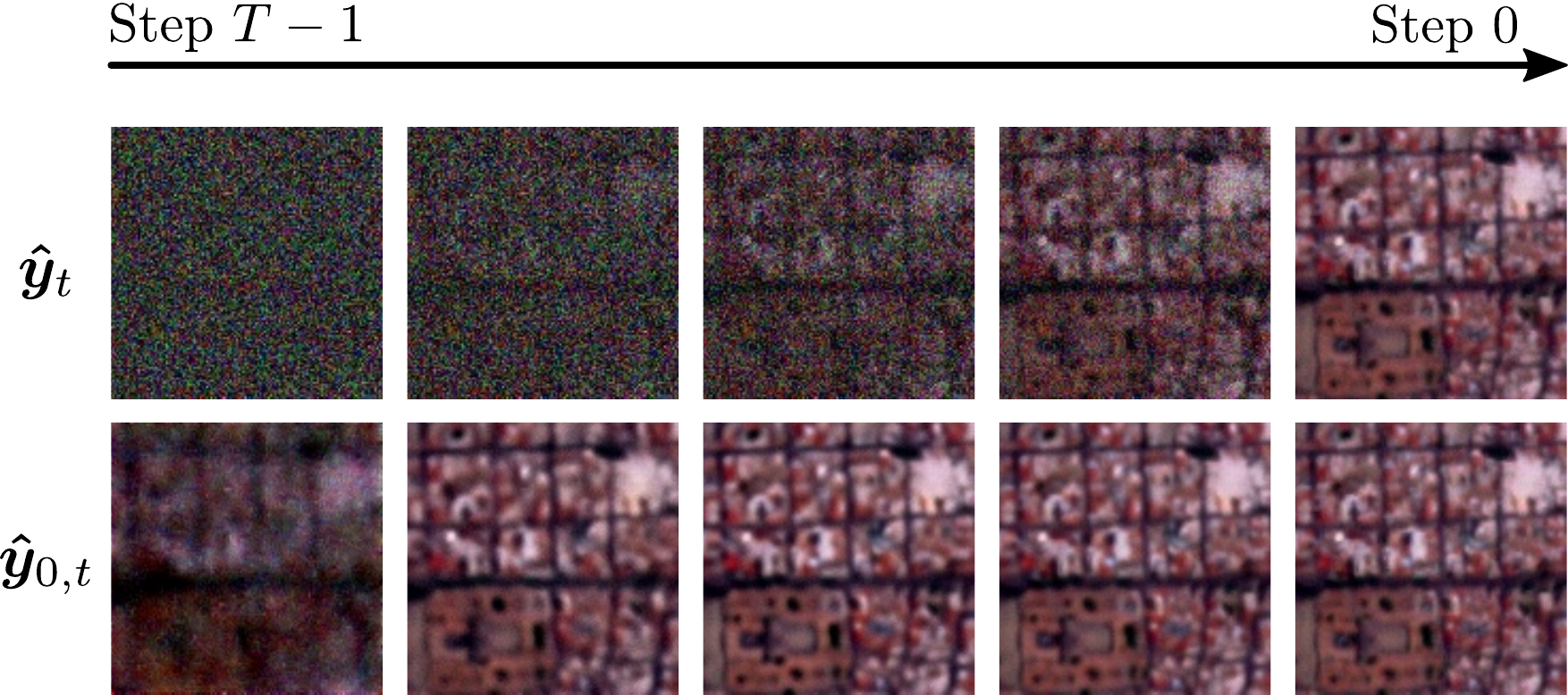}

    \caption{Visualization of the intermediate steps of the diffusion process. The first row shows the progression of the noisy image $\bm{y}_t$ along diffusion timesteps. The second row shows their corresponding denoised predictions $\bm{y}_{0,t}$.}
    \label{fig:interm}
\end{figure*}
To demonstrate the dynamics of the forward and reverse diffusion processes for this specific task, Figure \ref{fig:interm} shows an example of the intermediate steps of the denoising procedure. The first row shows the progression of the noisy image $\bm{y}_t$ along diffusion steps $t \in {0, 1, ..., T-1}$, whereas the second row shows their corresponding denoised outputs $\bm{y}_{0,t}$.
During the very initial steps, most of the low-frequency information is recovered, resulting in blurry low-quality outputs. Around halfway of the process, the denoised image contains almost all high-frequency features of the final prediction. Thus, subsequent steps serve only to recover the finest details for the final prediction.

\begin{algorithm}
\setstretch{1.7}
\caption{Targetless Change Detection}
\label{alg:cd}
\begin{algorithmic}[1]
\REQUIRE $\bm{I}_{\text{pre}}$, $\bm{I}_{\text{post}}$, $l$, $n_{\text{ch}}$, $h$, $w$,  $\omega$, $w_{\text{gauss}}$, $w_{\text{otsu}}, e_{\text{max}}, n_{\text{min}}$

\ENSURE $\bm{I}_{\text{diff}}$

\STATE $\bm{I}_{\text{pre}} \gets$ {\fontsize{8.5}{8.5}\selectfont
$\left[\min\left(I_{\text{pre}}^{i,j,k}, \mu\left(\bm{I}_{\text{pre}}\right) + 6\sigma_{\bm{I}_{\text{pre}}}\right)\right]_{1\leq i \leq n_{\text{ch}}, \:, 1 \leq j \leq h, \: 1 \leq k \leq w}$ }\label{alg:cd:clip_pre}

\STATE $\bm{I}_{\text{post}} \gets $ {\fontsize{8.4}{8.4}\selectfont
$\left[\min\left(I_{\text{post}}^{i,j,k}, \mu\left(\bm{I}_{\text{post}}\right) + 6\sigma_{\bm{I}_{\text{post}}}\right)\right]_{1\leq i \leq n_{\text{ch}}, \:, 1 \leq j \leq h, \: 1 \leq k \leq w}$} \label{alg:cd:clip_post}

\STATE $\bm{I}_{\text{pre}} \gets \text{GaussianBlur}\left(\frac{\bm{I}_{\text{pre}} - \min \bm{I}_{\text{pre}} \cdot \mathbf{J}_{n_\text{ch} \times h \times w}}{\max \bm{I}_{\text{pre}} - \min \bm{I}_{\text{pre}}}, w_{\text{gauss}}\right)$ \label{alg:cd:gausspre}

\STATE $\bm{I}_{\text{post}} \gets \text{GaussianBlur}\left(\frac{\bm{I}_{\text{post}} - \min \bm{I}_{\text{post}} \cdot \mathbf{J}_{n_\text{ch} \times h \times w}}{\max \bm{I}_{\text{post}} - \min \bm{I}_{\text{post}}}, w_{\text{gauss}}\right)$ \label{alg:cd:gausspost}

\STATE $\bm{I}_{\text{pre}} \gets \frac{\bm{I}_{\text{pre}} - \mu \left(\bm{I}_{\text{pre}}\right)\mathbf{J}_{n_\text{ch} \times h \times w}}{\sigma\left(\bm{I}_{\text{pre}}\right)}$ \label{alg:cd:norm_pre}

\STATE $\bm{I}_{\text{post}} \gets \frac{\bm{I}_{\text{post}} - \mu \left(\bm{I}_{\text{post}}\right)\mathbf{J}_{n_\text{ch} \times h \times w}}{\sigma\left(\bm{I}_{\text{post}}\right)}$ \label{alg:cd:norm_post}

\STATE $\bm{I}_{\text{diff}} \gets \left(\bm{I}_{\text{post}} - \bm{I}_{\text{pre}}\right)^{\odot 2}$ \label{alg:cd:i_diff}

\STATE $\bm{I}_{\text{diff}} \gets \frac{1}{n_{\text{ch}}} \sum_{i=1}^{n_{\text{ch}}} \bm{I}^i_{\text{diff}} $ \label{alg:cd:diffchmean}

\STATE $\bm{I}_{\text{diff}} \gets \frac{\bm{I}_{\text{diff}} - \min \bm{I}_{\text{diff}} \cdot \mathbf{J}_{n_\text{ch} \times h \times w}}{\max \bm{I}_{\text{diff}} - \min \bm{I}_{\text{diff}}}$ \label{alg:cd:scalediff}

\STATE $\bm{I}_{\text{diff}} \gets \mathbf{H}\left({\bm{I}}_{\text{diff}} - \omega\mathbf{J}_{n_\text{ch} \times h \times w} \right) \odot \bm{I}_{\text{diff}}$ \label{alg:cd:threshold}

\STATE $\bm{I}_{\text{diff}} \gets 
\text{Otsu}\left(\bm{I}_{\text{diff}}, w_{\text{otsu}}\right)$ \label{alg:cd:otsu}

\STATE $\bm{L}_{\text{diff}} \gets \text{DBSCAN}\left(\bm{I}_{\text{diff}}, e_{\text{max}}, n_{\text{min}}\right)$ \label{alg:cd:dbscan_label}

\STATE $\bm{I}_{\text{diff}} \gets 
\begin{cases}
    I_{\text{diff}}^{i,j,k} & \text{if } L_{\text{diff}}^{i,j,k} \neq -1\\
    0 & \text{otherwise} 
\end{cases} $\\
{\quad \quad \quad \quad\fontsize{8.5}{8.5}\selectfont${1\leq i \leq n_{\text{ch}}, \:, 1 \leq j \leq h, \: 1 \leq k \leq w}$} \label{alg:cd:dbscan_out}

\end{algorithmic}
\end{algorithm}

\subsection{Change Detection as a Use Case}
\label{sec:experiments:change_detection}
By transforming images from one domain to another, image-to-image translation methods can effectively highlight differences and changes between images taken at different times or under different conditions. Here, we explore the use of our image-to-image translation method for change detection, demonstrating its practical utility in this specific context. It is important to note that change detection is just one possible use case for our method. The results presented here aim to showcase how our image-to-image translation approach can be beneficial for practical applications, without limiting its potential to other domains.

\subsubsection{Rationale for Selective Comparisons}
We focus on evaluating our model’s performance against one of the regression models previously compared: Regression + $\mathbf{W,C}$. Previous experiments have already demonstrated the superiority of our model in image generation compared to state-of-the-art algorithms. Specifically, our method outperforms others in image quality, particularly with respect to the LPIPS metric, which is a patch-by-patch comparative metric. A lower LPIPS score indicates better translation and fewer artifacts, which correlates with higher quality change detection results.
To maintain a concise and focused manuscript, we limit our change detection experiments to comparisons against regression models. This approach allows us to highlight the power of using diffusion models against a regression-based objective, which is also employed by the methods ShuffleMixer, SRDenseNet, and SwinIR.

It is important to note that the tested change detection algorithm does not necessarily extract the full potential of our image translation method, as it contains steps that do not fully utilize the added visual resolution to our advantage. Additionally, using this specific simple HCD method with images generated by the proposed method does not necessarily produce better HCD results when compared to other methods such as ShuffleMixer, SRDenseNet, and SwinIR. These methods have performed poorly in super resolution but very well in patch-wise feature consistency, and Algorithm \ref{alg:cd} does not make use of the higher resolution features produced by our image translation method. However, employing a more advanced change detection algorithm is beyond the scope of this paper. The HCD experiments are intended to demonstrate that our proposed method could be potentially useful for change detection. A more advanced learning-based change detection procedure using the produced synthetic images will be explored in future work.
Furthermore, Regression + $\mathbf{W,C}$ is trained with the same neural network backbone as the proposed method. By doing so, we effectively showcase the advantages of our diffusion-based method under consistent conditions, ensuring a clear and concise presentation of our results.

\subsubsection{Procedure}
Algorithm \ref{alg:cd} defines a simple change detection procedure for a pair of images, $\textbf{I}_{\text{pre}}$ and $\textbf{I}_{\text{post}}$, taken before and after a change event, respectively. It is targetless, meaning that no specific class of detections is to be highlighted, but instead, all changes are treated equally.
The algorithm, which is used in the change detection experiments presented henceforth, is a structured approach to targetless change detection, leveraging the robustness of Otsu’s method \cite{Crapsu1979}, a well-known adaptive thresholding technique. It works by automatically selecting the optimal threshold value to separate pixels into two classes, foreground, and background, based on their grayscale levels, in a sliding-window fashion. Intuitively, it calculates the threshold that minimizes intra-class variance, effectively distinguishing between changed and unchanged areas in an image.

Lines \ref{alg:cd:clip_pre} and \ref{alg:cd:clip_post} of Algorithm \ref{alg:cd} individually clip $\textbf{I}_{\text{pre}}$ and $\textbf{I}_{\text{post}}$ to their mean plus six times their standard deviations, to remove extreme outliers that could affect the difference image calculation. 
Lines \ref{alg:cd:gausspre} and \ref{alg:cd:gausspost} apply Gaussian filtering to $[0, 1]$-normalized versions of the images, to smooth the images, which alleviates the value transitions, suppressing potential false alarms. Following, in Lines \ref{alg:cd:norm_pre} and \ref{alg:cd:norm_post}, the images are standardized so that the pixel values of both images become comparable.
Line \ref{alg:cd:i_diff} computes the squared difference between the two images in order to highlight the changes, followed by a channel-wise mean in Line \ref{alg:cd:diffchmean}, to compact the change map into a 2D matrix.
Line \ref{alg:cd:threshold} zeroes out elements of the difference image that are smaller than $\omega$. $\mathbf{H}$ is the element-wise Heaviside step function. By setting such a global threshold, we effectively filter out minor variations in pixel values that are likely due to noise. This pre-processing step simplifies the image data, ensuring that when Otsu’s method is applied, it can more accurately focus on the substantial changes—those that are of real interest in the context of the event being analyzed.
Then, in Line \ref{alg:cd:otsu}, the Otsu's adaptive thresholding \cite{Crapsu1979} is applied with window size of $w_{\text{otsu}}$, to generate a binary map whose nonzero elements correspond to potential changes. In Lines \ref{alg:cd:dbscan_label} and \ref{alg:cd:dbscan_out}, to alleviate the presence of detection noise, i.e., to remove positive pixels that are completely isolated and scattered through the binary map, Density-Based Spatial Clustering of Applications with Noise (DBSCAN) \cite{DBSCAN1996} algorithm is applied. It locates clusters of positive pixels of significant size and discards pixels that do not belong to any detected cluster, classifying them hereby as noisy pixels. Its parameters are the maximum Euclidean distance between pixels of a cluster, $e_{\text{max}}$, and the minimum number of pixels that a cluster should contain, $n_{\text{min}}$. The output of Line \ref{alg:cd:dbscan_label}, $\bm{L}_{\text{diff}}$, is a matrix with the dimensions of $\bm{\textit{I}}_{\text{diff}}$, whose pixels are integers from -1 to the number of identified clusters. All pixels equal to -1 are classified as noise and are finally discarded in the last line of Algorithm \ref{alg:cd}, resulting in a clean binary map, where nonzero pixels are classified as changed pixels.
For the executed experiments, the values of $w_{\text{gauss}}$,  $w_{\text{otsu}}$, $e_{\text{max}}$, and $n_{\text{min}}$ were set to 11, 1023, 5, and 48, respectively.

\begin{figure*}
\centering
\subfloat[Sentinel-II (Pre) + Planet (Post)]{
    \includegraphics[width=.8\linewidth]{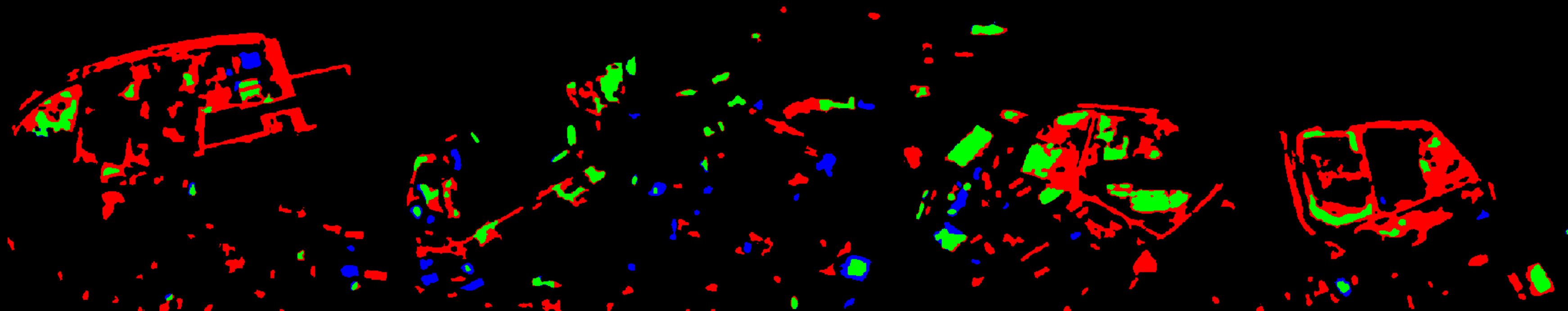}
    \label{fig:cd_beirut:beirut_s2planet}
}

\hfill

\subfloat[Regression + $\mathbf{W,C}$ Synthetic Planet (Pre) + Planet (Post)]{
    \includegraphics[width=.8\linewidth]{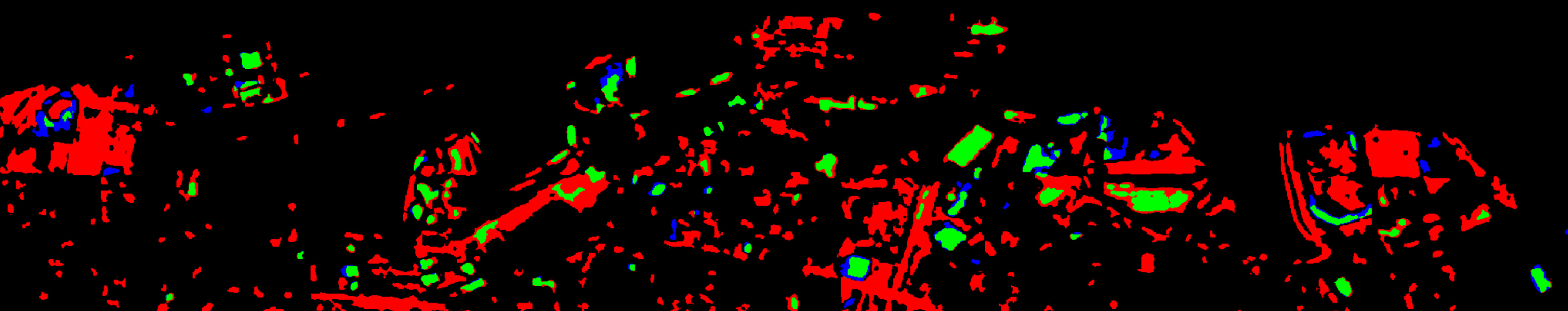}
    \label{fig:cd_beirut:beirut_regression}
}
\hfill

\subfloat[Proposed Synthetic Planet (Pre) + Planet (Post)]{
    \includegraphics[width=.8\linewidth]{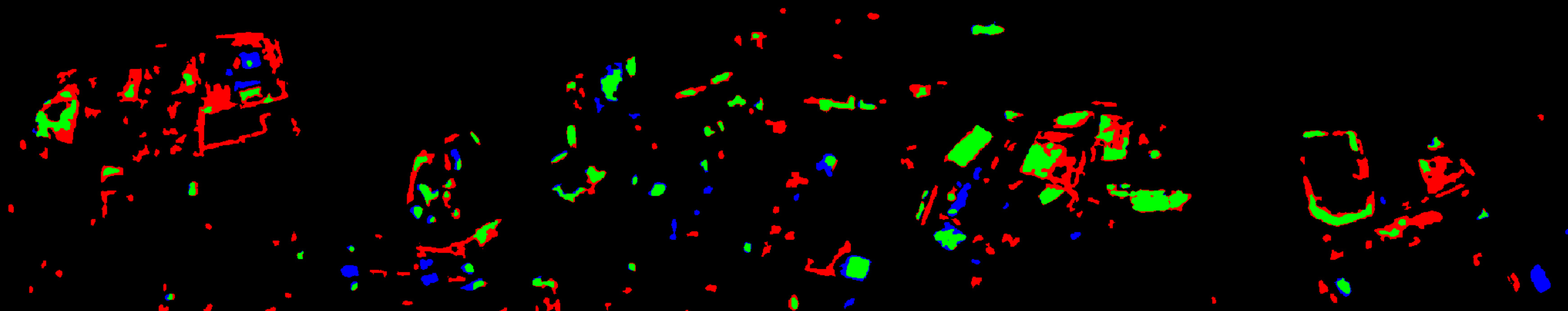}
    \label{fig:cd_beirut:beirut_ourmethod}
}

\subfloat[Ground Truth---Planet Pre and Post]{
\includegraphics[width=.8\linewidth]{figs/beirut/beirut_cd_gt.jpg}
\label{fig:cd_beirut:gt}
}
\caption{Change detection results in Beirut Port Region for a fixed detection rate of 0.75. Pixels in green, blue, and red are true positives, false negatives, and false positives, respectively.}
\label{fig:cd_beirut}
\end{figure*}

\subsubsection{HCD Experiments}
Two pairs of Sentinel-II pre-event and Planet Dove post-event images from two different cities were selected for a change detection comparison experiment. We also use their respective pre-event Planet Dove images to produce a change detection ground truth. The Sentinel-II pre-event images are part of the test set mentioned in Section \ref{sec:dataset}. The first pair of images is in Beirut, Lebanon, surrounding the port region. It has been extracted from the northern part of the images in Figure \ref{fig:beirut_all}, where no clouds are present. This image has been considered because it showcases multiple different types of changes, making targetless change detection useful. Moreover, it is a good example to test the robustness of the proposed method. The second pair of images is in Austin, USA, where many construction and terrain-related changes are present. The metrics used henceforth are the detection rate (DR) (recall) and the false alarm rate (FAR), which is defined as the ratio between the number of pixels misclassified as changes and the total number of unchanged pixels. These are metrics that properly quantify the trade-off between over and under detection.

\paragraph{Visual Evaluation}

\begin{figure*}[t]
\centering
\subfloat[Sentinel-II (Pre) + Planet (Post)]{
    \includegraphics[width=.7\linewidth]{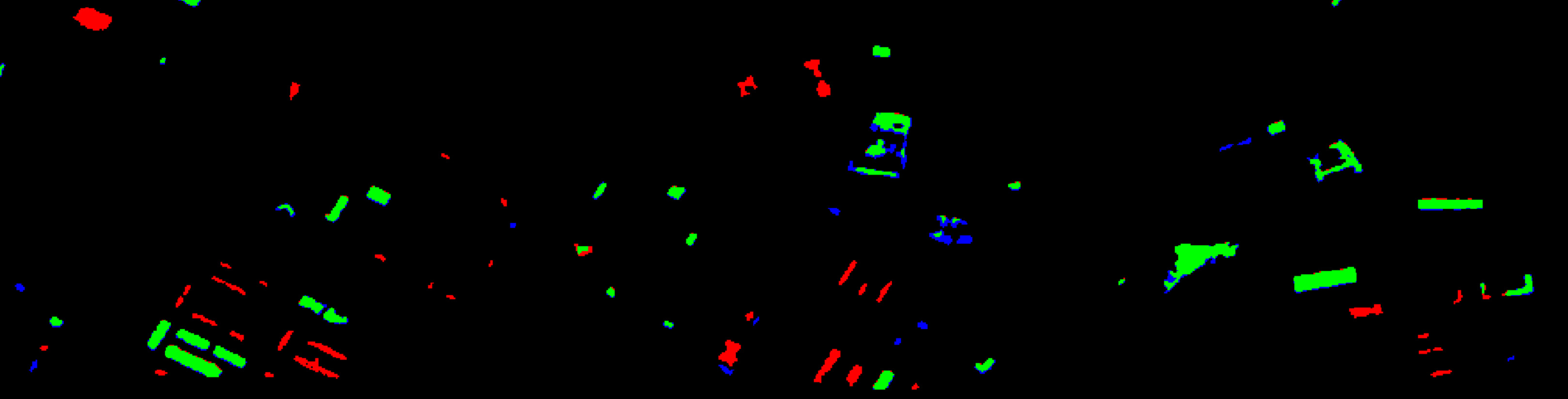}
    \label{fig:cd_austin:austin_s2planet}
}
\hfill
\subfloat[Regression + $\mathbf{W,C}$ Synthetic Planet (Pre) + Planet (Post)]{
    \includegraphics[width=.7\linewidth]{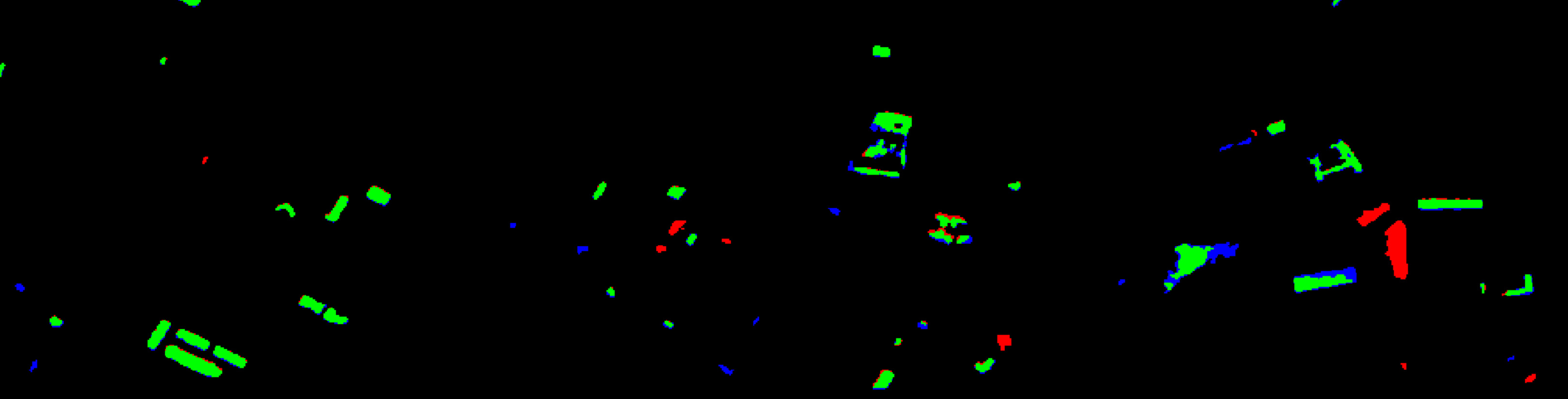}
    \label{fig:cd_austin:austin_regression}
}
\hfill
\subfloat[Proposed Synthetic Planet (Pre) + Planet (Post)]{
    \includegraphics[width=.7\linewidth]{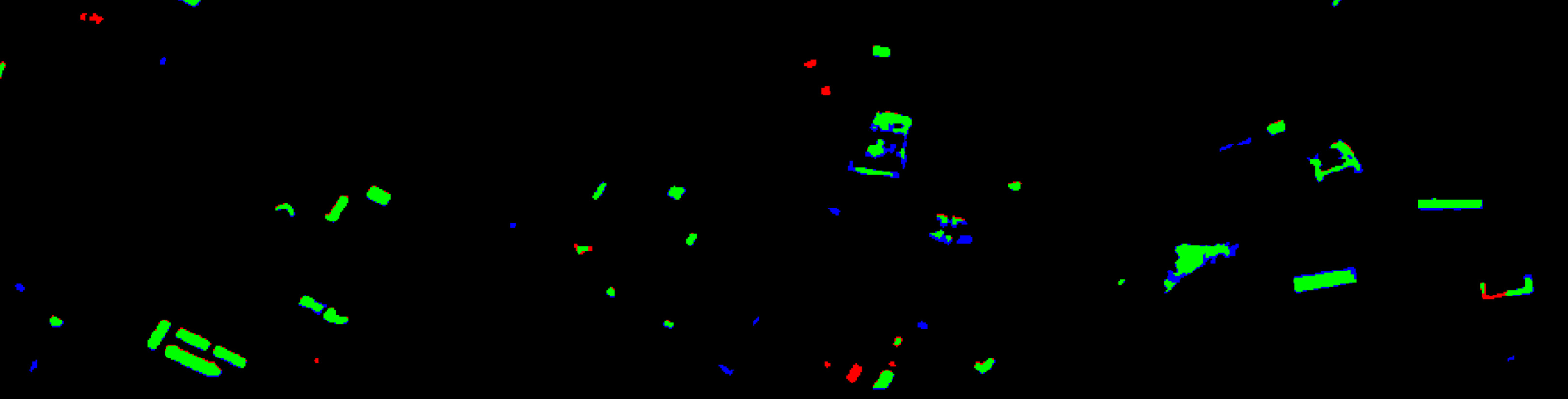}
    \label{fig:cd_austin:austin_ourmethod}
}
\hfill
\subfloat[Ground Truth---Planet Pre and Post]{
\includegraphics[width=.7\linewidth]{figs/austin/austin_cd_gt.jpg}
\label{fig:cd_austin:gt}
}
\caption{Change detection results in Austin Region for a fixed detection rate of 0.75. Pixels in green, blue, and red are true positives, false negatives, and false positives, respectively.}
\label{fig:cd_austin}
\end{figure*}

Using the images presented in Figures \ref{fig:genres} and \ref{fig:genres_austin}, quantitative change detection results were obtained and presented in Figures \ref{fig:cd_beirut} and \ref{fig:cd_austin} for Beirut and Austin, respectively.
Pixels in green correspond to true positives, in red to false positives, and in blue to false negatives. The first, second, and third rows individually contain change detection results generated using pre-event images: a Sentinel-II image (Figs. \ref{fig:cd_beirut:beirut_s2planet} and \ref{fig:cd_austin:austin_s2planet}), a synthetic Planet Dove image using the Regression + $\mathbf{W}, \mathbf{C}$ model (Figs. \ref{fig:cd_beirut:beirut_regression} and \ref{fig:cd_austin:austin_regression}), described in the previous section, and a synthetic Planet Dove image generated with the proposed DDM-based algorithm (Figs. \ref{fig:cd_beirut:beirut_ourmethod} and \ref{fig:cd_austin:austin_ourmethod}).
The Coloring procedure executed in the generation of the synthetic images made use of color information extracted from the post-Planet Dove image patches. 
A threshold $\omega$ has been chosen for each generated change map such that $\text{DR} = 0.75$, which is an operating point that highlights the effects of the proposed algorithm. 

\begin{figure*}
    \centering
    \subfloat[Beirut Port Region]{
    \includegraphics[width=.5\linewidth]{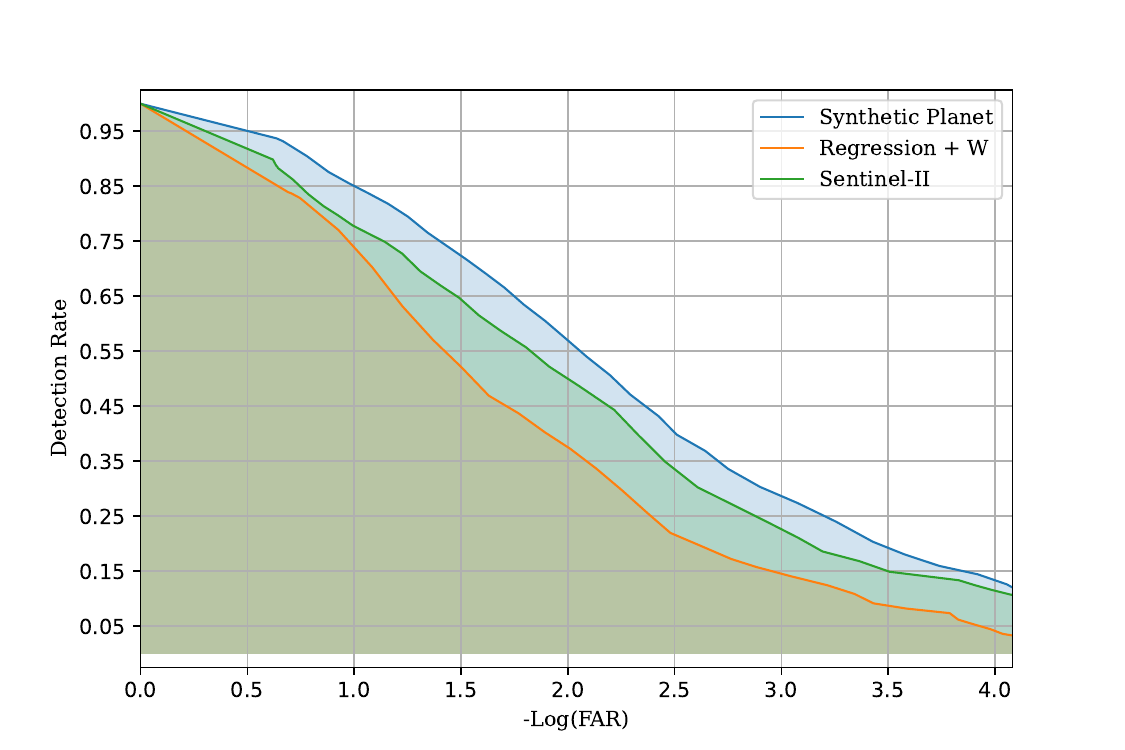}
    \label{fig:roc_cd:beirut}
    }
    \subfloat[Austin Region]{
    \includegraphics[width=.5\linewidth]{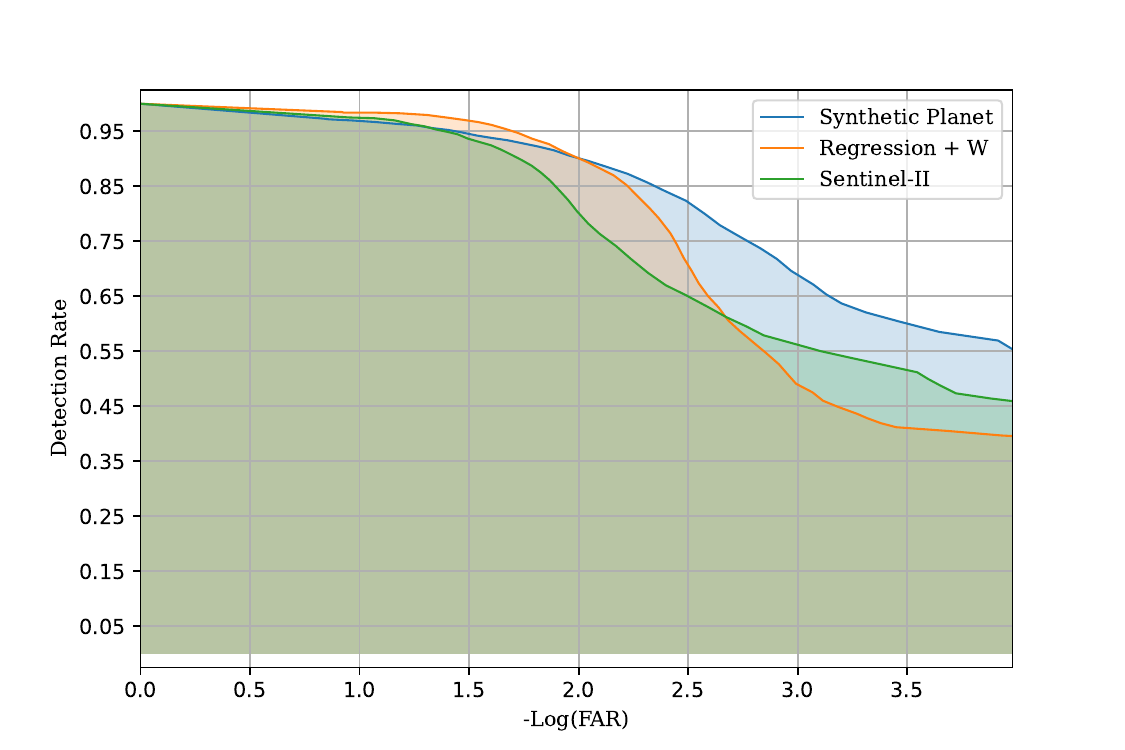}
    \label{fig:roc_cd:austin}
    }
    \caption{Performance comparison of heterogeneous change detection region using a Planet Dove frame as a post-event image. \textit{Sentinel-II} refers to when a Sentinel-II image is directly used as a pre-event image. \textit{Regression + W} refers to when using the regression-based model with the whitening operation to generate a synthetic pre-event image.
    \textit{Synthetic Planet} refers to when a synthetic Planet Dove image generated by the proposed I2I method is used as a pre-event image.}
    \label{fig:roc_cd}
\end{figure*}

Comparing the results in Fig. \ref{fig:cd_beirut}, the proposed method has been able to heavily suppress false alarms. In fact, the change maps for Sentinel-II and Regression + $\textbf{W}$ as pre-images show a staggering amount of false alarms, greatly surpassing the number of true positives. When the proposed method is applied, however, the majority of such false alarms are completely avoided. This visual superiority is further confirmed by the calculated metrics.
For the Beirut region in Fig. \ref{fig:cd_beirut}, with $\text{DR} = 0.75$, the proposed method reached a FAR of 0.03933 ($-log_{10}(\text{FAR})=-1.405$), whereas when using the Sentinel-II pre-event image, and the Regression + $\mathbf{W}$ synthetic pre-event image, the FAR was of 0.0731 ($-log_{10}(\text{FAR})=-1.136$) and 0.1055 ($-log_{10}(\text{FAR})=-0.9769$), respectively. 

For the Austin region in Fig. \ref{fig:cd_austin}, the improvements are less visually evident but still numerically expressive. In Fig. \ref{fig:cd_austin:austin_s2planet}, when the Sentinel-II image is used as pre-event image, multiple false alarms related to unchanged buildings can be spotted, specially in the bottom of the image, which are not observed in the change map for the proposed method's change map (Fig. \ref{fig:cd_austin:austin_ourmethod}). Moreover, for the regression-based change detection in Fig. \ref{fig:cd_austin:austin_regression}, a big false alarm is spotted in the eastern part of the image, which is not present in the change map of the proposed method. The metrics again prove the superiority of our technique:
for $\text{DR} = 0.75$, using the synthetic image generated by the proposed algorithm as the pre-event image leads to a FAR of 0.001678 ($-log_{10}(\text{FAR})=-2.775$), whereas when using Sentinel-II leads to a FAR of 0.007194 ($-log_{10}(\text{FAR})=-2.143$), and to a FAR of 0.003648 ($-log_{10}(\text{FAR})=-2.438$) when Regression + $\mathbf{W}$ is used.

\paragraph{Quantitative Evaluation}
To better evaluate the performance of the proposed algorithm for all possible choices of clipping threshold $\omega$, employed in Line \ref{alg:cd:threshold} of Algorithm \ref{alg:cd}, Figures \ref{fig:roc_cd:beirut} and \ref{fig:roc_cd:austin} present receiving operating characteristic (ROC) comparisons between the baseline change detection and change detection using the proposed method, for Beirut port and Austin regions, respectively. These curves were obtained by varying $\omega$ between 0 and 1.0, with steps of 0.01. As can be visualized in the curves, $\omega$ adjusts how permissive or restrictive should the algorithm be towards potential changes: a trade-off between higher detection rates and lower false alarm rates.

In Fig. \ref{fig:roc_cd:beirut}, the proposed method beats both compared methods for all choices of $\omega$, which displays how it has been able to significantly reduce false alarms for a desired detection rate. For fixed detection rates, the difference between the false alarm rate of the proposed method and the other two cases is up to 0.7 on the logarithmic scale or 5 times lower on the linear scale (DR=0.45). 

Meanwhile, in Fig. \ref{fig:roc_cd:austin}, the performance of the proposed method is on par with the other two methods up to $-log_{10}(\text{FAR}) = 2$, displaying a slightly lower DR compared to the regression-based model: a maximum difference of 0.01 in the DR. After this point, however, the proposed method beats by a large margin the other two approaches, reaching a FAR difference of up to 1.25 in the logarithmic scale, or 17.8 times lower in the linear scale.

\section{Discussion}
The results of our study demonstrate the efficacy of leveraging our proposed Denoising Diffusion method for large-scale optical image translation. Our approach successfully addresses the limitations of existing Image-to-Image (I2I) methods, particularly in handling large-scale multi-patch imagery. By focusing on super-resolving low spatial resolution images into high-resolution equivalents, our method achieves significant improvements in both radiometric accuracy and feature representation.

One of the key findings is the ability of our method to maintain uniformity across hundreds of patches, which is crucial for applications such as heterogeneous change detection (HCD). The comparative study against the standard classifier-free guided DDIM framework and other leading methods highlights the superior performance of our approach in terms of domain adaptation and artifact reduction. The HCD tasks in urban settings like Beirut, Lebanon, and Austin, USA, further validate the practical applicability of our method.

However, it is important to highlight that the simple change detection algorithm employed in our experiments does not fully exploit the enhanced visual resolution provided by our image translation method. This limitation is partly due to the algorithm’s design, which does not leverage the higher-resolution features effectively. Consequently, while our method shows promise, it does not necessarily outperform other methods, such as ShuffleMixer, SRDenseNet, and SwinIR, in HCD tasks using the chosen change detection algorithm. Despite their poor super-resolution performance, these methods maintain patch-wise feature consistency. The HCD experiments in this study are intended to demonstrate the potential utility of our approach rather than to establish it as the definitive solution for change detection. Future work will explore more advanced learning-based change detection procedures that can fully utilize the synthetic images generated by our method.

\color{dgreen}
As described in Section \ref{sec:pmet}, when a domain \textit{B} HR image is not available for color extraction during inference, it is possible to extract the needed radiometry information directly from the input LR domain \textit{A} to restore the overall colorization of the domain \textit{B} output. 
Given the overall radiometric shifts between these domains, relying on overall radiometry information from domain \textit{A} may result in synthetic images that may not perfectly represent the radiometry expected from the target domain. However, in cases where radiometry information from domain \textit{B} is not available, this compromise can still produce images that are adequate for most downstream tasks, seen that the radiometric shifts between domains are rarely  large enough to cause problems.


While our method effectively handles large-scale imagery, the computational complexity remains challenging. Future work will also explore optimization techniques to reduce the computational load without compromising the model’s performance. Additionally, extending our approach to other types of remote sensing data, such as radar or hyperspectral images, could broaden its applicability and impact. Notably, when the number of channels ($n_{\text{ch}}$) is reduced to one, as in panchromatic or single-band SAR images, the whitening and colorization processes become redundant due to the subsequent normalization. In such cases, an alternative approach would be to remove the $[-1, 1]$ normalization and instead normalize the values of all patches by dividing them by a fixed maximum value to keep them within a reasonable range. This would be followed by mean subtraction and posterior mean addition in the resulting outputs. We did not, however, performed these tests, as working with panchromatic images is out of the scope of the manuscript.
 
\color{black}

\section{Conclusion}
\color{dgreen}
We introduced an innovative deep learning-based method that employs denoising diffusion models to generate synthetic high-spatial-resolution satellite images from low-spatial-resolution inputs of a different sensor. Our method incorporates novel forward and reverse diffusion procedures, including color standardization and initial noise selection techniques, which align input and output patches and reduce model hallucinations and color inconsistencies. Trained and tested on a large and diverse dataset of paired images, our method demonstrated superior performance compared to existing solutions, including the popular classifier-free guided DDIM framework.

We also applied our method to the task of heterogeneous change detection between images from different sensors, demonstrating its effectiveness in two urban areas: Beirut, Lebanon, and Austin, USA. While our method effectively handles large-scale imagery and maintains uniformity across patches, certain limitations remain---such as when color information from the target domain is unavailable during inference, and the high computational inference complexity.
To enhance the robustness of our approach, future work will focus on addressing these limitations. We plan to extend our method to other combinations of sensors and investigate the use of synthetic images in target change detection. 

In summary, by addressing the limitations of existing methods and by providing a foundation for future advancements, our work contributes to the progression of remote sensing technologies and their practical applications across various domains.
\color{black}
\FloatBarrier
\printbibliography

@article{Izmailov2018,
      title={{Averaging Weights Leads to Wider Optima and Better Generalization}}, 
      author={Pavel Izmailov and Dmitrii Podoprikhin and Timur Garipov and Dmitry Vetrov and Andrew Gordon Wilson},
      year={2019},
      journal={{arXiv}},
}

@article{Park2019,
      title={{Semantic Image Synthesis with Spatially-Adaptive Normalization}}, 
      author={Taesung Park and Ming-Yu Liu and Ting-Chun Wang and Jun-Yan Zhu},
      year={2019},
      journal={{arXiv}},
}

@article{Wang2022,
      title={{Semantic Image Synthesis via Diffusion Models}}, 
      author={Weilun Wang and Jianmin Bao and Wengang Zhou and Dongdong Chen and Dong Chen and Lu Yuan and Houqiang Li},
      year={2022},
      journal={{arXiv}},
}

@article{Song2020,
      title={{Denoising Diffusion Implicit Models}}, 
      author={{Jiaming Song and Chenlin Meng and Stefano Ermon}},
      year={2022},
      journal={{arXiv}},
}

@article{Wu2022,
   author = {Yue Wu and Jiaheng Li and Yongzhe Yuan and A. K. Qin and Qi Guang Miao and Mao Guo Gong},
   doi = {10.1109/TNNLS.2021.3056238},
   issue = {9},
   journal = {IEEE Transactions on Neural Networks and Learning Systems},
   month = {9},
   pages = {4257-4270},
   pmid = {33600325},
   publisher = {Institute of Electrical and Electronics Engineers Inc.},
   title = {{Commonality Autoencoder: Learning Common Features for Change Detection From Heterogeneous Images}},
   volume = {33},
   year = {2022},
}

@article{Touati2020,
   author = {Redha Touati and Max Mignotte and Mohamed Dahmane},
   doi = {10.1109/JSTARS.2020.2964409},
   journal = {IEEE Journal of Selected Topics in Applied Earth Observations and Remote Sensing},
   pages = {588-600},
   publisher = {Institute of Electrical and Electronics Engineers},
   title = {{Anomaly Feature Learning for Unsupervised Change Detection in Heterogeneous Images: A Deep Sparse Residual Model}},
   volume = {13},
   year = {2020},
}

@article{I2ISeo2023,
      title={{Improved Flood Insights: Diffusion-Based SAR to EO Image Translation}}, 
      author={Minseok Seo and Youngtack Oh and Doyi Kim and Dongmin Kang and Yeji Choi},
      year={2023},
      journal={{arXiv}},
}

@article{Nichol2021,
      title={{Improved Denoising Diffusion Probabilistic Models}}, 
      author={Alex Nichol and Prafulla Dhariwal},
      year={2021},
      journal={{arXiv}},
}

@article{Zhang2022,
   author = {Chenxiao Zhang and Yukang Feng and Lei Hu and Deodato Tapete and Li Pan and Zheheng Liang and Francesca Cigna and Peng Yue},
   doi = {10.1016/j.jag.2022.102769},
   journal = {International Journal of Applied Earth Observation and Geoinformation},
   month = {5},
   publisher = {Elsevier B.V.},
   title = {{A domain adaptation neural network for change detection with heterogeneous optical and SAR remote sensing images}},
   volume = {109},
   year = {2022},
}

@article{Lv2022,
   author = {Zhi Yong Lv and Hai Tao Huang and Xinghua Li and Ming Hua Zhao and Jon Atli Benediktsson and Wei Wei Sun and Nicola Falco},
   doi = {10.1109/JPROC.2022.3219376},
   issue = {12},
   journal = {Proceedings of the IEEE},
   month = {12},
   pages = {1976-1991},
   publisher = {Institute of Electrical and Electronics Engineers Inc.},
   title = {{Land Cover Change Detection with Heterogeneous Remote Sensing Images: Review, Progress, and Perspective}},
   volume = {110},
   year = {2022},
}

@article{Dhariwal2021,
      title={{Diffusion Models Beat GANs on Image Synthesis}}, 
      author={Prafulla Dhariwal and Alex Nichol},
      year={2021},
      journal={{arXiv}},
}

@article{Saharia2021,
      title={{Palette: Image-to-Image Diffusion Models}}, 
      author={Chitwan Saharia and William Chan and Huiwen Chang and Chris A. Lee and Jonathan Ho and Tim Salimans and David J. Fleet and Mohammad Norouzi},
      year={2022},
      journal={{arXiv}}
}

@article{CDLi2021,
   author = {Xinghua Li and Zhengshun Du and Yanyuan Huang and Zhenyu Tan},
   doi = {10.1016/j.isprsjprs.2021.07.007},
   journal = {ISPRS Journal of Photogrammetry and Remote Sensing},
   pages = {14-34},
   publisher = {Elsevier B.V.},
   title = {{A deep translation (GAN) based change detection network for optical and SAR remote sensing images}},
   volume = {179},
   year = {2021},
}

@article{WangCD2022,
      title={{CD-GAN: a robust fusion-based generative adversarial network for unsupervised remote sensing change detection with heterogeneous sensors}}, 
      author={Jin-Ju Wang and Nicolas Dobigeon and Marie Chabert and Ding-Cheng Wang and Ting-Zhu Huang and Jie Huang},
      year={2023},
      journal={{arXiv}},
}

@article{Ho2020,
      title={{Denoising Diffusion Probabilistic Models}}, 
      author={Jonathan Ho and Ajay Jain and Pieter Abbeel},
      year={2020},
      journal={{arXiv}},
}

@article{mansourifar_i3eaccess2022,
  author={Mansourifar, Hadi and Moskovitz, Alexander and Klingensmith, Ben and Mintas, Dino and Simske, Steven J.},
  journal={IEEE Access}, 
  title={{GAN-Based Satellite Imaging: A Survey on Techniques and Applications}}, 
  year={2022},
  volume={10},
  number={},
  pages={118123-118140},
  doi={10.1109/ACCESS.2022.3221123}}

@article{ho2022classifierfree,
      title={{Classifier-Free Diffusion Guidance}}, 
      author={Jonathan Ho and Tim Salimans},
      year={2022},
      journal={{arXiv}},
}

@article{Daras2023,
      title={{Consistent Diffusion Models: Mitigating Sampling Drift by Learning to be Consistent}}, 
      author={Giannis Daras and Yuval Dagan and Alexandros G. Dimakis and Constantinos Daskalakis},
      year={2023},
      journal={{arXiv}},
}

@article{Hang2023,
      title={{Efficient Diffusion Training via Min-SNR Weighting Strategy}}, 
      author={Tiankai Hang and Shuyang Gu and Chen Li and Jianmin Bao and Dong Chen and Han Hu and Xin Geng and Baining Guo},
      year={2023},
      journal={{arXiv}},
}

@article{Saharia2021SR3,
      title={{Image Super-Resolution via Iterative Refinement}}, 
      author={Chitwan Saharia and Jonathan Ho and William Chan and Tim Salimans and David J. Fleet and Mohammad Norouzi},
      year={2021},
      journal={{arXiv}},
}

@article{liu2021variance,
      title={{On the Variance of the Adaptive Learning Rate and Beyond}}, 
      author={Liyuan Liu and Haoming Jiang and Pengcheng He and Weizhu Chen and Xiaodong Liu and Jianfeng Gao and Jiawei Han},
      year={2021},
      journal={{arXiv}},
}

@article{meyer2020alternative,
      title={{An Alternative Probabilistic Interpretation of the Huber Loss}}, 
      author={Gregory P. Meyer},
      year={2020},
      journal={{arXiv}},
}

@inproceedings{Heusel2017,
 author = {Heusel, Martin and Ramsauer, Hubert and Unterthiner, Thomas and Nessler, Bernhard and Hochreiter, Sepp},
 booktitle = {Advances in Neural Information Processing Systems},
 editor = {I. Guyon and U. Von Luxburg and S. Bengio and H. Wallach and R. Fergus and S. Vishwanathan and R. Garnett},
 publisher = {Curran Associates, Inc.},
 title = {{GANs Trained by a Two Time-Scale Update Rule Converge to a Local Nash Equilibrium}},
 volume = {30},
 year = {2017}
}

@article{garipov2018loss,
      title={{Loss Surfaces, Mode Connectivity, and Fast Ensembling of DNNs}}, 
      author={Timur Garipov and Pavel Izmailov and Dmitrii Podoprikhin and Dmitry Vetrov and Andrew Gordon Wilson},
      year={2018},
      journal={{arXiv}},
}

@article{Zhang2018LPIPS,
      title={{The Unreasonable Effectiveness of Deep Features as a Perceptual Metric}}, 
      author={Richard Zhang and Phillip Isola and Alexei A. Efros and Eli Shechtman and Oliver Wang},
      year={2018},
      journal={{arXiv}},
}

@inproceedings{Hore2010,
  author={Horé, Alain and Ziou, Djemel},
  booktitle={{20th International Conference on Pattern Recognition}}, 
  title={{Image Quality Metrics: PSNR vs. SSIM}}, 
  year={2010},
  volume={},
  number={},
  pages={2366-2369},
  doi={10.1109/ICPR.2010.579}}

@ARTICLE{Crapsu1979,
  author={Otsu, Nobuyuki},
  journal={IEEE Transactions on Systems, Man, and Cybernetics}, 
  title={{A Threshold Selection Method from Gray-Level Histograms}}, 
  year={1979},
  volume={9},
  number={1},
  pages={62-66},
  doi={10.1109/TSMC.1979.4310076}}

@article{I2IIsmael2023,
   author = {Sarmad F. Ismael and Koray Kayabol and Erchan Aptoula},
   doi = {10.1109/LGRS.2023.3281458},
   journal = {IEEE Geoscience and Remote Sensing Letters},
   publisher = {Institute of Electrical and Electronics Engineers Inc.},
   title = {{Unsupervised Domain Adaptation for the Semantic Segmentation of Remote Sensing Images via One-Shot Image-to-Image Translation}},
   volume = {20},
   year = {2023},
}

@inproceedings{I2ITasar2020,
   author = {Onur Tasar and S. L. Happy and Yuliya Tarabalka and Pierre Alliez},
   doi = {10.1109/IGARSS39084.2020.9323711},
   isbn = {9781728163741},
   publisher = {2020 International Geoscience and Remote Sensing Symposium (IGARSS) },
   month = {9},
   pages = {1837-1840},
   title = {{SEMI2I: Semantically Consistent Image-to-Image Translation for Domain Adaptation of Remote Sensing Data}},
   year = {2020},
}

@article{I2ISokolov2023,
   author = {Mikhail Sokolov and Christopher Henry and Joni Storie and Christopher Storie and Victor Alhassan and Mathieu Turgeon-Pelchat},
   doi = {10.1109/JSTARS.2022.3226705},
   journal = {IEEE Journal of Selected Topics in Applied Earth Observations and Remote Sensing},
   pages = {482-492},
   publisher = {Institute of Electrical and Electronics Engineers Inc.},
   title = {{High-Resolution Semantically Consistent Image-to-Image Translation}},
   volume = {16},
   year = {2023},
}

@article{I2IZhang2023,
      title={{SAR-to-Optical Image Translation via Thermodynamics-inspired Network}}, 
      author={Mingjin Zhang and Jiamin Xu and Chengyu He and Wenteng Shang and Yunsong Li and Xinbo Gao},
      year={2023},
    publisher={{arXiv}}
}

@article{Xiao2023,
   author = {Yi Xiao and Qiangqiang Yuan and Kui Jiang and Jiang He and Xianyu Jin and Liangpei Zhang},
   doi = {10.1109/TGRS.2023.3341437},
   journal = {IEEE Transactions on Geoscience and Remote Sensing},
   publisher = {Institute of Electrical and Electronics Engineers Inc.},
   title = {{EDiffSR: An Efficient Diffusion Probabilistic Model for Remote Sensing Image Super-Resolution}},
   year = {2023},
}

@article{Lathuiliere2020,
   title={A Comprehensive Analysis of Deep Regression},
   DOI={10.1109/tpami.2019.2910523},
   number={9},
   journal={IEEE Transactions on Pattern Analysis and Machine Intelligence},
   publisher={Institute of Electrical and Electronics Engineers (IEEE)},
   author={Lathuiliere, Stephane and Mesejo, Pablo and Alameda-Pineda, Xavier and Horaud, Radu},
   year={2020},
   month=09, pages={2065–2081} }

@inproceedings{DBSCAN1996,
author = {Ester, Martin and Kriegel, Hans-Peter and Sander, J\"{o}rg and Xu, Xiaowei},
title = {{A density-based algorithm for discovering clusters in large spatial databases with noise}},
year = {1996},
publisher = {AAAI Press},
booktitle = {Proceedings of the Second International Conference on Knowledge Discovery and Data Mining},
pages = {226–231},
numpages = {6},
location = {Portland, Oregon},
series = {KDD'96}
}

@article{Chini2023,
title = {Fourier domain structural relationship analysis for unsupervised multimodal change detection},
journal = {ISPRS Journal of Photogrammetry and Remote Sensing},
volume = {198},
pages = {99-114},
year = {2023},
issn = {0924-2716},
doi = {10.1016/j.isprsjprs.2023.03.004},
author = {Hongruixuan Chen and Naoto Yokoya and Marco Chini},
}

@INPROCEEDINGS{Pix2Pix2017,
  author={Isola, Phillip and Zhu, Jun-Yan and Zhou, Tinghui and Efros, Alexei A.},
  booktitle={2017 IEEE Conference on Computer Vision and Pattern Recognition (CVPR)}, 
  title={Image-to-Image Translation with Conditional Adversarial Networks}, 
  year={2017},
  pages={5967-5976},
  doi={10.1109/CVPR.2017.632}}

@article{luo_understanding_2022,
	title = {Understanding Diffusion Models: A Unified Perspective},
	journal = {{arXiv}},
	author = {Luo, Calvin},
	year = {2022},
}

@article{liang2021swinir,
    title={{SwinIR: Image Restoration Using Swin Transformer}}, 
    author={Jingyun Liang and Jiezhang Cao and Guolei Sun and Kai Zhang and Luc Van Gool and Radu Timofte},
    year={2021},
    journal = {{arXiv}},
}

@article{sun2022shufflemixer,
      title={{ShuffleMixer: An Efficient ConvNet for Image Super-Resolution}}, 
      author={Long Sun and Jinshan Pan and Jinhui Tang},
      year={2022},
      journal={{arXiv}},
      primaryClass={cs.CV}
}

@article{zhang2018residual,
      title={{Residual Dense Network for Image Super-Resolution}}, 
      author={Yulun Zhang and Yapeng Tian and Yu Kong and Bineng Zhong and Yun Fu},
      year={2018},
      journal={{arXiv}},
}

@Article{CuiColor2020,
AUTHOR = {Cui, H. and Zhang, G.},
TITLE = {HIGH-RESOLUTION OPTICAL SATELLITE IMAGES COLOR CONSISTENCY METHOD BASED ON EXTERNAL COLOR REFERENCES},
JOURNAL = {ISPRS Annals of the Photogrammetry, Remote Sensing and Spatial Information Sciences},
VOLUME = {V-3-2020},
YEAR = {2020},
PAGES = {663--668},
DOI = {10.5194/isprs-annals-V-3-2020-663-2020}
}

@article{deck2023easing,
      title={Easing Color Shifts in Score-Based Diffusion Models}, 
      author={Katherine Deck and Tobias Bischoff},
      year={2023},
      journal={{arXiv}},
}

@article{wan_psc_2024,
	title = {{PSC} diffusion: patch-based simplified conditional diffusion model for low-light image enhancement},
	volume = {30},
	issn = {0942-4962, 1432-1882},
	doi = {10.1007/s00530-024-01391-z},
	shorttitle = {{PSC} diffusion},
	pages = {187},
	number = {4},
	journaltitle = {Multimedia Systems},
	shortjournal = {Multimedia Systems},
	author = {Wan, Fei and Xu, Bingxin and Pan, Weiguo and Liu, Hongzhe},
	urldate = {2024-09-05},
	date = {2024-08},
	langid = {english},

}

@article{yi_tip2024,
author = {Xiao, Yi and Yuan, Qiangqiang and Jiang, Kui and He, Jiang and Lin, Chia-Wen and Zhang, Liangpei},
year = {2024},
month = {01},
pages = {738-752},
title = {{TTST: A Top-k Token Selective Transformer for Remote Sensing Image Super-Resolution}},
volume = {33},
journal = {IEEE Transactions on Image Processing},
doi = {10.1109/TIP.2023.3349004}
}

@article{Min2024,
  author={Min, Jeongho and Lee, Yejun and Kim, Dongyoung and Yoo, Jaejun},
  journal={{IEEE Geoscience and Remote Sensing Letters}}, 
  title={{Bridging the Domain Gap: A Simple Domain Matching Method for Reference-Based Image Super-Resolution in Remote Sensing}}, 
  year={2024},
  volume={21},
  number={},
  pages={1-5},
  doi={10.1109/LGRS.2023.3336680}}

@article{Haut2018,
  author={Haut, Juan Mario and Fernandez-Beltran, Ruben and Paoletti, Mercedes E. and Plaza, Javier and Plaza, Antonio and Pla, Filiberto},
  journal={{IEEE Transactions on Geoscience and Remote Sensing}}, 
  title={{A New Deep Generative Network for Unsupervised Remote Sensing Single-Image Super-Resolution}}, 
  year={2018},
  volume={56},
  number={11},
  pages={6792-6810},
  doi={10.1109/TGRS.2018.2843525}}
\begin{appendices}
\section{Implementation Details}
\label{sec:apx:model_s}
\begin{table*}[t]
\setstretch{1.25}
\centering
\caption{Hyperparameter values and their purpose for the performed experiments.}
\label{tab:hyper}
\begin{tabularx}{\linewidth}{llX}
\hline
\hline
\textbf{Parameter}  & \textbf{Value} & \textbf{Purpose}  \\
\hline
Patch Size & $128\times 128$ & Width $w$ and height $h$ of the image patches fed to the neural network.          \\
$N_{\text{final}}$ & 64 & Number of DDIM iterations for the final reverse diffusion in Line \ref{alg:inference:y_0_hat} of Alg. \ref{alg:inference}.\\
$N_{\text{pre}}$ & 8 & Number of DDIM iterations for the PSNR voting procedure in Line \ref{alg:inference:eps_t_best} of Alg. \ref{alg:inference}. \\
$n_{\text{noisy}}$ & 8 & Number of DDIM runs for the estimation of $\bm{\varepsilon}_T^{\text{best}}$, used in Line \ref{alg:inference:Y_0_hat_lq} of Alg. \ref{alg:inference}.\\
$\bm{\gamma}$ & $ \left[\cos\left(\frac{t_i/ T + 0.008}{1.008 }\frac{\pi}{2}\right)^ 2\right]_{t_i}$ & Diffusion noise level parameter for each timestep $t_i \in \{0, 1,\dots, T-1\}$ \cite{Nichol2021}. \\
$T$ & $1024$ & Number of forward diffusion noising steps.\\
$p_{\text{uncond}}$ & $10^{-1}$ & Classifier-free guidance unconditional diffusion probability \cite{ho2022classifierfree}.\\
$\omega_{\text{uncond}}$ & $1$ & Classifier-free guidance inference weighting parameter \cite{ho2022classifierfree}.\\
$\lambda_{\text{consist}}$ & $10^{-1}$ & Weight of the consistency loss over the final loss \cite{Daras2023}.\\
$\varepsilon_{\text{consist}}$ & $10$ & It tells how far from the sampled timestep $t$ can the consistency timestep $t^\prime$ be sampled \cite{Daras2023}. That is, $t^\prime \in \{t-\varepsilon_{\text{consist}}, \dots, t\}$. \\
$n_{\text{consist}}$ & $1$ & Number of consistency training steps to be executed in a single overall training step \cite{Daras2023}.\\
$\gamma_{\text{SNR}}$ & $5$ & Clamping parameter for the Min-SNR-$\gamma$ training weighting strategy.\\
\hline
\hline
\end{tabularx}
\end{table*}
SDM \cite{Wang2022}, the chosen backbone for the proposed method's experiments, is a U-Net-style model that uses attention mechanisms to improve image fidelity. Differently from the diffusion-based models presented in \cite{Saharia2021, Saharia2021SR3, ho2022classifierfree}, this model processes the noisy input $\bm{y}_t$ and the conditioning input $\bm{x}$ independently. That is $\bm{y}_t$ is fed right into the encoder and is processed normally, whereas $\bm{x}$ is re-introduced into the decoder by inserting it into each residual block. To properly match the spatial resolution required by each residual block, $\bm{x}$ is processed by spatially adaptive normalization (SPADE) blocks. 
As confirmed by the performed experiments, we observed that this modification improved significantly the generation quality and fidelity to the conditioning input. Some internal parameters of the model were adjusted to maximize performance while limiting its size due to hardware constraints. The number of channels has been set to be a multiple of $96$. The number of residual blocks for each downsampling operation has been set to $2$. Attention resolutions were set to $2, 4, 8$ and $16$. Finally, the number of attention heads has been set to $4$. This configuration resulted in a model with around 195 million trainable parameters.
For training, batch size has been set to $5$, and the learning rate has been set to $10^{-4}$. 

We use the Huber loss function \cite{meyer2020alternative} with its parameter $\delta$ set to $0.5$ since it has been shown to reduce the impact of outliers in the gradients while maintaining the benefits of the mean square distance for non-outliers. Implementation details for the Huber loss are presented in Appendix \ref{sec:apx:to}. RAdam optimizer \cite{liu2021variance} has been chosen due to its proven training speed and its low sensitivity to the learning rate choice. The training ran for $31$ epochs, which was more than enough for the loss function to converge completely. 

To stabilize training and avoid catastrophic forgetting of features, Exponential Moving Average (EMA) has been used in several DDM-based methods \cite{Ho2020, Saharia2021, Wang2022, Dhariwal2021}. With the same goal, we instead use Stochastic Weight Averaging (SWA) \cite{Izmailov2018}, which consists of averaging weights after the training loss shows convergence behavior. Differently from EMA, SWA does not give higher importance to the weights of more recent epochs but instead considers all epochs after the beginning of weight averaging to have the same importance for overall learning. Since we have observed that the loss function converged after a number of epochs, it was reasonable to consider that newer weight updates would not bring an overall improvement of the model, thus applying SWA instead of EMA made more sense.
The work presented in \cite{garipov2018loss} suggests that averaging weights of different epochs can indeed result in higher performance than when only using the last weights of a training experiment. We observed that training stabilized after 10 epochs, so we started the SWA process at this moment. Also, SWA suggests increasing the learning rate during the epochs where weight averaging takes place. However, we noticed that increasing the learning rate over $10^{-4}$ made training unstable. For this reason, it is kept constant during the whole process. 

For the performed experiments, multiple hyperparameters not directly related to the model setup have been chosen. They are presented in Table \ref{tab:hyper}, along with their purpose. Some choices were based on a grid search, and others were based on the articles that defined them in the first place. Moreover, two recently proposed training techniques for denoising diffusion models were included in our experiments: Consistency Diffusion Loss and Min-SNR-$\gamma$ Weighting. They are further discussed in Appendices \ref{sec:apx:cdm} and \ref{sec:apx:minsnr}. The training and experiments were performed using one NVIDIA RTX A6000 GPU.

\subsection{Model Pruning}
To elucidate the model's performance in relation to its size, we have included Table \ref{tab:comp_prune}, which presents the effects of various unstructured pruning rates. The experiments, executed with parameters \(N_{\text{final}}=16\), \(N_{\text{pre}}=2\), and \(n_{\text{noisy}}=2\), demonstrate that a light pruning—specifically removing 1/8 of the weights by magnitude—significantly improved the model's performance metrics on the test set used in the Experiments section. This enhancement suggests that such pruning can mitigate overfitting, streamlining the model to better generalize under different data regimes.

However, applying more aggressive pruning rates of 1/4 and 1/2 led to a considerable degradation in performance, indicating that such high levels of pruning do not preserve the model's ability to perform effectively. This outcome suggests that there is a threshold beyond which pruning negatively impacts the model's capacity to learn and generalize.

Regarding numerical efficiency, despite the model being converted into a sparse format in PyTorch post-pruning with the intention of exploiting sparsity for computational speedups, no such acceleration has been observed. The expected efficiency gains have not materialized, possibly due to the hardware's inability to fully exploit the model's sparsity properties. Another possibility is that the sparsity of the pruned models is not sufficiently large to result in noticeable speedups.

\subsection{Discussion on the Selection of Hyperparameters}

In response to a possible question on the absence of ablation studies for the network parameters, it is important to note that the original paper presenting the network did not perform such studies either. The reason for this is twofold: firstly, diffusion models require extensive computational resources and time to train, often taking several days, even on high-end hardware. In our case, the hardware constraints were even more pronounced, which made the training process particularly challenging. Secondly, the potential gains from conducting ablation studies on different hyperparameters are not evidently clear. Diffusion models are complex, and their performance is not solely dependent on individual hyperparameters but rather on the intricate interplay between them. As such, varying single parameters may not result in significant changes in performance, making the exhaustive process of ablation less justifiable.

Furthermore, the chosen hyperparameters were set based on a combination of grid search results and recommendations from the literature, ensuring a well-informed starting point for our experiments. The internal parameters were carefully adjusted to balance performance optimization with the limitations imposed by our hardware. For instance, the number of channels, residual blocks, attention resolutions, and attention heads were all set to values that would maximize the model’s efficiency without exceeding our computational budget.

The training process itself was meticulously monitored, with the batch size and learning rate being set to values that allowed for stable and complete convergence of the loss function within 31 epochs. The use of the Huber loss function and the RAdam optimizer further contributed to the robustness and speed of the training, as detailed in Appendix \ref{sec:apx:to}.

In light of these considerations, we believe that the chosen approach for parameter selection and training was the most appropriate given the constraints and the nature of diffusion models. The inclusion of novel training techniques such as Consistency Diffusion Loss and \mbox{Min-SNR-$\gamma$} Weighting further underscores our commitment to leveraging the latest advancements in the field to enhance our model’s performance.

\begin{table}[t]
\begingroup
\setstretch{1.25}
\centering
\caption{Model Performance Under Different Pruning Rates for $N_{\text{final}}=16$, $N_{\text{pre}}=2$, $n_{\text{noisy}}=2$}
\label{tab:comp_prune}
\begin{tabularx}{\linewidth}{YYYY}
\hline
\hline
Rate & \textbf{mLPIPS}$\downarrow$ & \textbf{FID}$\downarrow$ & \textbf{mPSNR}$\uparrow$  \\
\hline
No Pruning & 0.2229 & 52.46 & 14.60 \\
\textbf{1/8} & \textbf{0.2007} & \textbf{45.50} & \textbf{14.81}\\
1/4 & 0.2636 & 66.33 & 13.85 \\
1/2 & 0.4325 & 245.1 & 10.47\\
\hline
\hline
\end{tabularx}
\endgroup

\end{table} 
\section{Fundamental Concepts}\label{sec:apx:fund}
\subsection{Conditional Denoising Diffusion Probabilistic Models}
\label{sec:apx:dif}
\subsubsection{Forward Diffusion}
\label{sec:apx:fdif}
Let $\bm{y}_0$ be a data matrix sampled from $q(\bm{y}_0)$. The forward diffusion process takes place by progressively adding Gaussian noise to $\bm{y}_0$. It is a Markov chain with $T$ transitions, or steps, described by the following conditional probability functions:
\begin{gather}
    q(\bm{y}_{1:T}|\bm{y}_0) = \prod_{t=1}^{T} q(\bm{y}_{t}|\bm{y}_{t-1})\\
    q(\bm{y}_{t}|\bm{y}_{t-1}) = \mathcal{N} \left(\sqrt{\alpha_t} \bm{y}_{t-1}, (1 - \alpha_{t})\mathbf{I}\right)\text{.}
\end{gather}
It can be marginalized for each step as:
\begin{gather}
    q(\bm{y}_{t}|\bm{y}_0) :=  \int q(\bm{y}_{1:t}|\bm{y}_0) \: \mathrm{\mathbf{d}}\bm{y}_{1:(t-1)} \\ = \mathcal{N} \left(\sqrt{\gamma_t} \bm{y}_{0}, (1 - \gamma_{t})\mathbf{I}\right) \label{eq:diff:yty0}
\end{gather}
where $\gamma_t \in (0, 1)$ is the noise schedule parameter at time $t$, and $\alpha_t = \gamma_t /\gamma_{t-1}$. 

\subsubsection{Reverse Diffusion}
\label{sec:apx:revdif}
Applying Bayes', the posterior probability $q(\bm{y}_{t-1}|\bm{y}_0, \bm{y}_t)$ can be expressed as \cite{luo_understanding_2022}:
\begin{gather}
    q(\bm{y}_{t-1}|\bm{y}_0, \bm{y}_t) = \mathcal{N} (\bm{\mu}_q, \bm{\Sigma}_q)\label{eq:diff:ytm1y0yt} \\
    \bm{\mu}_q = \frac{1}{1-\gamma_t} \left[ \sqrt{\gamma_{t-1}} (1 - \alpha_t)\bm{y}_0 + \sqrt{\alpha_t}(1-\gamma_{t-1})\bm{y}_t \right] \\
    \bm{\Sigma}_q = \dfrac{(1-\gamma_{t-1})(1-\alpha_t)}{1-\gamma_t}\text{.}
\end{gather}
This means that it is possible to draw a sample from a previous diffusion step $t-1$ if $\bm{y}_t$ and $\bm{y}_0$ are known. Since the denoised data $\bm{y}_0$ is not known---as it is exactly what we want to retrieve at the end of the process---a neural network $\bm{\theta}_{t}$ can learn to predict an approximation of $\bm{y}_0$, $\hat{\bm{y}}_0$, given $\bm{y}_t$ and $\gamma_t$. In the context of conditional models, a conditional input $\bm{x}$ is also fed to the neural network. 

From Eq. \ref{eq:diff:yty0},
\begin{gather}
    \bm{y}_t = \sqrt{\gamma_t} \bm{y}_0 + \sqrt{1-\gamma_t} \bm{\epsilon}_t \\
    \text{where }\bm{\epsilon}_t \sim \mathcal{N}(0, \mathbf{I})\text{.}
\end{gather}
Thus, $\bm{y}_0$ can be parameterized as a linear combination between $\bm{y}_t$ and $\bm{\epsilon}_t$:
\begin{gather}
    \bm{y}_0 = \frac{1}{\sqrt{\gamma_t}} \left( \bm{y}_t - \sqrt{1-\gamma_t} \bm{\epsilon}_t \right)\text{.} \label{eq:diff:y0}
\end{gather}
This parametrization is useful since Ho et al. \cite{Ho2020} verified experimentally that higher quality approximations of $\bm{y}_0$ are obtained if the neural network is trained to instead predict $\bm{\epsilon}_t$, i.e., $\bm{\theta}_t(\bm{y}_t, \gamma_t, \bm{x}) = \hat{\bm{\epsilon}}_t$. This prediction can be plugged into Eq. \ref{eq:diff:y0} to obtain $\hat{\bm{y}}_0$. Finally, with $\bm{y}_t$ and $\hat{\bm{y}}_0$, an approximation of $\bm{y}_{t-1}$ is obtained by sampling from Eq. \ref{eq:diff:ytm1y0yt}. Combining Eqs. \ref{eq:diff:ytm1y0yt} and \ref{eq:diff:y0}, such sampling procedure can be expressed as:
\begin{gather}
    \hat{\bm{y}}_{t-1} = \frac{1}{\sqrt{\alpha_t}}\left[ \bm{y}_t - \frac{1-\alpha_t}{\sqrt{1-\gamma_t}} \bm{\theta}_t(\bm{y}_t, \gamma_t, \bm{x})\right] \notag\\ + \eta\sqrt{\dfrac{(1-\gamma_{t-1})(1-\alpha_t)}{1-\gamma_t}}\bm{z} \\
    \text{where } \bm{z} \sim \mathcal{N}(0, \mathbf{I})\\
    \text{and } \eta \in [0, 1]\text{.} \label{eq:diff:ytm1hat}
\end{gather}
By repeating the reverse sampling until $t=1$, a higher quality $\hat{y}_0$ can be obtained.

\subsubsection{Training Objective}
\label{sec:apx:to}
As discussed, the neural network should be trained to predict $\bm{\epsilon}_t$.
Typically, the Mean Squared Error (MSE) serves as the default loss function \cite{Ho2020}. However, it is highly sensitive to outliers, which can negatively affect the loss. To mitigate this, Huber Loss can be used, as it operates like MSE loss for regular data but switches to L1 loss for outliers.

Let $\delta \in \mathds{R}_{++}$ be the thresholding parameter. The training objective for $\bm{\theta}$ at timestep $t$ is defined as:
\begin{gather}
    \mathcal{L}_{\delta}(\bm{\theta}_t(\bm{y}_t, \gamma_t, \bm{x}), \bm{\epsilon}_t) = \notag\\\begin{cases}
\frac{1}{2}(\bm{\theta}_t(\bm{y}_t, \gamma_t, \bm{x}) - \bm{\epsilon}_t)^2 & \text{for } |\bm{\theta}_t(\bm{y}_t, \gamma_t, \bm{x}) - \bm{\epsilon}_t| \leq \delta, \\
\delta(|\bm{\theta}_t(\bm{y}_t, \gamma_t, \bm{x}) - \bm{\epsilon}_t| - \frac{1}{2}\delta) & \text{otherwise.}
\end{cases}
\end{gather}
Finally, $\mathcal{L}_{\delta}$ is flattened and averaged to obtain a single loss value:
\begin{gather}
    \mathcal{L} = \frac{1}{N}\sum_{i=1}^{N} \ell_{\delta, i}
\end{gather}
where $\ell_{\delta, i}$ is the $i$-th element in the flattened $\mathcal{L}_{\delta}$ vector and $N$ is the number of elements in $\mathcal{L}_{\delta}$.

\subsection{Denoising Diffusion Implicit Models}
\label{sec:apx:ddim}
Song et al. \cite{Song2020} proposed Denoising Diffusion Implicit Models (DDIM). It consists of jumping steps in the reverse diffusion process to save computing resources while maintaining acceptable sampling quality. They observed that, when reducing the number of steps, setting $\eta = 0$ in Eq. \ref{eq:diff:ytm1hat} yielded higher quality samples. This renders the reverse diffusion process deterministic, i.e., given fixed inputs, the output of the reverse diffusion is always the same. DDIM redefines the noise scheduling sequence $\bm{\gamma}$ by selecting a subset of the $T$ defined timesteps. In our experiments, we follow their linear selection procedure, which consists of selecting equally-spaced points of $\bm{\gamma}$. Such spacing between points is defined by the total number of desired sampling steps $T^{\text{DDIM}}$, which is a divisor of the original sampling steps $T$. Thus, the new noise schedule sequence $\bm{\gamma}^{\text{DDIM}}$ is defined as
\begin{gather}
    \gamma^{\text{DDIM}}_{t} = \gamma_{at} \:\:\: \forall \: 0 \leq t < T^{\text{DDIM}} \\
    \text{where } a = \frac{T^{\text{DDIM}}}{T} \text{.}
\end{gather}

\subsection{Classifier-Free Guidance}
\label{sec:apx:cfg}
Classifier-free guidance is an approach that enhances the controllability of conditional diffusion models.
Introduced in \cite{ho2022classifierfree}, it allows driving the reverse diffusion process towards desired data characteristics using input conditional data, as conditional diffusion models tend to ignore the conditioning input and produce samples that disregard the condition. The key idea is to train the model in a way that it can operate in two modes: one with guidance---conditioned on some input data---and one without---unconditioned. This is achieved by randomly dropping the conditioning input during training, which encourages the model to learn an internal representation that can be manipulated at sampling time to control the generation.

During training, $\bm{x}$ is dropped with probability $p$, effectively setting $\bm{x} = \O$. At sampling time, a guidance scale $\omega$ is introduced to amplify or diminish the influence of $\bm{x}$ by modifying the noise prediction $\bm{\theta}_t(\bm{y}_t, \gamma_t, \bm{x}) = \hat{\bm{\epsilon}}_t$:
\begin{gather}
    \tilde{\bm{\epsilon}}_t = (1 + \omega)\bm{\theta}_t(\bm{y}_t, \gamma_t, \bm{x}) - \omega \bm{\theta}_t(\bm{y}_t, \gamma_t, \O) \text{.}
\end{gather}
The modified noise matrix replaces $\bm{\theta}_t(\bm{y}_t, \gamma_t, \bm{x})$ in Eq. \ref{eq:diff:ytm1hat}.

\subsection{Consistent Diffusion Models}
\label{sec:apx:cdm}
The concept of Consistent Diffusion Models \cite{Daras2023} addresses the challenge of sampling drift in denoising diffusion models. It focuses on enforcing a consistency property on the model during training. This approach comes from the observation that if $\bm{\epsilon}$ is fixed, the model should produce similar---consistent---predictions for similar diffusion timesteps. Following this line of thought, such consistency can be enforced by adding an additional loss term to the training loss.

Let $t$ and $t'$ be diffusion timesteps such that $|t - t'| < \varepsilon_{\text{consist}}$. The timestep $t'$ is uniformly sampled from the interval $[t-\varepsilon_{\text{consist}}, t]$.
Following the line of thought of \cite{Daras2023}, the following loss term is summed to the diffusion loss:
\begin{gather}
\mathcal{L}_{\text{consist}} =    \lambda_{\text{consist}} 
 \mathcal{L}\left(\hat{\bm{y}}_{0,t}, \hat{\bm{y}}_{0,t'}\right)
\end{gather}
where $\mathcal{L}$ is the chosen diffusion loss function, $\hat{\bm{y}}_{0,t}$ and $\hat{\bm{y}}_{0,t'}$ are the estimations of $\bm{y}_{0}$ at diffusion steps $t$ and $t'$ respectively. For a single training step, $n_{\text{consist}}$ timesteps are sampled from $[t-\varepsilon_{\text{consist}}, t]$ and thus $n_{\text{consist}}$ consistent loss terms are added to the diffusion loss.

\subsection{Min-SNR Weighting Strategy}
\label{sec:apx:minsnr}
The Min-SNR Weighting Strategy \cite{Hang2023} tackles the issue of slow training convergence in denoising diffusion models caused by conflicting optimization directions across different timesteps.
To harmonize these trajectories and accelerate convergence, a clamped signal-to-noise ratio (SNR) is used to adaptively assign loss weights to each timestep. This method effectively suppresses the gradient conflicts among timesteps, leading to a more efficient training process.

Let $t$ be a diffusion timestep whose associated noise schedule value is $\gamma_t$, and $\gamma_{\text{SNR}} \in \mathds{R}_{++}$ be a clamping parameter. For the denoising diffusion objective, the Min-SNR weighting strategy defines the loss weight for timestep $t$ as:
\begin{gather}
w_t = \min\left(\frac{\gamma_{\text{SNR}}}{\text{SNR}_t}, 1\right) \\
\text{where } \text{SNR}_t = \frac{\gamma_t}{1 - \gamma_t}\text{.}
\end{gather}
This weighting scheme ensures that the model does not focus excessively on training samples with low noise values, which results in a significant convergence speedup, as observed in the experiments presented in \cite{Hang2023}. The calculated weights are multiplied by loss values for each sample in a training batch.

\end{appendices}
\vskip -2\baselineskip plus -1fil
\begin{IEEEbiography}
	[{\includegraphics[width=1in,height=1.25in,
	clip,keepaspectratio]
		{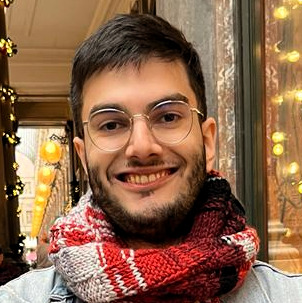}}]{João Gabriel Vinholi} received his Engineering degree in Electrical Engineering from the Federal University of Santa Catarina (UFSC), Florianópolis, Brazil, in 2020. From 2021 to 2023, he worked at ICEYE, Luxembourg, as a machine learning engineer. During his tenure, he focused on developing deep learning-based algorithms for multi-temporal flood mapping, deforestation mapping, building damage detection, and anomaly detection in X-band SAR imagery. He also specialized in MLOps for deploying these algorithms efficiently. Since 2023, he has been with the Luxembourg Institute of Science and Technology (LIST), Esch-sur-Alzette, Luxembourg, as a machine learning scientist, where he continues to conduct research and development of machine learning-based tools for remote sensing applications. Concurrently, since 2021, Vinholi has been a Doctoral Candidate of Electronics and Computer Engineering at the Aeronautics Institute of Technology (ITA), São José dos Campos, Brazil. His research interests lie in the fields of machine and deep learning, computer vision, digital signal processing, and their applications in remote sensing.
\end{IEEEbiography}
\vskip -2\baselineskip plus -1fil
\begin{IEEEbiography}
	[{\includegraphics[width=1in,height=1.25in,clip,keepaspectratio]
		{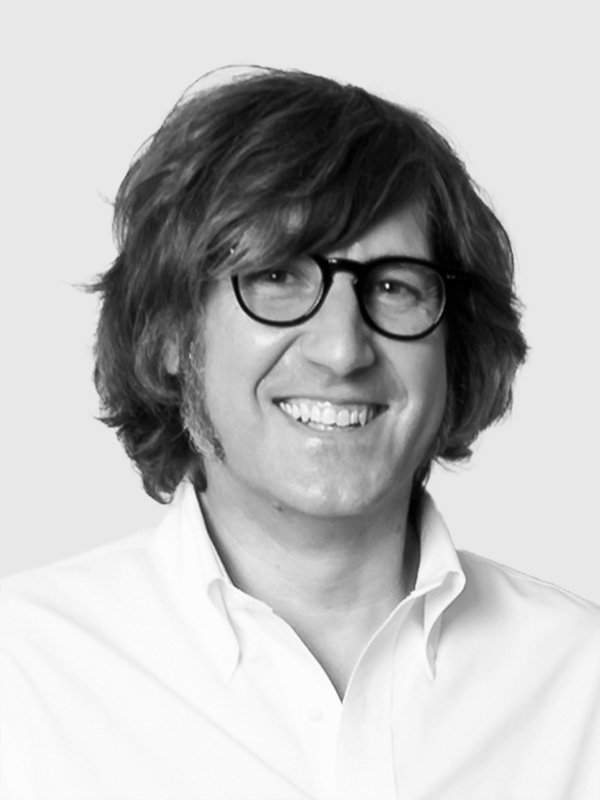}}]{Marco Chini} (Senior Member, IEEE) received the M.S. degree in electronic engineering from the Sapienza University of Rome, Rome, Italy, in 2003, and the Ph.D. degree in geophysics from the University of Bologna, Bologna, Italy, in 2008. From 2003 to 2012, he collaborated with the Department of Information Engineering, Electronics and Telecommunications, Sapienza University of Rome. From 2006 and 2008, he was a Visiting Researcher with the Department of Aerospace Engineering Science, University of Colorado, Boulder, CO, USA, where he was involved in very high-resolution optical images for urban monitoring. From 2008 to 2012, he was with the Remote Sensing Group, Istituto Nazionale di Geofisica e Vulcanologia, Rome, where his research topic was the development of Earth Observation (EO)-based innovative algorithms for detecting damages caused by earthquakes. Since 2013, he has been with the Luxembourg Institute of Science and Technology (LIST), Esch-sur-Alzette, Luxembourg, formerly the Centre de Recherche Public—Gabriel Lippmann. He has been involved in projects for mapping flooded areas, detecting damage caused by earthquakes, monitoring urban sprawl, detecting surface changes, and identifying risk-/hazard-prone areas. His research interests include analysis of multitemporal data, deep learning, feature extraction, data fusion and segmentation using synthetic aperture radar and optical data, soil moisture retrieval, and SAR interferometry technique for geophysical applications. Dr. Chini is an Associate Editor of the IEEE TRANSACTIONS ON GEOSCIENCE AND REMOTE SENSING, an Editorial Advisor of Springer’s publication program in the Remote Sensing/Photogrammetry, and an Associate Editor of the European Journal of Remote Sensing.
\end{IEEEbiography}
\vskip -2\baselineskip plus -1fil

\begin{IEEEbiography}
	[{\includegraphics[width=1in,height=1.25in,
	clip,keepaspectratio]
		{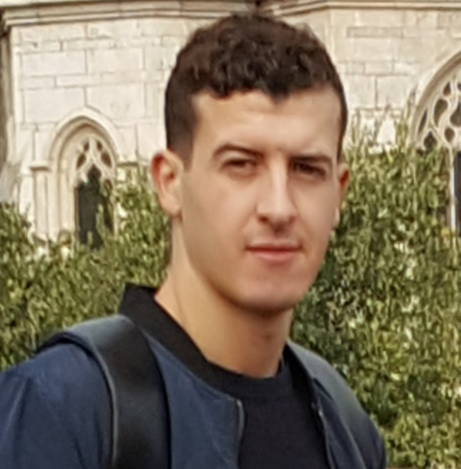}}]{Anis Amziane} received the M. S. degree in Signal and Image processing from the University of Burgundy, Dijon, France, in 2018, and the Ph.D. degree in Computer Engineering and Signal/Image processing from the University of Lille, Lille, France, in 2022. From 2021 to 2022, he was a Research and Teaching Associate at the Faculty of Science and Technology of the University of Lille, France. In 2023, he joined the Luxembourg Institute of Science and Technology (LIST), Esch-sur-Alzette, Luxembourg, as a Postdoctoral researcher, where he conducts the research and development of machine learning algorithms for remote sensing image super resolution. His research interests include multispectral proximal/remote imaging, illumination-robust feature extraction, spectral filter array (SFA) image analysis, and image super resolution.
\end{IEEEbiography}
\vskip -2\baselineskip plus -1fil

\begin{IEEEbiography}
	[{\includegraphics[width=1in,height=1.25in,clip,keepaspectratio]
		{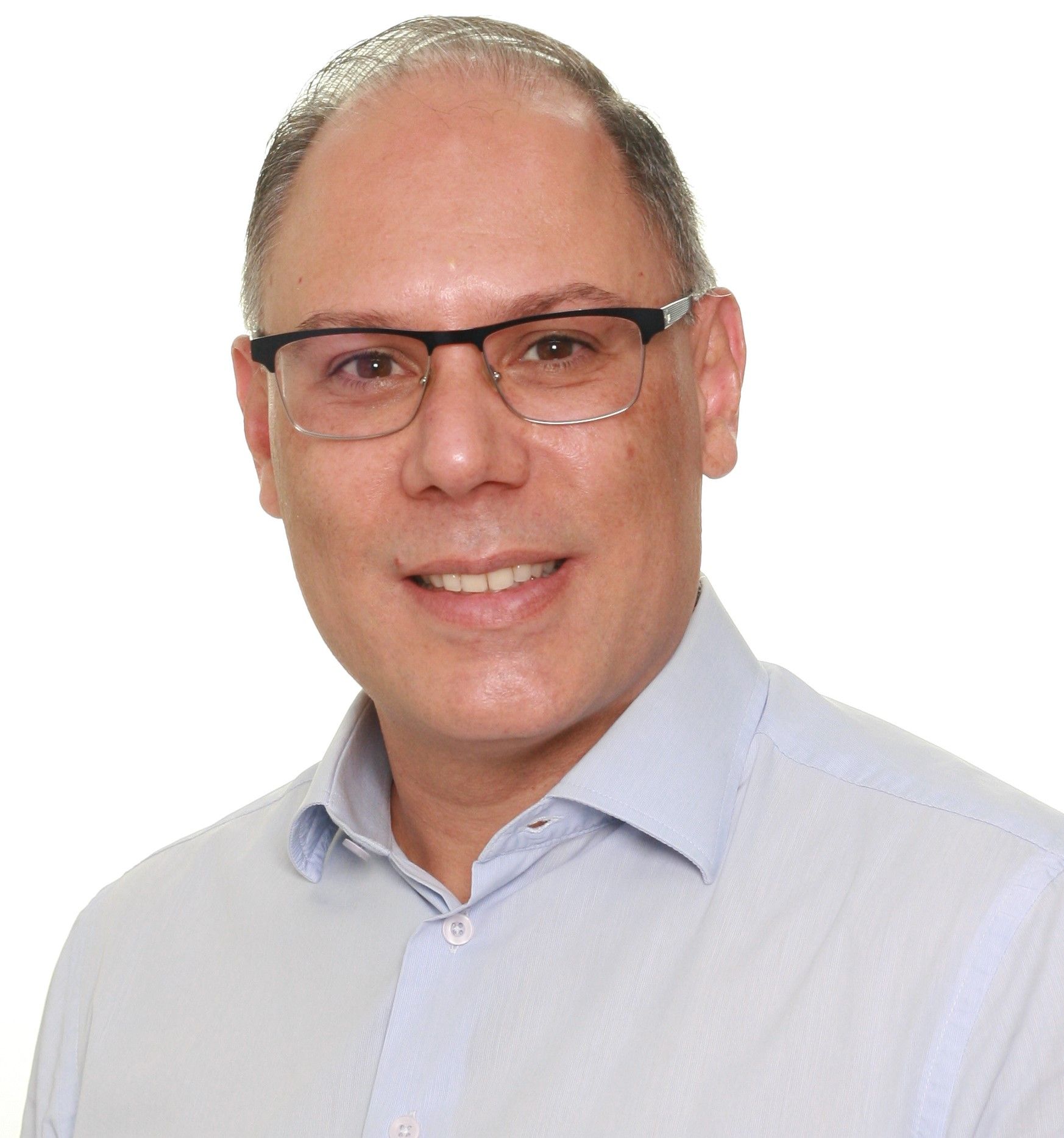}}]{Renato~Machado}
		received the B.S.E.E. degree from the São Paulo State University (UNESP),
Ilha Solteira, SP, Brazil, in 2001. He received the M.Sc. and Ph.D. degrees in Electrical
Engineering from the Federal University of Santa Catarina (UFSC), Florianópolis, SC, Brazil, in
2004 and 2008, respectively. From November 2013 to February 2015, Dr. Machado was a
visiting Research Fellow at Blekinge Institute of Technology (BTH) in partnership with Saab AB.
From August 2009 to December 2017, he was an Assistant (2008-2016) and Associate (2017)
Professor at the Federal University of Santa Maria, RS, Brazil, where he lectured many courses
in bachelor and graduate programs and assumed different positions in the institution, namely,
Researcher leader of the Communications and Signal Processing Research Group, Coordinator
of the Telecommunications Engineering Program, and Director of the Aerospace Science
Laboratory. Since December 2017, Dr. Machado has been an Associate Professor at the
Aeronautics Institute of Technology (ITA). He is the Research leader of the Digital and Signal
Processing Laboratory, ITA, and SAR and Radar Signal Processing Laboratory, ITA. Prof.
Machado is a Principal Investigator of BI0S (Brazilian Institute of Data Science), a Research
Center for Applied Science and Technology in Artificial Intelligence. - FAPESP/MCTI/CGI.
Between August 2021 and July 2022, he was the head of the Graduate Program in Electronics
and Computer Engineering, ITA. Since June 2022, Prof. Machado has been the head of the
Electronic Engineering Division, ITA. In February 2021, Dr. Machado received the Senior
Membership degree from the Institute of Electrical and Electronics Engineers (IEEE). His
research interests include SAR Processing, SAR Image Processing, Change Detection, Radar
Signal Processing, and Digital Signal Processing.
\end{IEEEbiography}

\vskip -2\baselineskip plus -1fil

\begin{IEEEbiography}
	[{\includegraphics[width=1in,height=1.25in,
	clip,keepaspectratio]
		{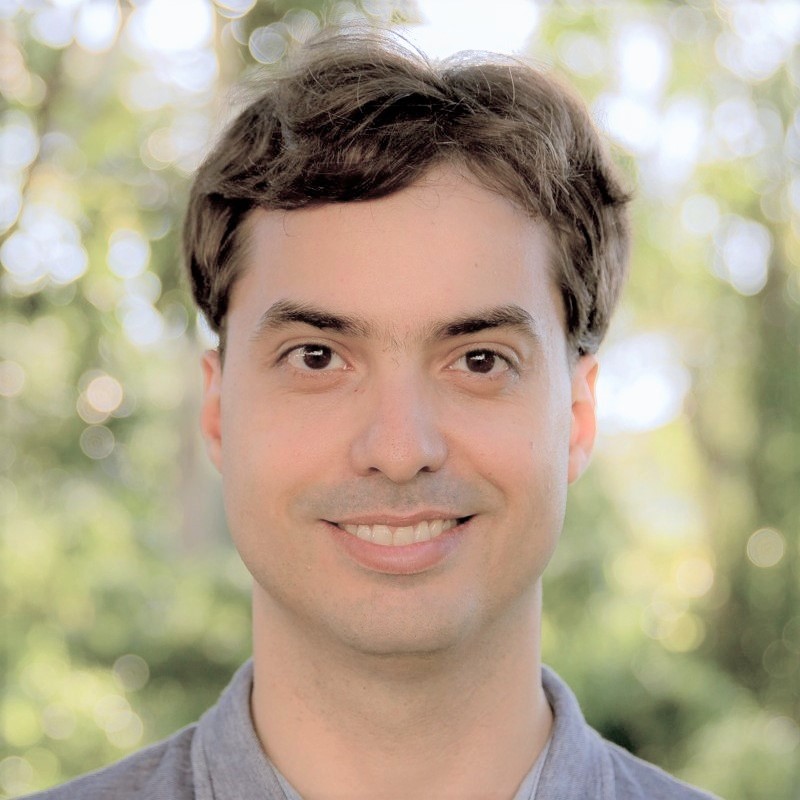}}]{Danilo Silva} received the B.Sc.\ degree from the Federal University of Pernambuco (UFPE), Recife, Brazil, in 2002, the M.Sc.\ degree from the Pontifical Catholic University of Rio de Janeiro (PUC-Rio), Rio de Janeiro, Brazil, in 2005, and the Ph.D. degree from the University of Toronto, Toronto, Canada, in 2009, all in electrical engineering.

        From 2009 to 2010, he was a Postdoctoral Fellow at the University of Toronto, at the \'Ecole Polytechnique F\'ed\'erale de Lausanne (EPFL), and at the State University of Campinas (UNICAMP). In 2010, he joined the Department of Electrical and Electronic Engineering, Federal University of Santa Catarina (UFSC), Florian\'opolis, Brazil, where he is currently an Associate Professor. His current research interests include machine learning, computer vision, signal processing, and their applications. His previous research interests include wireless communications, network coding, channel coding, and information theory.
        
        Dr. Silva is a member of the Brazilian Telecommunications Society (SBrT). He was a recipient of a CAPES Ph.D. Scholarship in 2005, the Shahid U.H.\ Qureshi Memorial Scholarship in 2009, and a FAPESP Postdoctoral Scholarship in 2010. He also served as Technical Program Co-Chair of the XXXVIII Brazilian Symposium on Telecommunications and Signal Processing (SBrT 2020).
\end{IEEEbiography}
\vskip -2\baselineskip plus -1fil

\begin{IEEEbiography}
	[{\includegraphics[width=1in,height=1.25in,
	clip,keepaspectratio]
		{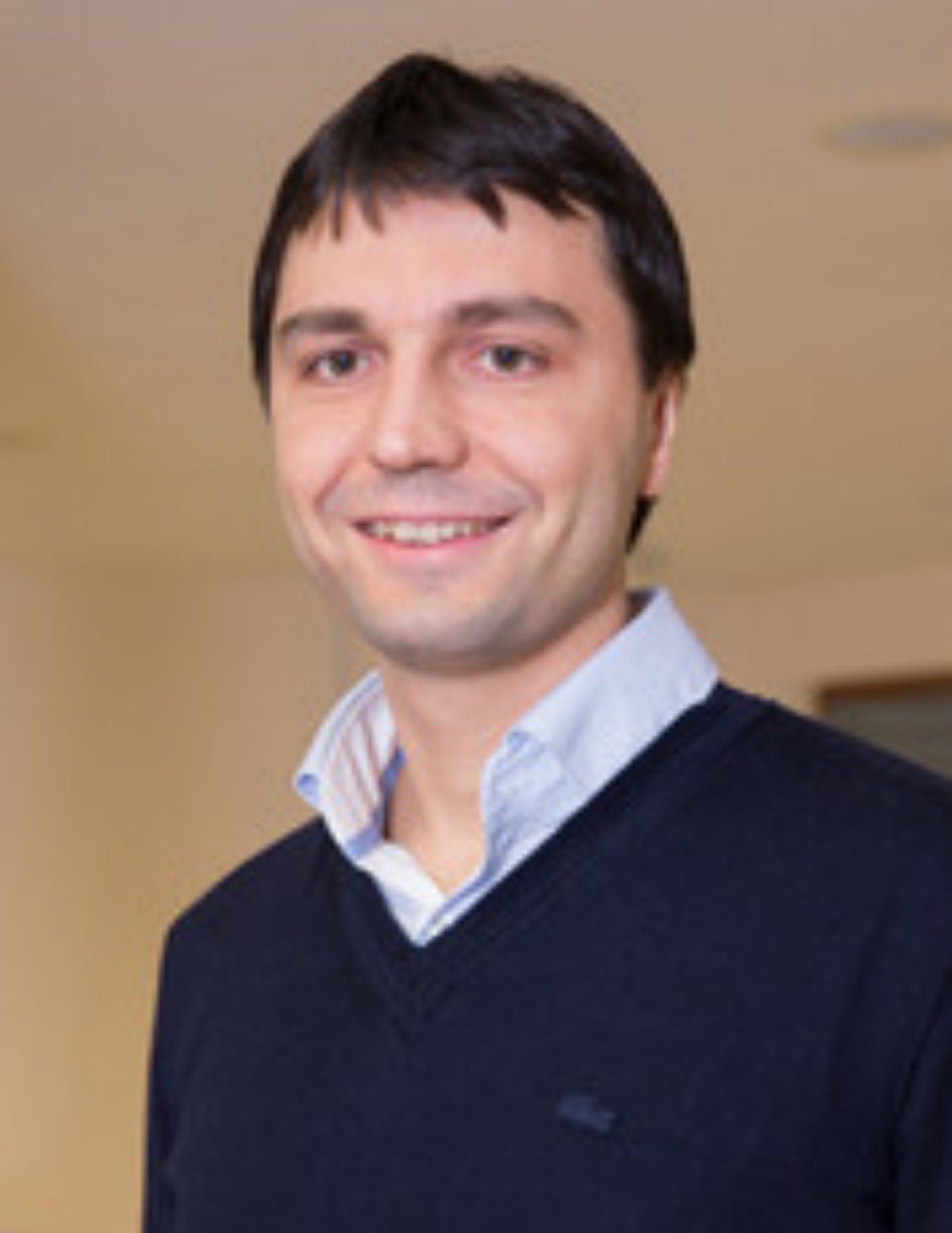}}]{Patrick Matgen}
Patrick Matgen earned his M.Sc. in Environmental Engineering at the Ecole Polytechnique Fédérale de Lausanne (Switzerland) in 2002. In 2011 he finalized his Ph.D dissertation “Surface and subsurface water from space: on the integration of microwave remote sensing observations with flood prediction systems” at the Technical University in Delft (The Netherlands). As a Project Manager from 2002 and then as a Lead Research and Technology Associate from 2015 until 2020 he was responsible for acquiring funding and managing Research Development and Innovation projects in Earth Observation and hydrological-hydraulic modelling within the research unit ‘Environmental sensing and modelling' of the Luxembourg Institute of Science and Technology (LIST). Since January 2020 he is leading the group “remote sensing and natural resources modelling” of the research unit. The group’s aim is to develop the synergistic use, processing and interpretation of data from multiple complementary active and passive sensors installed on both space- and airborne platforms and to provide deeper insights into the relations between spectral features and properties of Earth’s natural resources. Patrick Matgen’s personal research interests include the assimilation of remote sensing-derived observations into hydrodynamic models and the assessment of flood hazard and risk at large scale. 
\end{IEEEbiography}
\vskip -2\baselineskip plus -1fil

\end{document}